\pdfoutput=1
\documentclass[review]{elsarticle}
\usepackage{subfigure}
\usepackage{longtable}
\usepackage{algorithm, algorithmic}
\usepackage{bm}
\usepackage{booktabs}
\usepackage{multirow}
\usepackage{geometry}
\usepackage{amsfonts,amssymb}
\usepackage{float}

\usepackage{lineno,hyperref}
\modulolinenumbers[5]

\begin{document}
%%%%%%%%%%%%%%%%%%%%%%%%%%%%%%%%%%%%%%%%%%%%%%%%%%%%%%%%%%%%%%%%%%%%%
\begin{frontmatter}
\title{GMM Discriminant Analysis with Noisy Label for Each Class}

\author[mymainaddress]{Jian-wei Liu\corref{mycorrespondingauthor}}
\cortext[mycorrespondingauthor]{Corresponding author}%???
\ead{liujw@cup.edu.cn}%???

\author[mymainaddress]{Zheng-ping Ren}

\author[mymainaddress]{Run-kun Lu}

\author[mymainaddress]{Xiong-lin Luo}

\address[mymainaddress]{Department of Automation, College of Information Science and Engineering, China University of Petroleum , Beijing, Beijing, China}
	\begin{abstract}
		Real world datasets often contain noisy labels, and learning from such datasets using standard classification approaches may not produce the desired performance. In this paper, we propose a Gaussian Mixture Discriminant Analysis (GMDA) with noisy label for each class. We introduce flipping probability and class probability and use EM algorithms to solve the discriminant problem with label noise. We also provide the detail proofs of convergence. Experimental results on synthetic and real-world datasets show that the proposed approach notably outperforms other four state-of-art methods.
	\end{abstract}
\end{frontmatter}
%%%%%%%%%%%%%%%%%%%%%%%%%%%%%%%%%%%%%%%%%%%%%%%%%%%%%%%%%%%%%%%%%%%%%%%%%%%%
\section{INTRODUCTION}%1
Noisy label problem have been investigated for a long time in the machine learning literature and label noise-robust algorithms have numerous applications in medical image processing, spam filtering \cite{1nettleton2010study,2frenay2013classification,3brodley1999identifying}, Alzheimer disease prediction\cite{1nettleton2010study}, gene expression classification \cite{4libralon2009pre} , text processing \cite{8abellan2010bagging,19mathews2019gaussian,20Hastie2001TheEO,21bootkrajang2012label,25kearns1998efficient,27li2007classification,30long2010random}, image recognition \cite{34van2015learning,35du2015modelling,37scott2013classification,38scott2013classification,41manwani2013noise,421995Can}. Noisy labels are introduced by expert error and other unknown and unexpected factors. Mislabeled instances may lead to various potential negative consequences: bias the learning process, debase the prediction accuracy, and increase algorithm complexity of inferred models \cite{3brodley1999identifying,4libralon2009pre} and the number of necessary labeling training samples, which is often produced by an expensive and time-consuming hand-annotation process or inefficient automatic annotation \cite{1nettleton2010study,2frenay2013classification}, and increase difficulties in feature selection \cite{5zhang2006method,6shanab2012robustness}. The methods to deal with label noise can be classified into three \cite{1nettleton2010study}: 1) the label noise is ignored, and approaches that are robust to the presence of label noise, such as ensemble AdaBoost \cite{72001Soft} and decision trees \cite{8abellan2010bagging}, are searched; 2) mislabeled instances are detected and removed, and then cleaned training samples \cite{9wilson2000reduction,10brodley1996identifying} are used to learn; and 3) models considering label noise are designed, and label noise-tolerant methods are determined. Label noise-tolerant methods enable researchers to take advantage of noise knowledge and use more sample information than noise-cleansing methods. The disadvantages are the increment in algorithm complexity and the increase in the number of parameters to estimate.\\

Bootkrajang presented a robust normal discriminant analysis (rNDA) algorithm \cite{11bootkrajang2013supervised}. The algorithm solves the maximum likelihood estimate problem by employing the EM (Expectation Maximization) algorithm \cite{12moon1996expectation,46dempster1977maximum,471982On}. The rNDA model assumes that the examples in each class obey single Gaussian distribution; it is scarcely to verify. Thus, its performance on datasets that are not strictly Gaussian in each class seems insufficient.\\

Numerous studies on GMM have appeared in many fields, such as outlier mining \cite{14lathuiliere2018deepgum}, image processing \cite{16li2020dividemix,17kim2019nlnl}, clustering \cite{18lawrence2001estimating} and community detection\cite{19mathews2019gaussian}. \cite{14lathuiliere2018deepgum} devised a approach to adapt to a continuously evolving outlier distribution. \cite{15melnykov2012initializing} proposed initializing mean vectors by choosing points with higher concentrations of neighbors, and using a truncated normal distribution for the preliminary estimation of dispersion matrices. DivideMix models the per-sample loss distribution with a mixture model to dynamically divide the training data into a labeled set with clean samples and an unlabeled set with noisy samples \cite{16li2020dividemix}. \cite{17kim2019nlnl} addressed noisy labels issue and proposed selective negative learning and positive learning approach trained using a complementary label. \cite{18lawrence2001estimating} constructed a kernel Fisher discriminant (KFD) from training examples with noisy labels. \cite{19mathews2019gaussian} presented a procedure for community detection using GMMs that incorporates certain truncation and shrinkage effects that arise in the non-vanishing noise regime.\\

To solve this problem, we propose a new scheme to carry out the discriminant analysis with Gaussian mixture models (GMM) which has the ability to handle the non-Gaussian distributions. We employ a linear combination of Gaussian distributions to approximate the probabilistic distributions in each class and use the EM algorithm to solve the maximum likelihood estimate \cite{13boykin1999proceedings,46dempster1977maximum,471982On}. In the last several decades, researchers in the fields of statistics and computer vision have been interested in GMM.\\

The discriminant analysis discussed in this paper uses GMM to approximate data distributions and is applied to classification in the case of label noise. Maximum likelihood estimate method is used to determine the parameters. Moreover, this study derives the updating formulas of the parameters of the proposed Gaussian Mixture Discriminant Analysis (GMDA). The performance of GMDA is then compared with that of AdaBoost, rNDA, rLR, and rmLR \cite{19mathews2019gaussian} on two synthetic and six real-world datasets. Results show that our method can effectively and correctly estimate the parameters of both distribution and noise.\\

Our main contributions are as follows:\\
1) We propose a general discriminant analysis framework for attacking the noisy label problem. Different from previous approach, in this framework, the probabilistic distribution of each class on real data is captured by GMM instead of the single Gaussian distribution, this single Gaussian assumption for each class is apparently too harsh to be verified, and can scarcely reflect the actual scenarios.\\
2) We show that when flipping probability and class probability is introduced, the parameters of GMDA model and posterior probability to predict an unlabeled instance can be computed by using EM algorithm.\\
3) We provide the detail proofs of convergence for general situation, e.g., Gaussian classes with noisy labels and class-conditional Gaussian mixtures with noisy labels.\\
4) We have conducted extensive experiments on two synthetic datasets and six real-life datasets, which have different properties and scales, to demonstrate the effectiveness and efficiency of our proposed formulation.

The rest of this paper is organized as follows: the proposed GMMs discriminant analysis with noisy label for each class is formally introduced in Section II, convergence analysis for general situation, Gaussian classes with noisy labels and class-conditional Gaussian mixtures with noisy labels is presented in Section III, and related work in Section IV. Experimental results using synthetic and real-world datasets are discussed in Section V. Finally, the conclusion and future work are summarized in Section VI.

\section{DISCRIMINANT ANALYSIS BASED ON GAUSSIAN MIXTURE MODELS}%2
\subsection{Description of the problem with the noise labels}%2.1
Considering a statistical decision problem (pattern recognition, classification, and discrimination), we assume that some real data vectors 
\[{\bf{x}} = \left( {{x_1},...,{x_d}} \right) \in {\cal X},\quad \left( {{\cal X} = {{\cal R}^d}} \right)\]
have to be classified with respect to a finite set of classes $\Omega  = \left\{ {{\omega _1},{\omega _2}, \cdots ,{\omega _K}} \right\}$ . The data vectors ${\bf{x}} \in {\cal X}$  are supposed to occur randomly according to some unknown class-conditional pdfs $p\left( {{\bf{x}}|\omega } \right)$  and the respective priori class probabilities $p\left( \omega  \right),\;\omega  \in \Omega $. 

In case of supervised learning we are given a training set ${{\cal S}_\omega }$  for each class $\omega  \in \Omega $ :
\[{{\cal S}_\omega } = \left\{ {{\bf{x}} \in {\cal X}} \right\},\;\omega  \in \Omega \,\,;\;{\cal S} = \bigcup\limits_{\omega  \in \Omega } {{{\cal S}_\omega }} ,\;\left| {\cal S} \right| = \sum\limits_{\omega  \in \Omega } {\left| {{{\cal S}_\omega }} \right|} \]
where $\left| {\cal S} \right|$ and $\left| {{{\cal S}_\omega }} \right|$  denote the number of elements in set  ${{\cal S}_\omega }$. The decision problem can be solved by means of Bayes decision function by computing the maximum-likelihood estimates of the conditional densities $p\left( {{\bf{x}}|\omega } \right)$ . The related log-likelihood criterion ${L_\omega }$  is given by
\[{L_\omega } = \frac{1}{{\left| {{{\cal S}_\omega }} \right|}}\sum\limits_{{\bf{x}} \in {{\cal S}_\omega }} {log} p\left( {{\bf{x}}|\omega } \right)p\left( \omega  \right),\quad \left( {\omega  \in \Omega } \right)\]
with normalization coefficient $1/\left| {{{\cal S}_\omega }} \right|$  included for convenience. 

In case of noisy labels we assume that the true label $\omega  \in \Omega $  of a given observation ${\bf{x}} \in {\cal S}$  may be randomly interchanged (flipped, corrupted, substituted), i.e. for each data vector ${\bf{x}} \in {\cal S}$  we are given a unique observed label $\tilde \omega  \in \Omega $ , which may differ from the true label $\omega $ . Obviously, the probabilities $p(\tilde \omega )$  of the observed labels $\tilde \omega $  may differ from the true probabilities $p\left( \omega  \right)$  of the true labels $\omega $ . Assuming randomly substituted labels we denote $p\left( {\omega |\tilde \omega } \right)$  is the probability of the true label $\omega $  given the observed label  $\tilde \omega $. In this sense the conditional probability density  $\tilde p\left( {{\bf{x}}|\tilde \omega } \right)$ of  ${\bf{x}} \in {\cal X}$ given an observed label $\tilde \omega $  is a mixture
\[\tilde p\left( {{\bf{x}}|\tilde \omega } \right) = \sum\limits_{\omega  \in \Omega } {p\left( {{\bf{x}}|\omega } \right)} p\left( {\omega |\tilde \omega } \right),{\bf{x}} \in {\cal X},\;\tilde \omega  \in \Omega \]
consequently, with the probability $p\left( {\omega |\tilde \omega } \right)$ , any of the classes $\omega  \in \Omega $  can be the true source of the observation ${\bf{x}} \in {\cal X}$ . Note that the probabilities $p\left( {\omega |\tilde \omega } \right)$ , $\omega  \in \Omega $  , $p(\tilde \omega )$  , $\tilde \omega  \in \Omega $  are generally unknown and have to be estimated from data.
\subsection{Gaussian mixture model}%2.2
Lawrence and Schölkopf [19] proposed a probabilistic approach to label noise, and Bootkrajang \cite{10brodley1996identifying}, \cite{21bootkrajang2012label} extended the same model to multi-class case, assuming a Gaussian density for each class. We propose discriminant analysis based on GMM where GMM is used to approximate the probabilistic distribution of each class on real data.
Supposed that the data log-likelihood is:
\[\begin{array}{l}
L = \log \prod\limits_{{\bf{x}} \in {\cal S}} {p\left( {{\bf{x}}\left| {\tilde \omega ,{\theta _{\tilde \omega }}} \right.} \right)p\left( {\tilde \omega } \right)} \\
\;\;\; = \sum\limits_{{\bf{x}} \in {\cal S}} {\log } p\left( {{\bf{x}}\left| {\tilde \omega ,{\theta _{\tilde \omega }}} \right.} \right)p\left( {\tilde \omega } \right)
\end{array}\]
under our assumption conditions, the given model may be no longer valid. Thus, a hidden variable $\omega  \in \Omega $  is introduced to address the problem. The hidden variable $\omega $  is considered as the real class label:
\begin{equation}%%%公式1
	\begin{array}{l}
	L = \sum\limits_{{\bf{x}} \in {\cal S}} {\log } \sum\limits_{\omega  \in \Omega } {p\left( {{\bf{x}},\omega \left| {\tilde \omega ,{\theta _{\tilde 			\omega }}} \right.} \right)p\left( {\tilde \omega } \right)} \\
	\;\;\; = \log \prod\limits_{{\bf{x}} \in {\cal S}} {\sum\limits_{\omega  \in \Omega } {p\left( {{\bf{x}},\omega \left| {\tilde \omega ,{\theta 				_{\tilde \omega }}} \right.} \right)p\left( {\tilde \omega } \right)} } 
	\end{array}
\end{equation}

The observed labels are generated from true labels and random noise. The joint probability $\omega $ and  $\tilde \omega $ can be expressed as $p\left( {\omega ,{{\tilde \omega }_n}} \right) = p\left( {{{\tilde \omega }_n}\left| \omega  \right.} \right)p\left( \omega  \right)$. Input x is conditionally independent from observed label ${\tilde \omega _n}$  after knowing true label  $\omega $, because the label noise is random. Thus, we change Eq.(1) into
\begin{equation}%%%公式2
	\begin{array}{c}
		L\left( \Theta  \right) = \sum\limits_{{\bf{x}} \in {\cal S}} {\log \sum\limits_{\omega  \in \Omega } {p\left( {x\left| {\tilde \omega ,\omega ,			{\theta _{{{\tilde \omega }_n}}}} \right.} \right)p\left( {{{\tilde \omega }_n},\omega } \right)} } \\
		 = \sum\limits_{{\bf{x}} \in {\cal S}} {\log \sum\limits_{\omega  \in \Omega } {p\left( {x\left| {\omega ,{\theta _\omega }} \right.} 					\right)p\left( {\tilde \omega \left| \omega  \right.} \right)p\left( \omega  \right)} } \\
		 = \sum\limits_{\tilde \omega  \in \Omega } {I\left( {\tilde \omega  = \omega } \right)\sum\limits_{{\bf{x}} \in {\cal S}} {\log 						\sum\limits_{\omega  \in \Omega } {p\left( {x\left| {\omega ,{\theta _\omega }} \right.} \right)} } } p\left( {\tilde \omega \left| \omega  				\right.} \right)p\left( \omega  \right)
\end{array}
\end{equation}
where $I\left(  \cdot  \right)$  is an indicator function. Flipping probability is defined as ${\gamma _{\tilde \omega ,\omega }}\mathop  = \limits^{def} p\left( {\tilde \omega \left| \omega  \right.} \right)$  ,which is similar to reference [11], indicating the probability of true label $\omega $  flipped to the observed label $\tilde \omega $ . Class probability is defined as ${\pi _\omega }\mathop  = \limits^{def} p\left( \omega  \right)$, and the constraint conditions are  $\sum\limits_{\tilde \omega  \in \Omega } {{\gamma _{\tilde \omega ,\omega }}}  = 1$ and $\sum\limits_{\omega  \in \Omega } {{\pi _\omega }}  = 1$ . The set of flipping probability and class probability are denoted as $\Gamma  = {\left\{ {{\gamma _{\tilde \omega ,\omega }}} \right\}_{\tilde \omega  \in \Omega }}$ and $\Pi  = {\left\{ {{\pi _\omega }} \right\}_{\omega  \in \Omega }}$, respectively. The Gaussian mixture density function is 
\[p\left( {{\bf{x}}\left| {\omega ,{\theta _\omega }} \right.} \right) = \sum\limits_{m \in {\cal M}} {{w_m}} g\left( {{\bf{x}}\left| {\omega ,{{\bf{\mu }}_m},{{\bf{\Sigma }}_m}} \right.} \right)\]
where  ${\theta _\omega } = {\left\{ {{w_m},{{\bf{\mu }}_m},{{\bf{\Sigma }}_m}} \right\}_{m \in {\cal M}}}$ is the parameter set of the class $k$ , The elements are weight, mean vector, and covariance matrix of the component $m$,and $M$  is the number of components. The logarithm of a sum in (2) can be rewritten as:
\begin{equation}%%%公式3
	\begin{array}{l}
\log \sum\limits_{\omega  \in \Omega } {p\left( {{\bf{x}}\left| {\omega ;{\theta _\omega }} \right.} \right)p\left( {\tilde \omega \left| \omega  \right.} \right)p\left( \omega  \right)} \\
 = \log \sum\limits_{\omega  \in \Omega } {{\gamma _{\tilde \omega ,\omega }}{\pi _\omega }} \sum\limits_{m \in {\cal M}} {{w_m}} g\left( {{\bf{x}}\left| {\omega ,{{\bf{\mu }}_m},{{\bf{\Sigma }}_m}} \right.} \right)\\
 = \log \sum\limits_{\omega  \in \Omega } {q\left( \omega  \right)\frac{{{\gamma _{\tilde \omega ,\omega }}{\pi _\omega }\sum\limits_{m \in {\cal M}} {{w_m}} g\left( {{\bf{x}}\left| {\omega ,{{\bf{\mu }}_m},{{\bf{\Sigma }}_m}} \right.} \right)}}{{q\left( \omega  \right)}}} 
\end{array}
\end{equation}
where $q\left(  \cdot  \right)$  is an arbitrary distribution of  $\omega $, and the constraint condition is $\sum\limits_{\omega  \in \Omega } {q\left( \omega  \right){\rm{ = 1}}} $.

According to Jensen’s inequality \cite{22bishop1995neural}, if  $Z$ is a random variable, and $g\left(  \cdot  \right)$  is a concave function, then
\[g\left( {E\left( Z \right)} \right) \ge E\left( {g\left( Z \right)} \right)\]
Thus the lower bound of (3) is derived. Since $g\left(  \cdot  \right)$  is a concave function, and again by Jensen’s inequality we have:
\[\begin{array}{l}
f\left( {{E_{\omega \sim q\left( \omega  \right)}}\left( {\frac{{{\gamma _{\tilde \omega ,\omega }}{\pi _\omega }p\left( {{\bf{x}}\left| {\omega ,{\theta _\omega }} \right.} \right)}}{{q\left( \omega  \right)}}} \right)} \right)\\
 \ge {E_{\omega \sim q \left( \omega  \right)}}\left( {f\left( {\frac{{{\gamma _{\tilde \omega ,\omega }}{\pi _\omega }p\left( {{\bf{x}}\left| {\omega ,{\theta _\omega }} \right.} \right)}}{{q\left( \omega  \right)}}} \right)} \right)
\end{array}\]
i.e.
\[\begin{array}{l}
\log \sum\limits_{\omega  \in \Omega } {q\left( \omega  \right)\frac{{{\gamma _{\tilde \omega ,\omega }}{\pi _\omega }p\left( {{\bf{x}}\left| {\omega ,{\theta _\omega }} \right.} \right)}}{{q\left( \omega  \right)}}} \\
 \ge \sum\limits_{\omega  \in \Omega } {q\left( \omega  \right)\log \frac{{{\gamma _{\tilde \omega ,\omega }}{\pi _\omega }p\left( {{\bf{x}}\left| {\omega ,{\theta _\omega }} \right.} \right)}}{{q\left( \omega  \right)}}} 
\end{array}\]
A Lagrange multiplier ${\lambda _{{G_1}}}$  is introduced to find a local optimum of the objective function subject to $\sum\limits_{\omega  \in \Omega } {q\left( \omega  \right)}  = 1$, and the corresponding Lagrange function is:
\[\begin{array}{l}
{G_1} = \sum\limits_{\omega  \in \Omega } {q\left( \omega  \right)} \left[ {\log {\gamma _{\tilde \omega ,\omega }}{\pi _\omega }p\left( {{\bf{x}}\left| {\omega ,{\theta _\omega }} \right.} \right) - \log q\left( \omega  \right)} \right]\\
\;\;\;\;\;\;\; + {\lambda _{{G_1}}}\left[ {1 - \sum\limits_{\omega  \in \Omega } {q\left( \omega  \right)} } \right]
\end{array}\]
By setting the derivative w.r.t $q\left( \omega  \right)$  equals to zero, we obtain the formula as follow:
\begin{equation}%%%公式4
	\begin{array}{l}
\log p\left( {{\bf{x}}|\omega ,{\theta _\omega }} \right){\gamma _{\tilde \omega ,\omega }}{\pi _\omega } - \log q\left( \omega  \right) - 1 - {\lambda _{{G_1}}} = 0\\
 \Rightarrow \log q\left( \omega  \right) = \log p\left( {{\bf{x}}|\omega ,{\theta _\omega }} \right){\gamma _{\tilde \omega ,\omega }}{\pi _\omega } - 1 - {\lambda _{{G_1}}}\\
 \Rightarrow q\left( \omega  \right) = p\left( {{\bf{x}}|\omega ,{\theta _\omega }} \right){\gamma _{\tilde \omega ,\omega }}{\pi _\omega } \cdot {e^{ - \left( {1 + {\lambda _{{G_1}}}} \right)}}
\end{array}
\end{equation}
Further, we integrate $q\left( \omega  \right)$  in the field of $\Omega $ , and the optimal solution w.r.t  ${\lambda _{{G_1}}}$ is:
\[\begin{array}{l}
\sum\limits_{\omega  \in \Omega } {q\left( \omega  \right)}  = {e^{ - \left( {1 + {\lambda _{{G_1}}}} \right)}}\sum\limits_{\omega  \in \Omega } {p\left( {{\bf{x}}|\omega ,{\theta _\omega }} \right){\gamma _{\tilde \omega ,\omega }}{\pi _\omega }}  = 1\\
 \Rightarrow {\lambda _{{G_1}}} = \log \sum\limits_{\omega  \in \Omega } {p\left( {{\bf{x}}|\omega ,{\theta _\omega }} \right){\gamma _{\tilde \omega ,\omega }}{\pi _\omega }}  - 1
\end{array}\]
 ${\lambda _{{G_1}}}$ is plugged back into (4), and $q\left( \omega  \right)$  is solved as follows:
\begin{equation}%%%公式5
\begin{array}{c}
q\left( \omega  \right) = \frac{{p\left( {{\bf{x}}|\omega ,{\theta _\omega }} \right){\gamma _{\tilde \omega ,\omega }}{\pi _\omega }}}{{\sum\limits_{\omega  \in \Omega } {p\left( {{\bf{x}}|\omega ,{\theta _\omega }} \right){\gamma _{\tilde \omega ,\omega }}{\pi _\omega }} }}\\
 = p\left( {\omega \left| {x,\tilde \omega } \right.} \right)
\end{array}
\end{equation}
Jensen’s inequality is employed and the lower bound of $\log p\left( {{\bf{x}}\left| {\omega ,{\theta _\omega }} \right.} \right) = \log \sum\limits_{m \in {\cal M}} {{w_{\omega ,m}}} g\left( {{\bf{x}}\left| {\omega ,{{\bf{\mu }}_{\omega ,m}},{{\bf{\Sigma }}_{\omega ,m}}} \right.} \right)$ is derived as
\[\begin{array}{l}
\log p\left( {{\bf{x}}\left| {\omega ,{\theta _\omega }} \right.} \right)\\
 = \log \sum\limits_{m \in {\cal M}} {{w_{\omega ,m}}} g\left( {{\bf{x}}\left| {\omega ,{{\bf{\mu }}_{\omega ,m}},{{\bf{\Sigma }}_{\omega ,m}}} \right.} \right)\\
 = \log \sum\limits_{m \in {\cal M}} {{h_{\omega ,m}}\frac{{{w_{\omega ,m}}g\left( {{\bf{x}}\left| {\omega ,{{\bf{\mu }}_{\omega ,m}},{{\bf{\Sigma }}_{\omega ,m}}} \right.} \right)}}{{{h_{\omega ,m}}}}} \\
 \ge \sum\limits_{m \in {\cal M}} {{h_{\omega ,m}}\log \frac{{{w_{\omega ,m}}g\left( {{\bf{x}}\left| {\omega ,{{\bf{\mu }}_{\omega ,m}},{{\bf{\Sigma }}_{\omega ,m}}} \right.} \right)}}{{{h_{\omega ,m}}}}} \\
 = \sum\limits_{m \in {\cal M}} {{h_{\omega ,m}}\left[ {\log {w_{\omega ,m}}g\left( {{\bf{x}}\left| {\omega ,{{\bf{\mu }}_{\omega ,m}},{{\bf{\Sigma }}_{\omega ,m}}} \right.} \right) - \log {h_{\omega ,m}}} \right]} 
\end{array}\]
where $\sum\limits_{m \in {\cal M}} {{h_{\omega ,m}}}  = 1$ . Lagrange multiplier ${\lambda _{{G_2}}}$ is introduced again to solve for ${h_{\omega ,m}}$, the corresponding Lagrange function is as follows:
\[\begin{array}{l}
{G_2} = \sum\limits_{m \in {\cal M}} {{h_{\omega ,m}}\left[ {\log {w_{\omega ,m}}g\left( {{\bf{x}}\left| {\omega ,{{\bf{\mu }}_{\omega ,m}},{{\bf{\Sigma }}_{\omega ,m}}} \right.} \right) - \log {h_{\omega ,m}}} \right]} \\
\;\;\;\;\;\; + {\lambda _{{G_2}}}\left( {1 - \sum\limits_{m \in {\cal M}} {{h_{\omega ,m}}} } \right)
\end{array}\]
By setting the derivative w.r.t  ${h_{\omega ,m}}$ equals to zero, we obtain the formula as follow:
\[\begin{array}{l}
\log {w_{\omega ,m}}g\left( {{\bf{x}}\left| {\omega ,{{\bf{\mu }}_{\omega ,m}},{{\bf{\Sigma }}_{\omega ,m}}} \right.} \right) - \log {h_{\omega ,m}} - 1 - {\lambda _{{\lambda _{{G_2}}}}} = 0\\
 \Rightarrow \log {h_{\omega ,m}} = \log {w_{\omega ,m}}g\left( {{\bf{x}}\left| {\omega ,{{\bf{\mu }}_{\omega ,m}},{{\bf{\Sigma }}_{\omega ,m}}} \right.} \right) - 1 - {\lambda _{{\lambda _{{G_2}}}}}\\
 \Rightarrow {h_{\omega ,m}} = {w_{\omega ,m}}g\left( {{\bf{x}}\left| {\omega ,{{\bf{\mu }}_{\omega ,m}},{{\bf{\Sigma }}_{\omega ,m}}} \right.} \right) \cdot {e^{ - \left( {1 + {\lambda _{{\lambda _{{G_2}}}}}} \right)}}
\end{array}\]
Furthermore, we integrate ${h_{\omega ,m}}$  and the optimal solution w.r.t  ${\lambda _{{G_2}}}$ is: 
\[\begin{array}{l}
\sum\limits_{m \in {\cal M}} {{h_{\omega ,m}}}  = \sum\limits_{m \in {\cal M}} {{w_{\omega ,m}}g\left( {{\bf{x}}\left| {\omega ,{{\bf{\mu }}_{\omega ,m}},{{\bf{\Sigma }}_{\omega ,m}}} \right.} \right) \cdot {e^{ - \left( {1 + {\lambda _{{G_2}}}} \right)}}} \\
 \Rightarrow {\lambda _{{G_2}}} = \log \sum\limits_{m \in {\cal M}} {{w_{\omega ,m}}g\left( {{\bf{x}}\left| {\omega ,{{\bf{\mu }}_{\omega ,m}},{{\bf{\Sigma }}_{\omega ,m}}} \right.} \right)}  - 1
\end{array}\]
We plug ${\lambda _{{G_2}}}$  back and solve for ${h_{\omega ,m}}$  as follows:
\begin{equation}%%%公式6
\begin{array}{c}
{h_{\omega ,m}} = \frac{{{w_{\omega ,m}}g\left( {{\bf{x}}\left| {\omega ,{{\bf{\mu }}_{\omega ,m}},{{\bf{\Sigma }}_{\omega ,m}}} \right.} \right)}}{{\sum\limits_{m \in {\cal M}} {{w_{\omega ,m}}g\left( {{\bf{x}}\left| {\omega ,{{\bf{\mu }}_{\omega ,m}},{{\bf{\Sigma }}_{\omega ,m}}} \right.} \right)} }}\\
 = \Pr \left( {m\left| {{\bf{x}},\omega } \right.} \right)
\end{array}
\end{equation}
As a result, equation (5) and (6) are plugged back to derive the objective function (7):
\begin{equation}%%%公式7
	\begin{array}{c}
Q = \sum\limits_{\tilde \omega  \in \Omega } {I\left( {\tilde \omega  = \omega } \right)\sum\limits_{{\bf{x}} \in {\cal S}} {\sum\limits_{\omega  \in \Omega } {\left\{ {p\left( {\omega \left| {{\bf{x}},\tilde \omega } \right.} \right)} \right.} } } \\
\left. { \cdot \sum\limits_{m \in {\cal M}} {\Pr \left( {m\left| {{\bf{x}},\omega } \right.} \right)\log \frac{{{w_{\omega ,m}}g\left( {{\bf{x}}\left| {\omega ,{{\bf{\mu }}_{\omega ,m}},{{\bf{\Sigma }}_{\omega ,m}}} \right.} \right)}}{{\Pr \left( {m\left| {{\bf{x}},\omega } \right.} \right)}}} } \right\}\\
 + \sum\limits_{{\bf{x}} \in {\cal S}} {\sum\limits_{\omega  \in \Omega } {p\left( {\omega \left| {{\bf{x}},\tilde \omega } \right.} \right)\log {\gamma _{\tilde \omega ,\omega }}} }  + \sum\limits_{{\bf{x}} \in {\cal S}} {\sum\limits_{\omega  \in \Omega } {p\left( {\omega \left| {{\bf{x}},\tilde \omega } \right.} \right)\log {\pi _\omega }} } 
\end{array}
\end{equation}
The parameters to be estimated are $\Theta  = {\left\{ {{\theta _\omega }} \right\}_{\omega  \in \Omega }}$, $\Gamma  = {\left\{ {{\gamma _{\tilde \omega ,\omega }}} \right\}_{\tilde \omega  \in \Omega }}$ , and $\Pi  = {\left\{ {{\pi _\omega }} \right\}_{\omega  \in \Omega }}$ , where ${\theta _\omega } = {\left\{ {{w_{\omega ,m}},{{\bf{\mu }}_{\omega ,m}},{{\bf{\Sigma }}_{\omega ,m}}} \right\}_{m \in {\cal M}}}$ . The EM method is employed to solve for the parameters.

\textbf{E step:} $p\left( {\omega \left| {{\bf{x}},\tilde \omega } \right.} \right)$  and  $\Pr \left( {m\left| {{\bf{x}},\omega } \right.} \right)$ are calculated according to equations (5) and (6), which are listed as follows:
\begin{equation}%%%公式8
	p\left( {\omega \left| {{\bf{x}},\tilde \omega } \right.} \right) = \frac{{p\left( {{\bf{x}}\left| {\omega ,{\theta _\omega }} \right.} \right){\gamma _{\tilde \omega ,\omega }}{\pi _\omega }}}{{\sum\limits_{\omega  \in \Omega } {p\left( {{\bf{x}}|\omega ,{\theta _\omega }} \right){\gamma _{\tilde \omega ,\omega }}{\pi _\omega }} }}
\end{equation}
\begin{equation}%%%公式9
	\Pr \left( {m\left| {{{\bf{x}}_n},{\omega _n} = k} \right.} \right) = \frac{{{w_{\omega ,m}}g\left( {{\bf{x}}\left| {\omega ,{{\bf{\mu }}_{\omega ,m}},{{\bf{\Sigma }}_{\omega ,m}}} \right.} \right)}}{{\sum\limits_{m \in {\cal M}} {{w_{\omega ,m}}g\left( {{\bf{x}}\left| {\omega ,{{\bf{\mu }}_{\omega ,m}},{{\bf{\Sigma }}_{\omega ,m}}} \right.} \right)} }}
\end{equation}

\textbf{M step:} (7) is optimized to solve for the local optimum of $\Theta  = {\left\{ {{\theta _\omega }} \right\}_{\omega  \in \Omega }}$ ,$\Gamma  = {\left\{ {{\gamma _{\tilde \omega ,\omega }}} \right\}_{\tilde \omega  \in \Omega }}$  , and $\Pi  = {\left\{ {{\pi _\omega }} \right\}_{\omega  \in \Omega }}$ .
\subsection{Updating}%2.3
\textbf{Updating rule of ${{\bf{\mu }}_{\omega ,m}}$ :} by setting the derivative of equation (7) w.r.t ${{\bf{\mu }}_{\omega ,m}}$  equals to zero, we derive the updating rule as follow:
\begin{equation}%%%公式10
	\begin{array}{l}
\frac{{\partial Q}}{{\partial {{\bf{\mu }}_{\omega ,m}}}} = \sum\limits_{\tilde \omega  \in \Omega } {I\left( {\tilde \omega  = \omega } \right)\sum\limits_{{\bf{x}} \in {\cal S}} {\left\{ {p\left( {\omega \left| {{\bf{x}},\tilde \omega } \right.} \right)} \right.} } \\
\left. {\;\;\;\;\;\;\;\;\;\; \cdot \Pr \left( {m|{\bf{x}},\omega } \right) \cdot {\bf{\Sigma }}_{\omega ,m}^{ - 1} \cdot \left( {{\bf{x}} - {{\bf{\mu }}_{\omega ,m}}} \right)} \right\} = 0\\
 \Rightarrow {{\bf{\mu }}_{\omega ,m}} = \frac{{\sum\limits_{\tilde \omega  \in \Omega } {I\left( {\tilde \omega  = \omega } \right)\sum\limits_{{\bf{x}} \in {\cal S}} {p\left( {\omega \left| {{\bf{x}},\tilde \omega } \right.} \right)\Pr \left( {m|{\bf{x}},\omega } \right){\bf{x}}} } }}{{\sum\limits_{\tilde \omega  \in \Omega } {I\left( {\tilde \omega  = \omega } \right)\sum\limits_{{\bf{x}} \in {\cal S}} {p\left( {\omega \left| {{\bf{x}},\tilde \omega } \right.} \right)\Pr \left( {m|{\bf{x}},\omega } \right)} } }}
\end{array}
\end{equation}

\textbf{Updating rule of ${{\bf{\Sigma }}_{\omega ,m}}$ :} by setting the derivative of equation (7) w.r.t  ${{\bf{\Sigma }}_{\omega ,m}}$ equals to zero, we derive the updating rule as follow:
\begin{equation}%%%公式11
	\begin{array}{l}
\frac{{\partial Q}}{{\partial {{\bf{\Sigma }}_{\omega ,m}}}} =  - \sum\limits_{\tilde \omega  \in \Omega } {I\left( {\tilde \omega  = \omega } \right)\sum\limits_{{\bf{x}} \in {\cal S}} {\left\{ {p\left( {\omega \left| {{\bf{x}},\tilde \omega } \right.} \right)\Pr \left( {m\left| {{\bf{x}},\omega } \right.} \right) \cdot \left[ {{{\bf{\Sigma }}_{\omega ,m}}^{ - 1}} \right.} \right.} } \\
\;\;\;\;\;\;\;\;\;\;\;\left. {\left. { - {{\bf{\Sigma }}_{\omega ,m}}^{ - 1}\left( {{\bf{x}} - {{\bf{\mu }}_{\omega ,m}}} \right){{\left( {{\bf{x}} - {{\bf{\mu }}_{\omega ,m}}} \right)}^T}{{\bf{\Sigma }}_{\omega ,m}}^{ - 1}} \right]} \right\} = 0\\
 \Rightarrow {{\bf{\Sigma }}_{\omega ,m}}\\
 = \frac{{\sum\limits_{\tilde \omega  \in \Omega } {I\left( {\tilde \omega  = \omega } \right)\sum\limits_{{\bf{x}} \in {\cal S}} {p\left( {\omega \left| {{\bf{x}},\tilde \omega } \right.} \right)\Pr \left( {m|{\bf{x}},\omega } \right)\left( {{\bf{x}} - {{\bf{\mu }}_{\omega ,m}}} \right){{\left( {{\bf{x}} - {{\bf{\mu }}_{\omega ,m}}} \right)}^T}} } }}{{\sum\limits_{\tilde \omega  \in \Omega } {I\left( {\tilde \omega  = \omega } \right)\sum\limits_{{\bf{x}} \in {\cal S}} {p\left( {\omega \left| {{\bf{x}},\tilde \omega } \right.} \right)\Pr \left( {m|{\bf{x}},\omega } \right)} } }}
\end{array}
\end{equation}
\textbf{Updating rule of ${w_{\omega ,m}}$ :} a Lagrange multiplier  ${\lambda _{{w_{\omega ,m}}}}$ is introduced to guarantee the constraint condition $\sum\limits_{m \in {\cal M}} {{w_{\omega ,m}}}  = 1$, the corresponding Lagrange function is designed as follows:
\[{Q_{{\lambda _{{w_{\omega ,m}}}}}} = Q + {\lambda _{{w_{\omega ,m}}}}\left( {1 - \sum\limits_{m \in {\cal M}} {{w_{\omega ,m}}} } \right)\]
and by setting the derivative of  ${Q_{{\lambda _{{w_{\omega ,m}}}}}}$ w.r.t ${w_{\omega ,m}}$  equals to zero, we derive the formula as follow:
\[\begin{array}{l}
\frac{{\partial {Q_{{\lambda _{{w_{\omega ,m}}}}}}}}{{\partial {w_{\omega ,m}}}} = 0\\
 = \sum\limits_{\tilde \omega  \in \Omega } {I\left( {\tilde \omega  = \omega } \right)\sum\limits_{{\bf{x}} \in {\cal S}} {p\left( {\omega \left| {{\bf{x}},\tilde \omega } \right.} \right)\Pr \left( {m|{\bf{x}},\omega } \right)\frac{1}{{{w_{\omega ,m}}}}} }  - {\lambda _{{w_{\omega ,m}}}}\\
 \Rightarrow {\lambda _{{w_{\omega ,m}}}}{w_{\omega ,m}} = \sum\limits_{\tilde \omega  \in \Omega } {I\left( {\tilde \omega  = \omega } \right)\sum\limits_{{\bf{x}} \in {\cal S}} {p\left( {\omega \left| {{\bf{x}},\tilde \omega } \right.} \right)\Pr \left( {m|{\bf{x}},\omega } \right)} } \\
\mathop  \Rightarrow \limits^{{\rm{intergrate }}{w_{\omega ,m}}} {\lambda _{{w_{\omega ,m}}}}\sum\limits_{m \in {\cal M}} {{w_{\omega ,m}}}  = {\lambda _{{w_{\omega ,m}}}}\\
 = \sum\limits_{\tilde \omega  \in \Omega } {I\left( {\tilde \omega  = \omega } \right)\sum\limits_{{\bf{x}} \in {\cal S}} {p\left( {\omega \left| {{\bf{x}},\tilde \omega } \right.} \right)\sum\limits_{m \in {\cal M}} {\Pr \left( {m|{\bf{x}},\omega } \right)} } } 
\end{array}\]
and we plug ${\lambda _{{w_{\omega ,m}}}}$  back and solve for ${w_{\omega ,m}}$  as follows:
\begin{equation}%%%公式12
	{w_{\omega ,m}} = \frac{{\sum\limits_{\tilde \omega  \in \Omega } {I\left( {\tilde \omega  = \omega } \right)\sum\limits_{{\bf{x}} \in {\cal S}} {p\left( {\omega \left| {{\bf{x}},\tilde \omega } \right.} \right)\Pr \left( {m|{\bf{x}},\omega } \right)} } }}{{\sum\limits_{\tilde \omega  \in \Omega } {I\left( {\tilde \omega  = \omega } \right)\sum\limits_{{\bf{x}} \in {\cal S}} {p\left( {\omega \left| {{\bf{x}},\tilde \omega } \right.} \right)\sum\limits_{m \in {\cal M}} {\Pr \left( {m|{\bf{x}},\omega } \right)} } } }}
\end{equation}
\textbf{Updating the rule of ${\gamma _{\tilde \omega ,\omega }}$ :} a Lagrange multiplier ${\lambda _{{\gamma _{\tilde \omega ,\omega }}}}$  is introduced to guarantee the constraint condition , and the corresponding Lagrange function is designed as follows: 
\[\begin{array}{l}
\frac{{\partial {Q_{{\gamma _{\tilde \omega ,\omega }}}}}}{{\partial {\gamma _{\tilde \omega ,\omega }}}} = \sum\limits_{{\bf{x}} \in {\cal S}} {p\left( {\omega \left| {{\bf{x}},\tilde \omega } \right.} \right)} \frac{1}{{{\gamma _{\tilde \omega ,\omega }}}} - {\lambda _{{\gamma _{\tilde \omega ,\omega }}}} = 0\\
 \Rightarrow {\lambda _{{\gamma _{\tilde \omega ,\omega }}}}{\gamma _{\tilde \omega ,\omega }} = \sum\limits_{{\bf{x}} \in {\cal S}} {p\left( {\omega \left| {{\bf{x}},\tilde \omega } \right.} \right)} \\
\mathop  \Rightarrow \limits^{{\rm{intergrate }}{\gamma _{\tilde \omega ,\omega }}} {\lambda _{{\gamma _{\tilde \omega ,\omega }}}}\sum\limits_{\tilde \omega  \in \Omega } {{\gamma _{\tilde \omega ,\omega }}}  = {\lambda _{{\gamma _{\tilde \omega ,\omega }}}}\\
 = \sum\limits_{{\bf{x}} \in {\cal S}} {\sum\limits_{\tilde \omega  \in \Omega } {p\left( {\omega \left| {{\bf{x}},\tilde \omega } \right.} \right)} } 
\end{array}\]
and we plug ${\lambda _{{\gamma _{\tilde \omega ,\omega }}}}$  back and solve for ${\gamma _{\tilde \omega ,\omega }}$  as follows:
\begin{equation}%%%公式13
	{\gamma _{\tilde \omega ,\omega }} = \frac{{\sum\limits_{{\bf{x}} \in {\cal S}} {p\left( {\omega \left| {{\bf{x}},\tilde \omega } \right.} \right)} }}{{\sum\limits_{{\bf{x}} \in {\cal S}} {\sum\limits_{\tilde \omega  \in \Omega } {p\left( {\omega \left| {{\bf{x}},\tilde \omega } \right.} \right)} } }}
\end{equation}
\textbf{Updating class probability:} a Lagrange multiplier  ${\lambda _{{\pi _\omega }}}$ is introduced to guarantee the constraint condition $\sum\limits_{\omega  \in \Omega } {{\pi _\omega }}  = 1$ , a Lagrange function is designed as follows:
\[{Q_{{\lambda _{{\pi _\omega }}}}} = Q + {\lambda _{{\pi _\omega }}}\left( {1 - \sum\limits_{\omega  \in \Omega } {{\pi _\omega }} } \right)\]
and by setting the derivative of  ${Q_{{\lambda _{{\pi _\omega }}}}}$ w.r.t ${\pi _\omega }$  equals to zero, we derive the formula as follow:
\[\begin{array}{l}
\frac{{\partial {Q_{{\lambda _{{\pi _\omega }}}}}}}{{\partial {\pi _\omega }}}{\rm{ = }}\sum\limits_{\tilde \omega  \in \Omega } {I\left( {\tilde \omega  = \omega } \right)\sum\limits_{{\bf{x}} \in {\cal S}} {p\left( {\omega \left| {{\bf{x}},\tilde \omega } \right.} \right)\frac{1}{{{\pi _\omega }}}} }  - {\lambda _{{\pi _\omega }}}{\rm{ = 0}}\\
 \Rightarrow {\lambda _{{\pi _\omega }}}{\pi _\omega }{\rm{ = }}\sum\limits_{\tilde \omega  \in \Omega } {I\left( {\tilde \omega  = \omega } \right)\sum\limits_{{\bf{x}} \in {\cal S}} {p\left( {\omega \left| {{\bf{x}},\tilde \omega } \right.} \right)} } \\
\mathop  \Rightarrow \limits^{{\rm{intergrate }}{\pi _\omega }} {\lambda _{{\pi _\omega }}}{\rm{ = }}\sum\limits_{\tilde \omega  \in \Omega } {I\left( {\tilde \omega  = \omega } \right)\sum\limits_{{\bf{x}} \in {\cal S}} {\sum\limits_{\omega  \in \Omega } {p\left( {\omega \left| {{\bf{x}},\tilde \omega } \right.} \right)} } } 
\end{array}\]
and we plug  ${\lambda _{{\pi _\omega }}}$ back and solve for ${\pi _\omega }$  as follows:
\begin{equation}%%%公式14
	{\pi _\omega } = \frac{{\sum\limits_{\tilde \omega  \in \Omega } {I\left( {\tilde \omega  = \omega } \right)\sum\limits_{{\bf{x}} \in {\cal S}} {p\left( {\omega \left| {{\bf{x}},\tilde \omega } \right.} \right)} } }}{{\sum\limits_{\tilde \omega  \in \Omega } {I\left( {\tilde \omega  = \omega } \right)\sum\limits_{{\bf{x}} \in {\cal S}} {\sum\limits_{\omega  \in \Omega } {p\left( {\omega \left| {{\bf{x}},\tilde \omega } \right.} \right)} } } }}
\end{equation}
\textbf{Predicting posterior probability:} to predict an unlabeled instance ${{\bf{x}}_u}$, the posterior probability of each class is calculated as follows:
\[\begin{array}{c}
p\left( {\omega \left| {{{\bf{x}}_u}} \right.} \right) = \frac{{p\left( {{{\bf{x}}_u}\left| {\omega ,{\theta _\omega }} \right.} \right){\pi _\omega }}}{{\sum\limits_{\omega  \in \Omega } {p\left( {{{\bf{x}}_u}\left| {\omega ,{\theta _\omega }} \right.} \right){\pi _\omega }} }}\\
 = \frac{{{\pi _\omega }\sum\limits_{m \in {\cal M}} {{w_{\omega ,m}}g\left( {{\bf{x}}\left| {\omega ,{{\bf{\mu }}_{\omega ,m}},{{\bf{\Sigma }}_{\omega ,m}}} \right.} \right)} }}{{\sum\limits_{\omega  \in \Omega } {{\pi _\omega }\sum\limits_{m \in {\cal M}} {{w_{\omega ,m}}g\left( {{\bf{x}}\left| {\omega ,{{\bf{\mu }}_{\omega ,m}},{{\bf{\Sigma }}_{\omega ,m}}} \right.} \right)} } }}
\end{array}\]
The maximum class posterior probability decides the class of this unlabeled instance.

For convenience, we summarize the overall updating process in algorithm 1.
\begin{algorithm}%%%算法
	\caption{GMDA}
	\begin{algorithmic}
		\STATE \textbf{input:}$\Theta  = {\left\{ {{\theta _\omega }} \right\}_{\omega  \in \Omega }}$,$\Gamma  = {\left\{ {{\gamma _{\tilde \omega ,\omega }}} \right\}_{\tilde \omega  \in \Omega }}$ , and $\Pi  = {\left\{ {{\pi _\omega }} \right\}_{\omega  \in \Omega }}$ .
		\STATE \textbf{Initialize:} ${w_{\omega ,m}},{{\bf{\mu }}_{\omega ,m}},{{\bf{\Sigma }}_{\omega ,m}}$ is obtained by employing k-means method, and the initialization value of ${\pi _\omega }$  is:
		\[{\pi _\omega } = \frac{{\sum\limits_{\omega  \in \Omega } {I\left( {\tilde \omega } \right)} }}{{\left| \Omega  \right|}}\]
		\IF{iter < itermax}
			\STATE $p\left( {\omega  = k\left| {{{\bf{x}}_{n,}},\tilde \omega  = j} \right.} \right)$ and $\Pr \left( {m\left| {{{\bf{x}}_n},{\omega _n} = k} \right.} \right)$ are calculated according to (8) and (9).

		\STATE ${{\bf{\mu }}_{\omega ,m}},{{\bf{\Sigma }}_{\omega ,m}},{w_{\omega ,m}},{\gamma _{\tilde \omega ,\omega }},{\pi _\omega }$ is updated according to (10)–(14).
		\ENDIF
		\STATE \textbf{end}
		\STATE Output: $\Theta $ ,$\Gamma $ , and $\Pi $.
	\end{algorithmic}
\end{algorithm}
\section{CONVERGENCE ANALYSIS}%3
\subsection{General Solution}%3.1
The log-likelihood function for the general decision problem with randomly substituted labels is given by
\begin{equation}%%%公式15
	\begin{array}{l}
{L_0} = \frac{1}{{\left| {\cal S} \right|}}\sum\limits_{{\bf{x}} \in {\cal S}} {log} \;\tilde p\left( {{\bf{x}}|\psi \left( {\bf{x}} \right)} \right)\pi \left( {\psi \left( {\bf{x}} \right)} \right)\\
 = \frac{1}{{\left| {\cal S} \right|}}\sum\limits_{{\bf{x}} \in {\cal S}} {log} \sum\limits_{\omega  \in \Omega } {\left[ {p\left( {{\bf{x}}|\omega } \right)p\left( {\omega |\psi \left( {\bf{x}} \right)} \right)\pi \left( {\psi \left( {\bf{x}} \right)} \right)} \right]} 
\end{array}
\end{equation}
If we denote ${{\cal S}_{\tilde \omega }}$  the training set of data vectors with the observed label  $\tilde \omega $:
\[{{\cal S}_{\tilde \omega }} = \left\{ {{\bf{x}} \in {\cal S}:\psi \left( x \right) = \tilde \omega } \right\},\;\tilde \omega  \in \Omega ,{\cal S} = \bigcup\limits_{\tilde \omega  \in \Omega } {{{\cal S}_{\tilde \omega }},\left| {\cal S} \right| = \sum\limits_{\tilde \omega  \in \Omega } {\left| {{{\cal S}_{\tilde \omega }}} \right|} } \]
then, using the relation $\tilde \omega  = \psi \left( {\bf{x}} \right)$ , we can express the log-likelihood function (15) equivalently in the form
\[{L_0} = \frac{1}{{\left| {\cal S} \right|}}\sum\limits_{\tilde \omega  \in \Omega } {\sum\limits_{{\bf{x}} \in {{\cal S}_{\tilde \omega }}} {\left[ {log\left( {\sum\limits_{\omega  \in \Omega } {p\left( {{\bf{x}}|\omega } \right)p\left( {\omega |\tilde \omega } \right)} } \right) + log\pi \left( {\tilde \omega } \right)} \right]} } \;\]
and further
\begin{equation}%%%公式16
	{L_0} = \sum\limits_{\tilde \omega  \in \Omega } {\frac{{\left| {{{\cal S}_{\tilde \omega }}} \right|}}{{\left| {\cal S} \right|}}\,} log\pi (\tilde \omega )\, + \,\frac{1}{{\left| {\cal S} \right|}}\sum\limits_{{\bf{x}} \in {\cal S}} {log\left[ {p({\bf{x}}|\omega )p(\omega |\psi ({\bf{x}}))} \right]} \;
\end{equation}
We recall that for any two probability distributions $p\left( \omega  \right)$ , $\pi (\omega )$  the well-known Kullback-Leibler information divergence is non-negative
\[I\left( {p,\pi } \right) = \sum\limits_{\omega  \in \Omega } {p\left( \omega  \right)\;} log\;\frac{{p\left( \omega  \right)}}{{\pi \left( \omega  \right)}}\, \ge \,0\]
and the inequality (11) can be rewritten in the form
\begin{equation}%%%公式17
	\sum\limits_{\omega  \in \Omega } {p\left( \omega  \right)} \;log\pi \left( \omega  \right)\; \le \;\sum\limits_{\omega  \in \Omega } {p\left( \omega  \right)} \;logp\left( \omega  \right)\;
\end{equation}
Consequently, the left-hand part of the inequality (17) is maximized by setting $\pi (\omega ){\rm{ = }}p\left( \omega  \right)$,$\omega  \in \Omega $  . For the same reason the first sum in (16) is maximized by the maximum likelihood estimate  $\pi (\tilde \omega ){\rm{ = }}\left| {{{\cal S}_{\tilde \omega }}} \right|/\left| {\cal S} \right|$, $\tilde \omega  \in \Omega $   in terms of relative frequencies. In the following we repeatedly make use of this practically useful consequence of the inequality (17).

The second term in  ${L_0}$ can be maximized by EM algorithm. If we define the conditional weights
\begin{equation}%%%公式18
	\begin{array}{l}
q\left( {\omega |{\bf{x}},\psi \left( {\bf{x}} \right)} \right) = \frac{{p\left( {{\bf{x}}|\omega } \right)p\left( {\omega |\psi \left( {\bf{x}} \right)} \right)}}{{\sum\limits_{\omega  \in \Omega } {p\left( {{\bf{x}}|\omega } \right)p\left( {\omega |\psi \left( {\bf{x}} \right)} \right)} }}\\
\omega  \in \Omega ,\sum\limits_{\omega  \in \Omega } {q\left( {\omega |{\bf{x}},\psi \left( {\bf{x}} \right)} \right) = 1} , {\bf{x}} \in \;{\cal S}
\end{array}
\end{equation}
then the second part of the log-likelihood function (16):
\begin{equation}%%%公式19
	L = \frac{1}{{\left| {\cal S} \right|}}\sum\limits_{{\bf{x}} \in {\cal S}} {log} \left[ {\sum\limits_{\omega  \in \Omega } {p\left( {{\bf{x}}|\omega } \right)p\left( {\omega |\psi ({\bf{x}})} \right)} } \right]
\end{equation}
can be expanded in the form [46, 47]:
\begin{equation}%%%公式20
	\begin{array}{l}
L = \sum\limits_{\omega  \in \Omega } {\frac{1}{{\left| {\cal S} \right|}}} \sum\limits_{{\bf{x}} \in {\cal S}} \{  q\left( {\omega |{\bf{x}},\psi \left( {\bf{x}} \right)} \right)log[p\left( {{\bf{x}}|\omega } \right)p\left( {\omega |\psi ({\bf{x}})} \right)]\\
 - \;q\left( {\omega |{\bf{x}},\psi \left( {\bf{x}} \right)} \right)logq\left( {\omega |{\bf{x}},\psi \left( {\bf{x}} \right)} \right)\} 
\end{array}
\end{equation}
Similarly, denoting by apostrophe the mixture components and mixture parameters in the next iteration of EM algorithm, we can write
\begin{equation}%%%公式21
	\begin{array}{l}
L' = \frac{1}{{\left| {\cal S} \right|}}\sum\limits_{{\bf{x}} \in {\cal S}} {log} \left[ {\sum\limits_{\omega  \in \Omega } {p'\left( {{\bf{x}}|\omega } \right)p'\left( {\omega |\psi ({\bf{x}})} \right)} } \right]\\
 = \sum\limits_{\omega  \in \Omega } {\frac{1}{{\left| {\cal S} \right|}}} \sum\limits_{{\bf{x}} \in {\cal S}} {{\rm{\{ }}q\left( {\omega |{\bf{x}},\psi \left( {\bf{x}} \right)} \right)log[p'\left( {{\bf{x}}|\omega } \right)p'\left( {\omega |\psi ({\bf{x}})} \right)]} \;\\
 - \;q\left( {\omega |{\bf{x}},\psi \left( {\bf{x}} \right)} \right)logq'\left( {\omega |{\bf{x}},\psi \left( {\bf{x}} \right)} \right){\rm{\} }}
\end{array}
\end{equation}
where
\begin{equation}%%%公式22
	\begin{array}{l}
q'\left( {\omega |{\bf{x}},\psi \left( {\bf{x}} \right)} \right) = \frac{{p'\left( {{\bf{x}}|\omega } \right)p'\left( {\omega |\psi \left( {\bf{x}} \right)} \right)}}{{\sum\limits_{\omega  \in \Omega } {p'\left( {{\bf{x}}|\omega } \right)p'\left( {\omega |\psi \left( {\bf{x}} \right)} \right)} }},\\
\omega  \in \Omega ,{\bf{x}} \in \;{\cal S}
\end{array}
\end{equation}
Now, the increment of EM algorithm in one iteration can be expressed as follows
\begin{equation}%%%公式23
	\begin{array}{l}
L' - L\\
{\rm{ = }}\sum\limits_{\omega  \in \Omega } {\frac{1}{{\left| {\cal S} \right|}}} \sum\limits_{{\bf{x}} \in {\cal S}} {q\left( {\omega |{\bf{x}},\psi \left( {\bf{x}} \right)} \right)} log\left[ {\frac{{p'\left( {{\bf{x}}|\omega } \right)p'\left( {\omega |\psi ({\bf{x}})} \right)}}{{p\left( {{\bf{x}}|\omega } \right)p\left( {\omega |\psi ({\bf{x}})} \right)}}} \right]\\
 + \frac{1}{{\left| {\cal S} \right|}}\sum\limits_{{\bf{x}} \in {\cal S}} {\sum\limits_{\omega  \in \Omega } {q\left( {\omega |{\bf{x}},\psi \left( {\bf{x}} \right)} \right)log\frac{{q\left( {\omega |x,\psi \left( x \right)} \right)}}{{q'\left( {\omega |x,\psi \left( x \right)} \right)}}} } 
\end{array}
\end{equation}
The last sum in Eq. (23) is again the well-known non-negative Kullback-Leibler information divergence:
\begin{equation}%%%公式24
	I\left( {q,q'} \right) = \sum\limits_{\omega  \in \Omega } {q\left( {\omega |{\bf{x}},\psi \left( {\bf{x}} \right)} \right)} log\frac{{q\left( {\omega |{\bf{x}},\psi \left( {\bf{x}} \right)} \right)}}{{q'\left( {\omega |{\bf{x}},\psi \left( {\bf{x}} \right)} \right)}} \ge 0
\end{equation}
and therefore, we can write
\begin{equation}%%%公式25
	\begin{array}{l}
L' - L\\
 \ge \frac{1}{{\left| {\cal S} \right|}}\sum\limits_{{\bf{x}} \in {\cal S}} {\sum\limits_{\omega  \in \Omega } {q\left( {\omega |{\bf{x}},\psi \left( {\bf{x}} \right)} \right)} } log\left[ {\frac{{p'\left( {{\bf{x}}|\omega } \right)p'\left( {\omega |\psi ({\bf{x}})} \right)}}{{p\left( {{\bf{x}}|\omega } \right)p\left( {\omega |\psi ({\bf{x}})} \right)}}} \right]
\end{array}
\end{equation}
Thus, for the sake of the monotonic property of EM algorithm, we have to guarantee the inequality
\begin{equation}%%%公式26
	\begin{array}{l}
L' - L\\
 \ge \sum\limits_{\omega  \in \Omega } {\frac{1}{{\left| {\cal S} \right|}}} \sum\limits_{{\bf{x}} \in {\cal S}} {\left[ {q\left( {\omega |{\bf{x}},\psi \left( {\bf{x}} \right)} \right)log\frac{{p'\left( {{\bf{x}}|\omega } \right)}}{{p\left( {{\bf{x}}|\omega } \right)}}} \right]} \\
{\rm{ + }}\sum\limits_{\omega  \in \Omega } {\frac{1}{{\left| {\cal S} \right|}}} \sum\limits_{{\bf{x}} \in {\cal S}} {\left[ {q\left( {\omega |{\bf{x}},\psi \left( {\bf{x}} \right)} \right)log\frac{{p'\left( {\omega |\psi ({\bf{x}})} \right)}}{{p\left( {\omega |\psi ({\bf{x}})} \right)}}} \right]}  \ge 0
\end{array}
\end{equation}
Here the sum over $x \in {\cal S}$ in the second term can be decomposed into summing over  ${\bf{x}} \in {{\cal S}_{\tilde \omega }}$:
\begin{equation}%%%公式27
	\begin{array}{l}
L' - L\\
{\rm{ = }}\sum\limits_{\omega  \in \Omega } {\frac{1}{{\left| {\cal S} \right|}}} \sum\limits_{{\bf{x}} \in {\cal S}} {\left[ {q\left( {\omega |{\bf{x}},\psi \left( {\bf{x}} \right)} \right)log\frac{{p'\left( {{\bf{x}}|\omega } \right)}}{{p\left( {{\bf{x}}|\omega } \right)}}} \right]} \\
{\rm{ + }}\sum\limits_{\tilde \omega  \in \Omega } {\frac{{\left| {{{\cal S}_{\tilde \omega }}} \right|}}{{\left| {\cal S} \right|}}} \sum\limits_{\omega  \in \Omega } {\left[ {\frac{1}{{\left| {{{\cal S}_{\tilde \omega }}} \right|}}\sum\limits_{{\bf{x}} \in {{\cal S}_{\tilde \omega }}} {q\left( {\omega |{\bf{x}},\tilde \omega } \right)} } \right]} \;log\frac{{p'\left( {\omega |\tilde \omega } \right)}}{{p\left( {\omega |\tilde \omega } \right)}}\;
\end{array}
\end{equation}
Now, if we define the EM iteration equations in the form
\begin{equation}%%%公式28
	p'\left( {\omega |\tilde \omega } \right){\rm{ = }}\frac{1}{{\left| {{{\cal S}_{\tilde \omega }}} \right|}}\sum\limits_{{\bf{x}} \in {{\cal S}_{\tilde \omega }}} {q\left( {\omega |{\bf{x}},\tilde \omega } \right),\;\omega  \in \Omega ,\;} \tilde \omega  \in \Omega 
\end{equation}
\begin{equation}%%%公式29
	\begin{array}{l}
p'\left( { \cdot |\omega } \right){\rm{ = }}arg\;\mathop {max}\limits_{p\left( { \cdot |\omega } \right)} \left\{ {\frac{1}{{\left| {\cal S} \right|}}\sum\limits_{{\bf{x}} \in {\cal S}} {q\left( {\omega |{\bf{x}},\psi \left( {\bf{x}} \right)} \right)logp\left( {{\bf{x}}|\omega } \right)} } \right\}\\
\omega  \in \Omega 
\end{array}
\end{equation}
then the monotonic property of EM algorithm is guaranteed because, by substitution (28), the second term in (26) is nonnegative as a sum of nonnegative Kullback-Leibler information divergences:
\begin{equation}%%%公式30
\sum\limits_{\omega  \in \Omega } {p'\left( {\omega |\tilde \omega } \right)\;} log\frac{{p'\left( {\omega |\tilde \omega } \right)}}{{p\left( {\omega |\tilde \omega } \right)}} \ge 0,\tilde \omega  \in \Omega 
\end{equation}
and the Eq. (29) implies the inequalities
\[\begin{array}{l}
\frac{1}{{\left| {\cal S} \right|}}\sum\limits_{{\bf{x}} \in {\cal S}} {q\left( {\omega |{\bf{x}},\psi \left( {\bf{x}} \right)} \right)} \;log\;p'\left( {{\bf{x}}|\omega } \right)\\
 \ge \frac{1}{{\left| {\cal S} \right|}}\sum\limits_{{\bf{x}} \in {\cal S}} {q\left( {\omega |{\bf{x}},\psi \left( {\bf{x}} \right)} \right)} \;log\;p\left( {{\bf{x}}|\omega } \right),\omega  \in \Omega 
\end{array}\]
which can be rewritten in the form:
\begin{equation}%%%公式31
	\frac{1}{{\left| {\cal S} \right|}}\sum\limits_{{\bf{x}} \in {\cal S}} {q\left( {\omega |{\bf{x}},\psi \left( {\bf{x}} \right)} \right)log\frac{{p'\left( {{\bf{x}}|\omega } \right)}}{{p\left( {{\bf{x}}|\omega } \right)}}}  \ge 0,\omega  \in \Omega 
\end{equation}
Consequently, the first term in (27) is nonnegative and in view of the above inequalities (30), (31), the EM iteration equations in the general form (23), (28) and (29) imply the basic monotonic property of EM algorithm \cite{46dempster1977maximum,471982On}.
\subsection{Gaussian Classes with Noisy Labels}%3.2
Assuming a particular type, e.g. Gaussian class-conditional densities
\begin{equation}%%%公式32
	p({\bf{x}}|\omega ) = f({\bf{x}}|{{\bf{\mu }}_\omega },{\Sigma _\omega }),\;\omega  \in \Omega 
\end{equation}
we can write Eq. (29) in a more specific form
\begin{equation}%%%公式33
	\begin{array}{l}
\left\{ {{\bf{\mu }}_\omega ^\prime,\Sigma _\omega ^\prime} \right\}{\rm{ = }}\\
\arg \mathop {\max }\limits_{\left\{ {{\mu _\omega },{\Sigma _\omega }} \right\}} \left[ {\frac{1}{{\left| {\cal S} \right|}}\sum\limits_{{\bf{x}} \in {\cal S}} {q\left( {\omega |{\bf{x}},\psi \left( {\bf{x}} \right)} \right)} \;log\;f({\bf{x}}|{{\bf{\mu }}_\omega },{\Sigma _\omega })} \right]\\
\omega  \in \Omega 
\end{array}
\end{equation}
As the maximized expression in Eq. (33) is a weighted likelihood function, we can easily verify \cite{471982On} that the explicit solution can be expressed as a weighted analogy of the standard maximum likelihood estimate. In particular, let $F({\bf{x}}|\mu )$ be a probability density with a parameter $\mu $ having a standard maximum likelihood estimate  $\hat \mu $ :
\begin{equation}%%%公式34
	{L_{\bf{\mu }}} = \frac{1}{{\left| {\cal S} \right|}}\sum\limits_{{\bf{x}} \in {\cal S}} {\;log\;} f({\bf{x}}|{\bf{\mu }}) \to \max  \Rightarrow \;{\bf{\hat \mu }} = \frac{1}{{\left| {\cal S} \right|}}\sum\limits_{{\bf{x}} \in {\cal S}} {\bf{x}} 
\end{equation}
If $N({\bf{x}})$  is the number of repeated occurrences of ${\bf{x}} \in {\cal X}$  in ${\cal S}$  and $q({\bf{x}})$  denotes the relative frequency of  ${\bf{x}} \in {\cal S}$:
\[q({\bf{x}}) = \frac{{N({\bf{x}})}}{{\left| {\cal S} \right|}},\sum\limits_{{\bf{x}} \in {\cal X}} {q({\bf{x}}) = 1,({\bf{x}} \notin {\cal S} \Rightarrow q({\bf{x}}) = 0)} \]
then the Eq. (34) can be rewritten equivalently in the form
\begin{equation}%%%公式35
	{L_{\bf{\mu }}} = \sum\limits_{{\bf{x}} \in {\cal X}} {q({\bf{x}})\;\log \;} F({\bf{x}}|{\bf{\mu }}) \to \max  \Rightarrow {\bf{\hat \mu }} = \sum\limits_{{\bf{x}} \in {\cal X}} {q({\bf{x}})} \;{\bf{x}}
\end{equation}
From the comparison of Eq. (34) and (35) it follows that the weighted likelihood (35) is maximized by the corresponding weighted maximum likelihood estimate (for a detailed proof in [47]). Consequently, in view of (33), we can write:
\begin{equation}%%%公式36
	\begin{array}{l}
{\bf{\mu }}{'_\omega }{\rm{ = }}\frac{1}{{\sum\nolimits_{{\bf{x}} \in {\cal S}} {q\left( {\omega |{\bf{x}},\psi \left( {\bf{x}} \right)} \right)} }}\sum\limits_{{\bf{x}} \in {\cal S}} {q\left( {\omega |{\bf{x}},\psi \left( {\bf{x}} \right)} \right)} \;{\bf{x}}\\
\omega  \in \Omega 
\end{array}
\end{equation}
\begin{equation}%%%公式37
	\begin{array}{l}
\Sigma {'_\omega }{\rm{ = }}\frac{1}{{\sum\nolimits_{{\bf{x}} \in {\cal S}} {q\left( {\omega |{\bf{x}},\psi \left( {\bf{x}} \right)} \right)} }}\sum\limits_{{\bf{x}} \in {\cal S}} {q\left( {\omega |{\bf{x}},\psi \left( x \right)} \right)} \;{\bf{x}}{{\bf{x}}^T}\\
\;\;\;\;\;\;\; - {{\mu '}_\omega }{{\mu '}_\omega }^T,\;\;\omega  \in \Omega 
\end{array}
\end{equation}

We can conclude that the problem of parameter estimation for the Gaussian classes with noisy labels can be solved by repeating the EM iteration equations (18), (28), (36) and (37).
\subsection{Class-conditional Gaussian Mixtures with Noisy Labels}%3.3
The Gaussian assumption (32) is well known to be rather restrictive and can be essentially relaxed by approximating the unknown class-conditional densities  $ p\left( {{\bf{x}}|\omega } \right)$ by Gaussian mixtures. In particular, we assume
\begin{equation}%%%公式38
	\begin{array}{l}
p\left( {{\bf{x}}|\omega } \right){\rm{ = }}\sum\limits_{m \in {{\cal M}_\omega }} {{w_{m\omega }}} f({\bf{x}}|{{\bf{\mu }}_{m\omega }},{{\bf{\Sigma }}_{m\omega }}),\\
\sum\limits_{m \in {{\cal M}_\omega }} {{w_{m\omega }}}  = 1,\;\omega  \in \Omega 
\end{array}
\end{equation}
where $m \in {{\cal M}_\omega }$ denotes the component’s index set of the class-conditional mixture $P\left( {{\bf{x}}|\omega } \right)$. Making substitution (38) in (19) we obtain the log-likelihood function in the following more general form:
\[L = \frac{1}{{\left| {\cal S} \right|}}\sum\limits_{{\bf{x}} \in {\cal S}} {\log \;\left[ {\sum\limits_{\omega  \in \Omega } {\sum\limits_{m \in {{\cal M}_\omega }} {p\left( {\omega |\psi ({\bf{x}})} \right)} } \;{w_{m\omega }}f({\bf{x}}|{{\bf{\mu }}_{m\omega }},{{\bf{\Sigma }}_{m\omega }})} \right]} \]
If we introduce the conditional component weights:
\begin{equation}%%%公式39
	\begin{array}{l}
h\left( {m,\omega |{\bf{x}},\psi ({\bf{x}})} \right)\\
 = \frac{{p\left( {\omega |\psi ({\bf{x}})} \right){w_{m\omega }}f({\bf{x}}|{{\bf{\mu }}_{m\omega }},{{\bf{\Sigma }}_{m\omega }})}}{{\sum\limits_{\omega  \in \Omega } {\sum\limits_{m \in {{\cal M}_\omega }} {p\left( {\omega |\psi ({\bf{x}})} \right)} } \;{w_{m\omega }}f({\bf{x}}|{{\bf{\mu }}_{m\omega }},{{\bf{\Sigma }}_{m\omega }})}},\\
m \in {{\cal M}_\omega },\omega  \in \Omega ,{\bf{x}} \in {\cal S}
\end{array}
\end{equation}
then, in analogy with (20), we can expand the log-likelihood criterion (36) in the form:	
\[\begin{array}{l}
L = \frac{1}{{\left| {\cal S} \right|}}\sum\limits_{{\bf{x}} \in {\cal S}} {\sum\limits_{\omega  \in \Omega } {\sum\limits_{m \in {{\cal M}_\omega }} {h\left( {m,\omega |{\bf{x}},\psi ({\bf{x}})} \right)\;} } } \\
 \cdot log\;\left[ {p\left( {\omega |\psi ({\bf{x}})} \right){w_{m\omega }}f({\bf{x}}|{\mu _{m\omega }},{{\bf{\Sigma }}_{m\omega }})} \right] - \\
 - h\left( {m,\omega |{\bf{x}},\psi ({\bf{x}})} \right)\log h\left( {m,\omega |{\bf{x}},\psi ({\bf{x}})} \right)\\
s.t.\;\sum\limits_{\omega  \in \Omega } {\sum\limits_{m \in {{\cal M}_\omega }} {h\left( {m,\omega |{\bf{x}},\psi ({\bf{x}})} \right){\rm{ = }}1} } 
\end{array}\]
Again, in analogy with (17) - (25), we come to the inequality
\begin{equation}%%%公式40
	\begin{array}{l}
L' - L\\
 \ge \frac{1}{{\left| {\cal S} \right|}}\sum\limits_{x \in {\cal S}} {\sum\limits_{\omega  \in \Omega } {\sum\limits_{m \in {{\cal M}_\omega }} {h\left( {m,\omega |{\bf{x}},\psi ({\bf{x}})} \right)\;} } } \\
log\;\left[ {\frac{{p'\left( {\omega |\psi ({\bf{x}})} \right){w_{m\omega }}f({\bf{x}}|\mu {'_{m\omega }},{\bf{\Sigma }}{'_{m\omega }})}}{{p\left( {\omega |\psi ({\bf{x}})} \right){w_{m\omega }}f({\bf{x}}|{\mu _{m\omega }},{{\bf{\Sigma }}_{m\omega }})}}} \right] \ge 0
\end{array}
\end{equation}
to be guaranteed for the sake of the monotonic property of EM algorithm. The right-hand side of (40) can be satisfied separately in two parts:
\begin{equation}%%%公式41
	\frac{1}{{\left| {\cal S} \right|}}\sum\limits_{{\bf{x}} \in {\cal S}} {\sum\limits_{\omega  \in \Omega } {\sum\limits_{m \in {{\cal M}_\omega }} {h\left( {m,\omega |{\bf{x}},\psi ({\bf{x}})} \right)\;} } } log\frac{{f({\bf{x}}|{{{\bf{\mu '}}}_{m\omega }},{\bf{\Sigma }}{'_{m\omega }})}}{{f({\bf{x}}|{{\bf{\mu }}_{m\omega }},{{\bf{\Sigma }}_{m\omega }})}} \ge 0
\end{equation}
\begin{equation}%%%公式42
	\sum\limits_{\omega  \in \Omega } {\frac{1}{{\left| {\cal S} \right|}}\sum\limits_{{\bf{x}} \in {\cal S}} {\sum\limits_{m \in {{\cal M}_\omega }} {h\left( {m,\omega |{\bf{x}},\psi ({\bf{x}})} \right)\;} } } log\frac{{p'\left( {\omega |\psi ({\bf{x}})} \right){{w'}_{m\omega }}}}{{p\left( {\omega |\psi ({\bf{x}})} \right){w_{m\omega }}}} \ge 0
\end{equation}
The first inequality (41) is satisfied if we define the new parameters ${\mu '_{m\omega }}$ ,${\bf{\Sigma }}{'_{m\omega }}$  by Eq. (33)
\begin{equation}%%%公式43
	\begin{array}{l}
\left\{ {{{\mu '}_{m\omega }},{\bf{\Sigma }}{'_{m\omega }}} \right\}{\rm{ = }}\\
\arg \;\mathop {\max }\limits_{\left\{ {{\mu _{m\omega }},{{\bf{\Sigma }}_{m\omega }}} \right\}} \left[ {\sum\limits_{{\bf{x}} \in {\cal S}} {h\left( {m,\omega |{\bf{x}},\psi ({\bf{x}})} \right)\;} log\;f({\bf{x}}|{\mu _{m\omega }},{{\bf{\Sigma }}_{m\omega }})} \right],\\
\omega  \in \Omega 
\end{array}
\end{equation}
Again, using the weighted likelihood analogy of the standard maximum likelihood estimate [46,47], we can write the following explicit solution of the Eq. (43) :
\begin{equation}%%%公式44
	\begin{array}{l}
{{\mu '}_{m\omega }} = \frac{1}{{\sum\nolimits_{{\bf{x}} \in {\cal S}} {h\left( {m,\omega |{\bf{x}},\psi ({\bf{x}})} \right)} }}\sum\limits_{{\bf{x}} \in {\cal S}} {h\left( {m,\omega |{\bf{x}},\psi ({\bf{x}})} \right)} \;{\bf{x}}\\
m \in {{\cal M}_\omega },\omega  \in \Omega 
\end{array}
\end{equation}
\begin{equation}%%%公式45
	\begin{array}{l}
{\bf{\Sigma }}{'_{m\omega }} = \frac{1}{{\sum\nolimits_{{\bf{x}} \in {\cal S}} {h\left( {m,\omega |{\bf{x}},\psi ({\bf{x}})} \right)} }}\\
 \cdot \sum\limits_{{\bf{x}} \in {\cal S}} {h\left( {m,\omega |{\bf{x}},\psi ({\bf{x}})} \right)} \;{\bf{x}}{{\bf{x}}^T} - {{\mu '}_{m\omega }}{{\mu '}_{m\omega }}^T\\
m \in {{\cal M}_\omega },\omega  \in \Omega 
\end{array}
\end{equation}
The inequality (42) can be further decomposed as follows
\begin{equation}%%%公式46
	\begin{array}{l}
\sum\limits_{\tilde \omega  \in \Omega } {\frac{{\left| {{{\cal S}_{\tilde \omega }}} \right|}}{{\left| {\cal S} \right|}}} \sum\limits_{\omega  \in \Omega } {\left[ {\frac{1}{{\left| {{{\cal S}_{\tilde \omega }}} \right|}}\sum\limits_{{\bf{x}} \in {{\cal S}_{\tilde \omega }}} {\sum\limits_{m \in {{\cal M}_\omega }} {h\left( {m,\omega |{\bf{x}},\tilde \omega } \right)} } } \right]} ,\\
log\frac{{p'\left( {\omega |\tilde \omega } \right)}}{{p\left( {\omega |\tilde \omega } \right)}} \ge 0
\end{array}
\end{equation}
\begin{equation}%%%公式47
	\sum\limits_{\omega  \in \Omega } {\sum\limits_{m \in {{\cal M}_\omega }} {\left[ {\frac{1}{{\left| {\cal S} \right|}}\sum\limits_{{\bf{x}} \in {\cal S}} {h\left( {m|\omega ,{\bf{x}},\psi ({\bf{x}})} \right)} } \right]\;} } log\frac{{{{w'}_{m\omega }}}}{{{w_{m\omega }}}} \ge 0
\end{equation}
where
\[h\left( {m|\omega ,{\bf{x}},\psi ({\bf{x}})} \right){\rm{ = }}\frac{{h\left( {m,\omega |{\bf{x}},\psi ({\bf{x}})} \right)}}{{\sum\nolimits_{m \in {{\cal M}_\omega }} {h\left( {m,\omega |{\bf{x}},\psi ({\bf{x}})} \right)} }}\]
Considering the inequality (17) we can write again the EM iteration equations for the parameters $p'\left( {\omega |\tilde \omega } \right)$ , ${w'_{m\omega }}$  in explicit form. In particular, the inequality (45) is satisfied if we set
\begin{equation}%%%公式48
	p'\left( {\omega |\tilde \omega } \right) = \frac{1}{{\left| {{{\cal S}_{\tilde \omega }}} \right|}}\sum\limits_{{\bf{x}} \in {{\cal S}_{\tilde \omega }}} {\sum\limits_{m \in {{\cal M}_\omega }} {h\left( {m,\omega |{\bf{x}},\tilde \omega } \right)} } ,\omega  \in \Omega ,\;\tilde \omega  \in \Omega 
\end{equation}
and the second inequality (47) is satisfied if we define the new component weights ${w'_{m\omega }}$ by equation:
\begin{equation}%%%公式49
	\begin{array}{l}
{{w'}_{m\omega }}{\rm{ = }}\frac{1}{{\left| {\cal S} \right|}}\sum\limits_{{\bf{x}} \in {\cal S}} {h\left( {m|\omega ,{\bf{x}},\psi ({\bf{x}})} \right)} \;\\
{\rm{ = }}\frac{1}{{\left| {\cal S} \right|}}\sum\limits_{{\bf{x}} \in {\cal S}} {\frac{{h\left( {m,\omega |{\bf{x}},\psi ({\bf{x}})} \right)}}{{\sum\nolimits_{m \in {{\cal M}_\omega }} {h\left( {m,\omega |{\bf{x}},\psi ({\bf{x}})} \right)} }},} \\
m \in {{\cal M}_\omega },\omega  \in \Omega 
\end{array}
\end{equation}
The EM algorithm for the problem of estimating class-conditional Gaussian mixtures with noisy labels can be summarized in terms of the iteration equations (39), (44), (46), (48) and (49).
\section{RELATED WORK}%4
The problem of discriminant analysis has been studied by researchers from many disciplines, such as physical, biological and social sciences, cognitive science, psychology, engineering, and medicine \cite{23ji2008generalized}. Recently, the discriminant analysis with label noise has gained substantial research attention. Various solution strategies have been proposed to prevent a learning algorithm from overfitting the noisy data, the robust classifiers with capability to diminish the effect of label noise to a certain extent have obtained varying levels of success. [24, 25, 26, 27, 28, 29, 30, 31, 32]\cite{24durrant2012error,25kearns1998efficient,26angluin1988learning,27li2007classification,28natarajan2013learning,29gomez2006boosting,30long2010random,31khoshgoftaar2010supervised,32lee2003learning}. For instance, in \cite{29gomez2006boosting}, the emphasis functions that combine both sample errors and their proximity to the classification border are explored. Long and Servedio demonstrated in \cite{30long2010random} that for a broad class of convex potential functions, any boosting algorithm was highly susceptible to random classification noise. They also emphasized that the result was unsuitable for non-convex potential function. In \cite{31khoshgoftaar2010supervised}, a comprehensive empirical investigation using neural network algorithms to learn from imbalanced data with labeling errors was explored.\\

Lee and Liu transformed the learning problem with positive and unlabeled examples into a problem of learning with noise by labeling all unlabeled examples as negative and using logistic regression to learn from the weighting noisy examples \cite{32lee2003learning}. In \cite{33liu2015classification}, based on consistency assurance that the label noise ultimately did not hinder the search for the optimal classifier of the noise-free sample, the study proved that any surrogate loss function could be used for classification with noisy labels by using importance reweighting. \cite{33liu2015classification} also showed that the noise rate that could be estimated was upper bounded by the conditional probability of the noisy sample. Bootkrajang and Kabán built a discriminative model by modeling class noise distributions and reinterpreted existing discriminative models from the class noise perspective \cite{21bootkrajang2012label}. They proved that the error of label-noise robust logistic regression was bounded, and that label-noise robust logistic regression behaved in the same way as logistic regression when label noise did not exist or when the label flipping was symmetric. They also demonstrated that the weighting mechanism of label-noise robust logistic regression improved upon logistic regression with asymmetric label flipping. However, in \cite{21bootkrajang2012label}, the loss function did not define the latent true label but defined the observed noisy label instead. Rooyen et al. proposed in \cite{34van2015learning} a convex classification calibrated unhinged loss and proved that it is robust under symmetric label noise. The loss further avoided minimization of any convex potential over a linear function class that could result in classification performance equivalent to random guessing. In \cite{362015Learning}, corruption problems that were classified as mutually contaminated distributions were considered, and authors argued that optimized balanced error on corrupted data was equivalently optimized as the binary label error on clean data.\\

Based on the boundary conditional class noise assumption, instead of modeling data generation or conditional class probability both for symmetric and asymmetric cases, Jun and Cai assumed that the class noise was distributed as an unnormalized Gaussian and an unnormalized Laplace centered on the linear class boundary, and proposed Gaussian noise model and Laplace noise model, respectively \cite{35du2015modelling}. They then further reinterpreted logistic regression and probit regression by using the proposed class noise probability.\\

Previous theoretical work on the label noise problem assumed that the two classes were separable, and the label noise was independent of the true class label or that the noise proportions for each class were known. \cite{362015Learning,37scott2013classification} introduced a general mixture proportion estimation framework for classification with label noise that eliminated these assumptions. When the class-conditional distributions overlapped and the label noise was not symmetric, \cite{362015Learning,37scott2013classification} presented assumptions ensuring identifiability and the existence of a consistent estimator of the optimal risk and given associated estimation strategies. For any arbitrary pair of contaminated distributions, a unique pair of non-contaminated distributions satisfied the proposed assumptions. Scott argued in \cite{38scott2013classification} that a solution to mixture proportion estimation led to solutions to various weakly supervised learning problems, such as anomaly detection, learning from positive and unlabeled examples, domain adaptation, and classification with label noise. He established a rate of convergence for mixture proportion estimation under an appropriate distributional assumption based on surrogate risk minimization and showed that this rate of convergence can derive the consistency of the algorithm and provide a practical implementation of mixture proportion estimation and demonstrate its efficacy in classification with noisy labels \cite{38scott2013classification}.\\

By modeling the corruption process through a Markov kernel and defining the corrupted learning problem to be the corrupted experiment, Brendan and Williamson developed a general framework for tackling corrupted learning problems as well as introduced minimax upper and lower bounds on the risk for learning in the presence of corruption \cite{392015}.\\

Manwani and Sastry studies in \cite{40van2015learning} noise tolerance under risk minimization. They assume that the actual training set given to the learning algorithm was obtained from the noise-free data set, the class label of each example is corrupted and that a learning method was noise tolerant if the classifiers learned with noise-free data and with noisy data, and both have the same classification accuracy on the noise-free data. They showed that risk minimization under 0-1 loss function was a promising approach for learning from noisy training data, and that Fisher linear discriminant and linear least squares under squared error loss were noise tolerant under uniform noise, but not under non-uniform noise. The risk minimization under other loss functions was not noise tolerant \cite{41manwani2013noise}.\\

A great deal of research has been conducted on both theory and applications for such label noise problem. Despite much attention paid to discriminant analysis for noisy data \cite{421995Can}, the investigation focused on the instances of generating a single Gaussian model. Furthermore, symmetric and asymmetric label noise was introduced to describe the contaminated distribution of corrupted binary labels. However, the instances that belonged to the same class usually were ruled by multiple GMM because of the presence of non-Gaussian distribution, which is mixed proportionally by Gaussian distribution of different means and variances. However, to the best of our knowledge, the discriminant analysis with noisy labels based on GMM has received limited research attention mainly because of mathematical difficulties. In particular, the commonly used approaches, such as matrix analysis, are no longer directly applicable to deal with both symmetric and asymmetric label noise problem because the presence of asymmetric label noise cannot be expressed in the normal form. In this paper, therefore, we intend to tackle such an important yet challenging problem. \cite{43Han2019DeepSF} presents a novel deep self-learning framework, which does not rely on any assumption on the distribution of the noisy labels, and train a robust network on the real noisy datasets without extra supervision.\\

Similar to our approach, Bouveyron also proposed to use the explicit global mixture model of more than two classes \cite{44bouveyron2009robust}, however, Bouveyron’s method is totally different from our approach. Bouveyron’s approach compare the supervised information given by the learning data with an unsupervised modelling based on the Gaussian mixture model, if some learning data have wrong labels, the comparison of the supervised information with an unsupervised modelling of the data allows to detect the inconsistent labels. Then it is possible afterward to build a supervised classifier by giving a low confidence to the learning observations with inconsistent labels.
\section{EXPERIMENTS AND DISCUSSION}%5
\subsection{Datasets and Preprocessing}%5.1
Synthetic datasets and real-world datasets are used in our experiments. Table I presents a summary of the datasets.
\begin{table}%%%表1
	\centering
	\caption{CHARACTERISTICS OF THE DATASETS}
	\begin{tabular}{c|ccc}
		\hline\hline
		\multirow{2}{*}{\textbf{Dataset}} & \multicolumn{3}{|c}{\textbf{Characteristics}}\\ 
		\cline{2-4} & \textbf{Samples}	& \textbf{Dimensionality} &	\textbf{Classes} \\
		\hline\hline
		Synth1 &	2000 &	30 &	5\\
		Synth2 &	1000 &	2 &	2\\
		Breast &	106 &	 9 &	6\\
		Iris &	150 &	 4 &	3\\
		Wine &	178 & 	13 &	3\\
		Heart &	267 & 	22 &	2\\
		Boston &	506 & 	13 &	2\\
		Waveform & 	5000 &	21 &	3\\
		\hline\hline
	\end{tabular}
\end{table}
Two synthetic datasets are created randomly by our Matlab code. We apply the following real-world datasets: Boston, Breast Issue, SPECT Heart, Waveform, Wine, and Iris. Real-world datasets are UCI datasets \cite{45lichman2013uci}.\\

The datasets are equally divided into training data and test data. The original class labels are treated as true labels. Symmetric and asymmetric label errors are injected into the datasets artificially. As the label noise was generated randomly, the label noise rate and label error rate were not equal.
\begin{figure}[htbp]%%%图1
	\centering
	\includegraphics[scale=0.4]{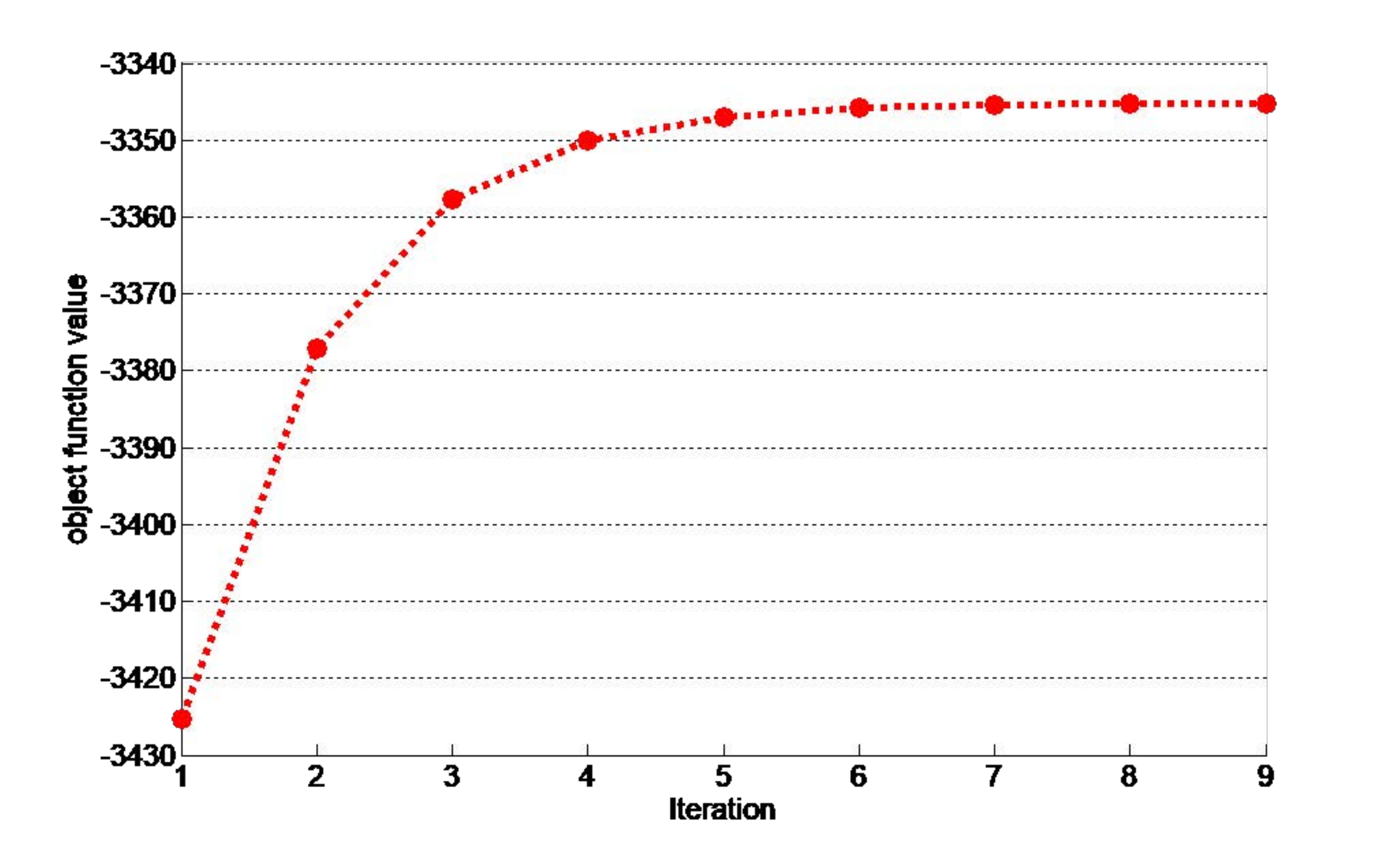}
	\caption{Convergence of the proposed method using EM algorithm}
\end{figure}
\subsection{Results and Discussion}%5.2
First, we illustrate the convergence of object function value of the proposed method. We use the synth1 dataset with 20\% label errors as a representative. The objective function value is almost stable after 5 iterations in Fig. 1. As shown in Fig. 2, the error rate of prediction after 5 iterations is 12.80\% and 12.80\% after 10 iterations. The error rate changes rapidly during the first 4 iterations, and the error rate stabilizes after 5 iterations. The proposed method using the EM approach is convergent and effective. The change of the mixture models centers at different iterations is illustrated in Fig. 3.
\begin{figure}[htbp]%%%图2
	\centering
	\includegraphics[scale=0.4]{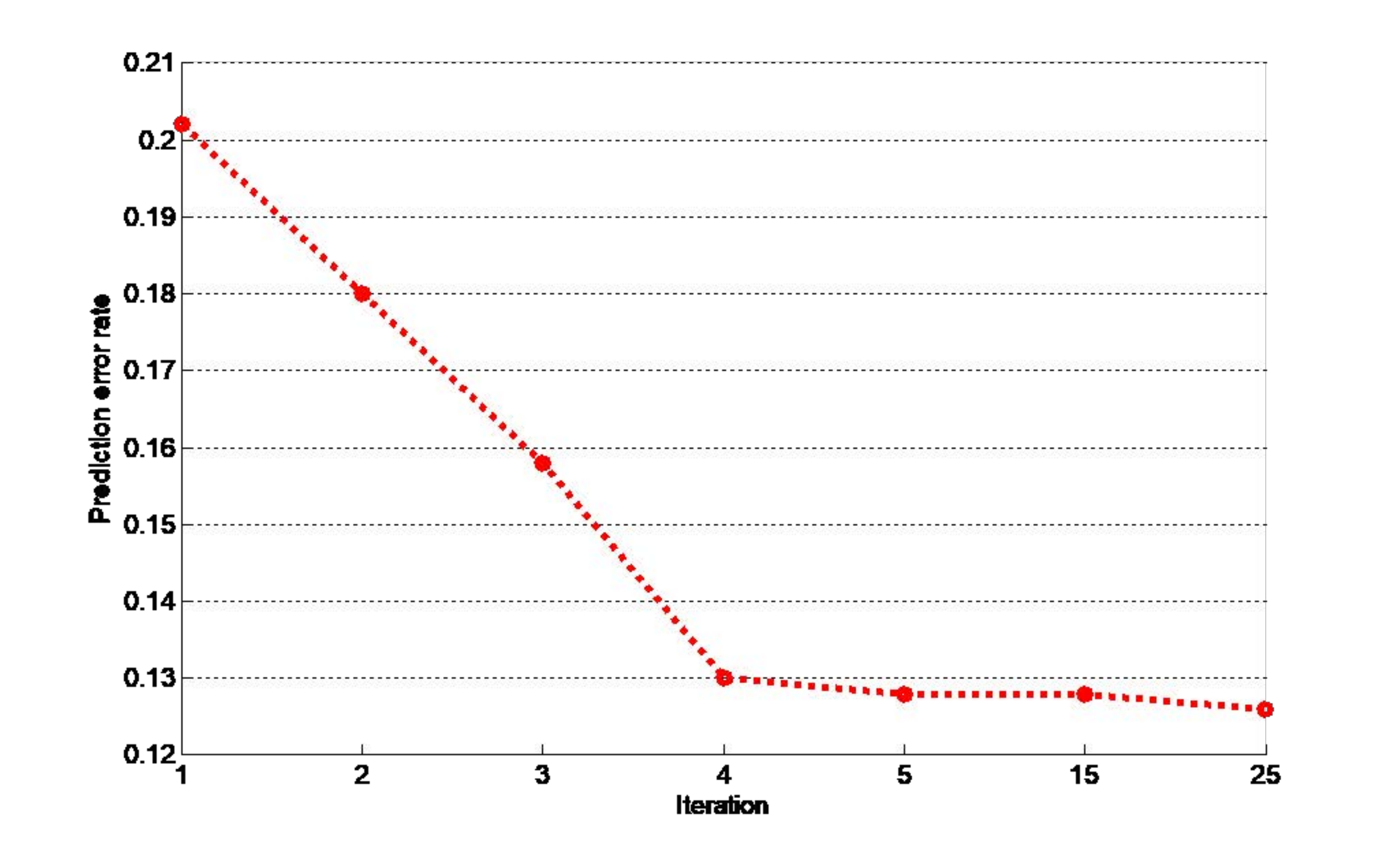}
	\caption{Error rates at different iterations}
\end{figure}
We compare the proposed method to rNDA, AdaBoost, rLR, and rmLR. As the maximum likelihood used in our method is totally dependent on the training data, we change the training sample size from 20\% to 80\% to determine the effect of the size of the training data on the performance of classifiers. Fig. 4 shows the effect of the number of training samples on error rates of the predictions of the methods. All the methods mentioned are affected by the number of training samples, and all methods perform best with 50\% samples. We run the following experiments with 50\% samples.\\

Fig. 5 shows the original dataset synth2 and the estimated mixture centers obtained by our method in comparison with 20\% label errors synth2 and its estimated mean vectors obtained by our method. Table 2 shows the parameters and the error rates of predictions obtained by our method using the original dataset synth2 and synth2 with 20\% label errors. The diagonal elements ${\gamma _{ij}}\left( {i = j} \right)$ indicate the probability of labels that are not flipped. Table II shows that the unflipped probabilities of original dataset synth2 are extremely close to 1, whereas that of 20\% label errors data are close to 0.8; and the real probability of each class is 0.5. The estimated class probabilities are all close to 0.5. The results confirm the reality. The error rates of predictions using the two datasets mentioned are 13.00\% and 12.60\%, respectively, which differ slightly. In the noisy case, prediction is better than the original case because our method has already considered flipping.
\begin{figure}[htbp]%%%图3
	\centering
	\subfigure[Iteration 1]{\includegraphics[width=5.5cm]{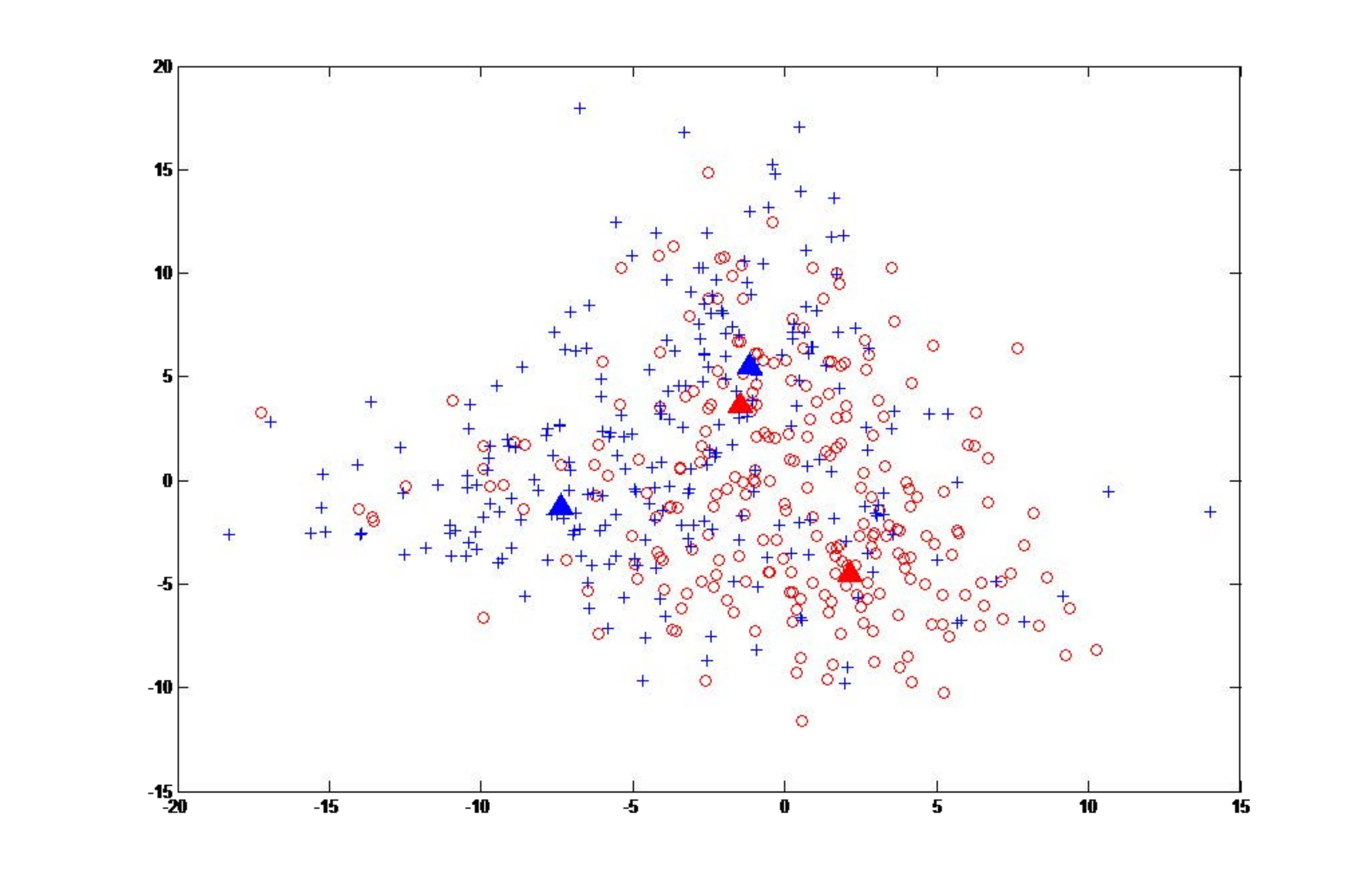}}
	\quad
	\subfigure[Iteration 5]{\includegraphics[width=5.5cm]{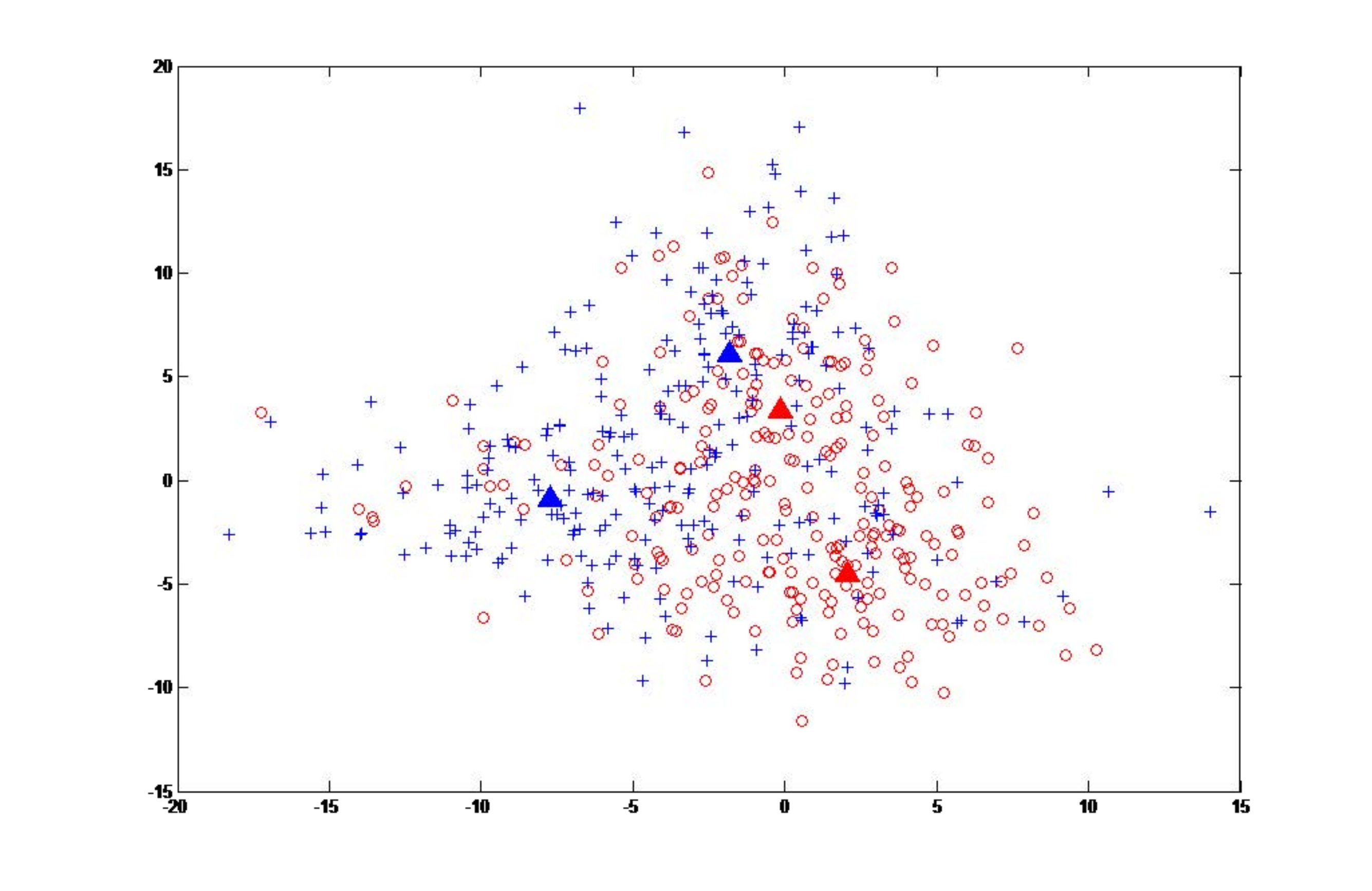}}
	\quad
	\subfigure[Iteration 15]{\includegraphics[width=5.5cm]{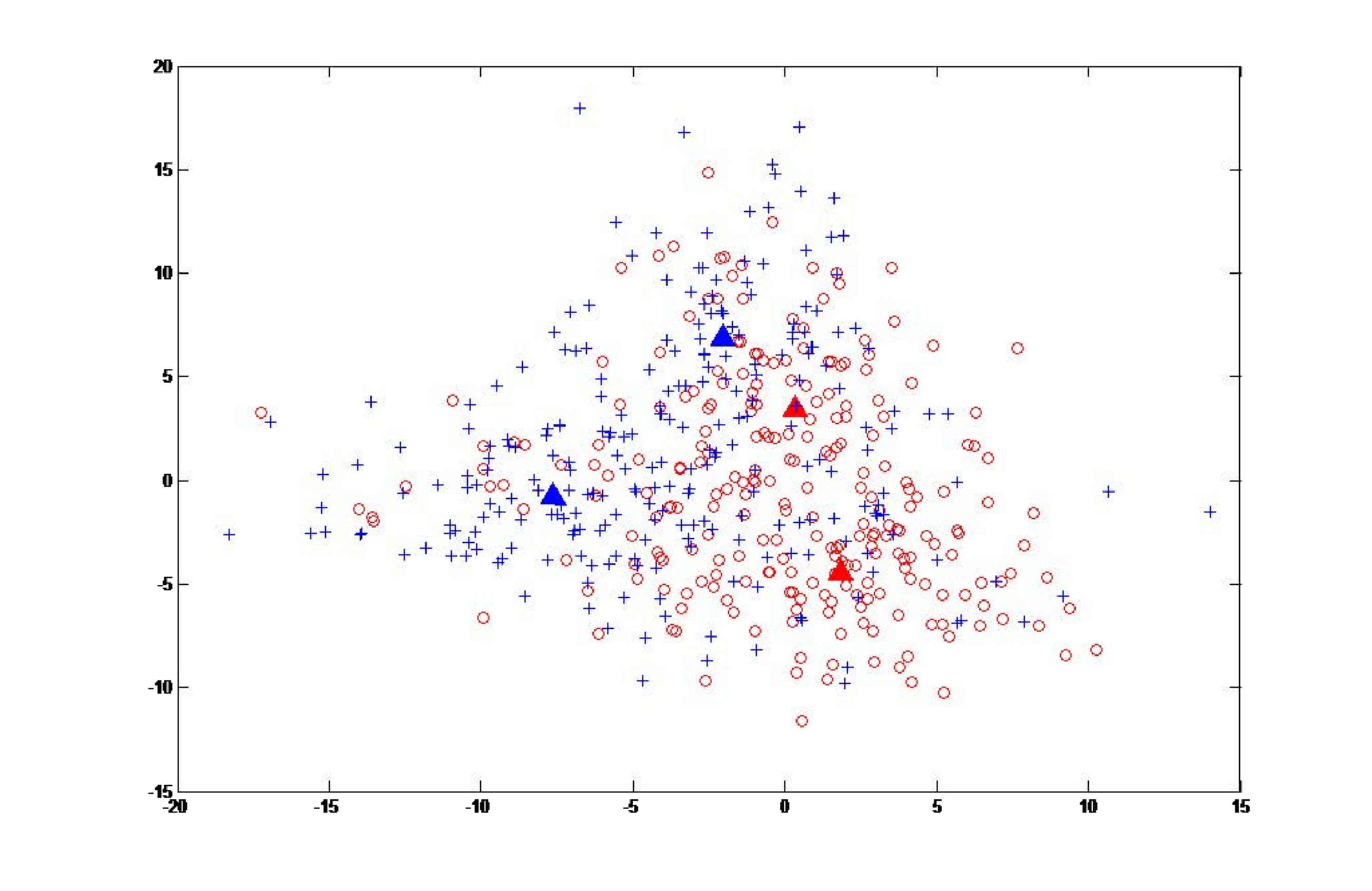}}
	\quad
	\subfigure[Iteration 20]{\includegraphics[width=5.5cm]{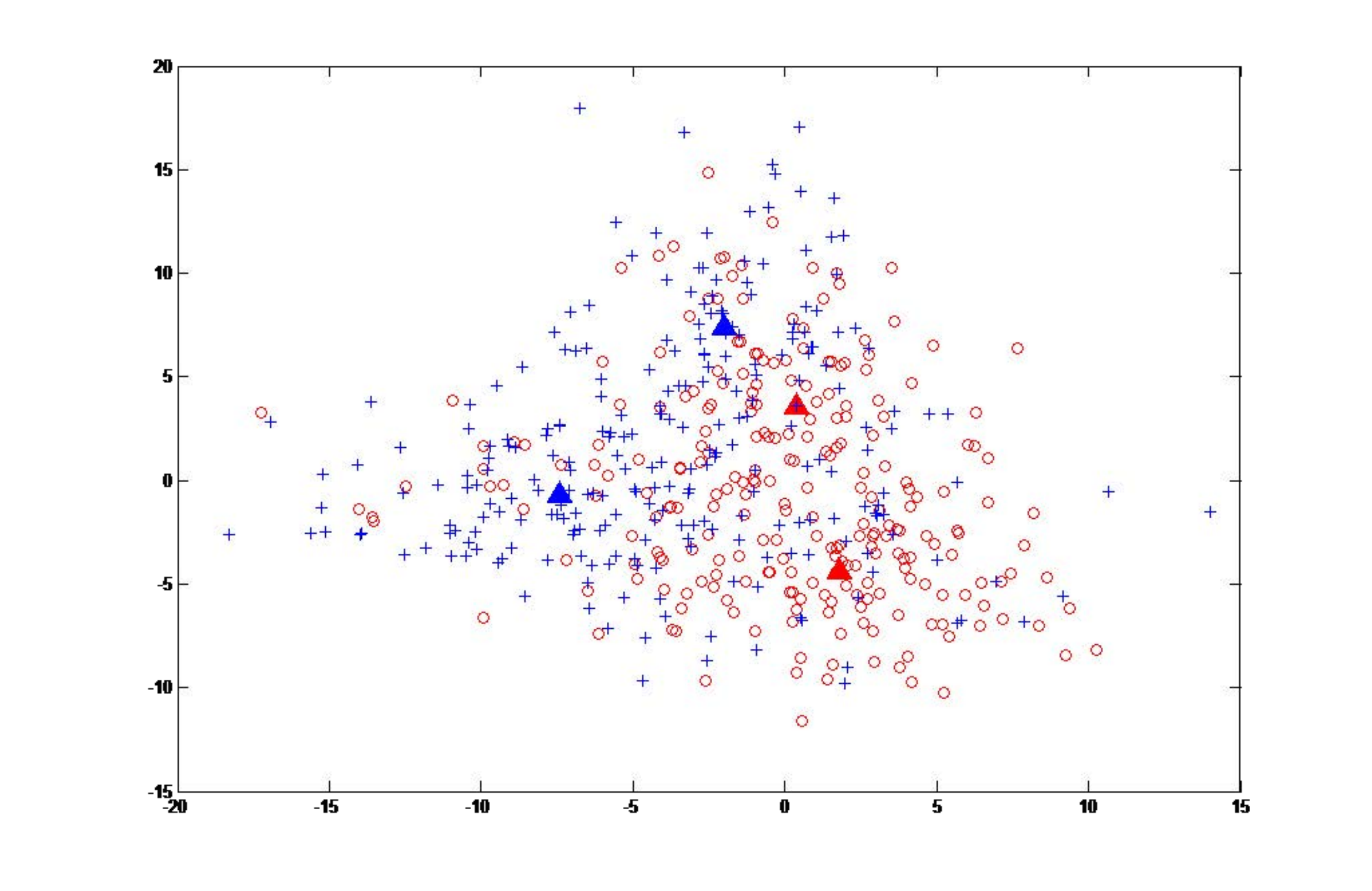}}
	\quad
	\caption{Model’s mixture centers at different iterations}
\end{figure}
\begin{table}%%%表2
	\centering
	\caption{RESULTS ON DATASET SYNTH2 USING THE PROPOSED METHOD}
	\begin{tabular}{c|ccc}
		\hline
		\multirow{2}{*}{\textbf{Dataset}} & \multicolumn{3}{|c}{\textbf{Results}}\\ 
		\cline{2-4} & \textbf{Flipping Probability}	& \textbf{Class Probability} &	\textbf{Error Rate} \\
		\hline
		\textbf{Original Synth2}	& $\left[ {\begin{array}{*{20}{c}}1&{2.49e - 05}\\{3.03{\rm{e}} - 04}&1\end{array}} \right]$ & $\left[ {\begin{array}{*{20}{c}}
{0.4994}&{0.5006}\end{array}} \right]$ & 13.00\% \\
		\hline
		\textbf{Synth2 with 20\% Label Error}	 & $\left[ {\begin{array}{*{20}{c}}{0.7825}&{0.2175}\\{0.1977}&{0.8023}\end{array}} \right]$ & $\left[ {\begin{array}{*{20}{c}}{0.5101}&{0.4899}\end{array}} \right]$ & 12.60\% \\
		\hline
	\end{tabular}
\end{table}
\begin{figure}[htbp]%%%图4
	\centering
	\includegraphics[scale=0.4]{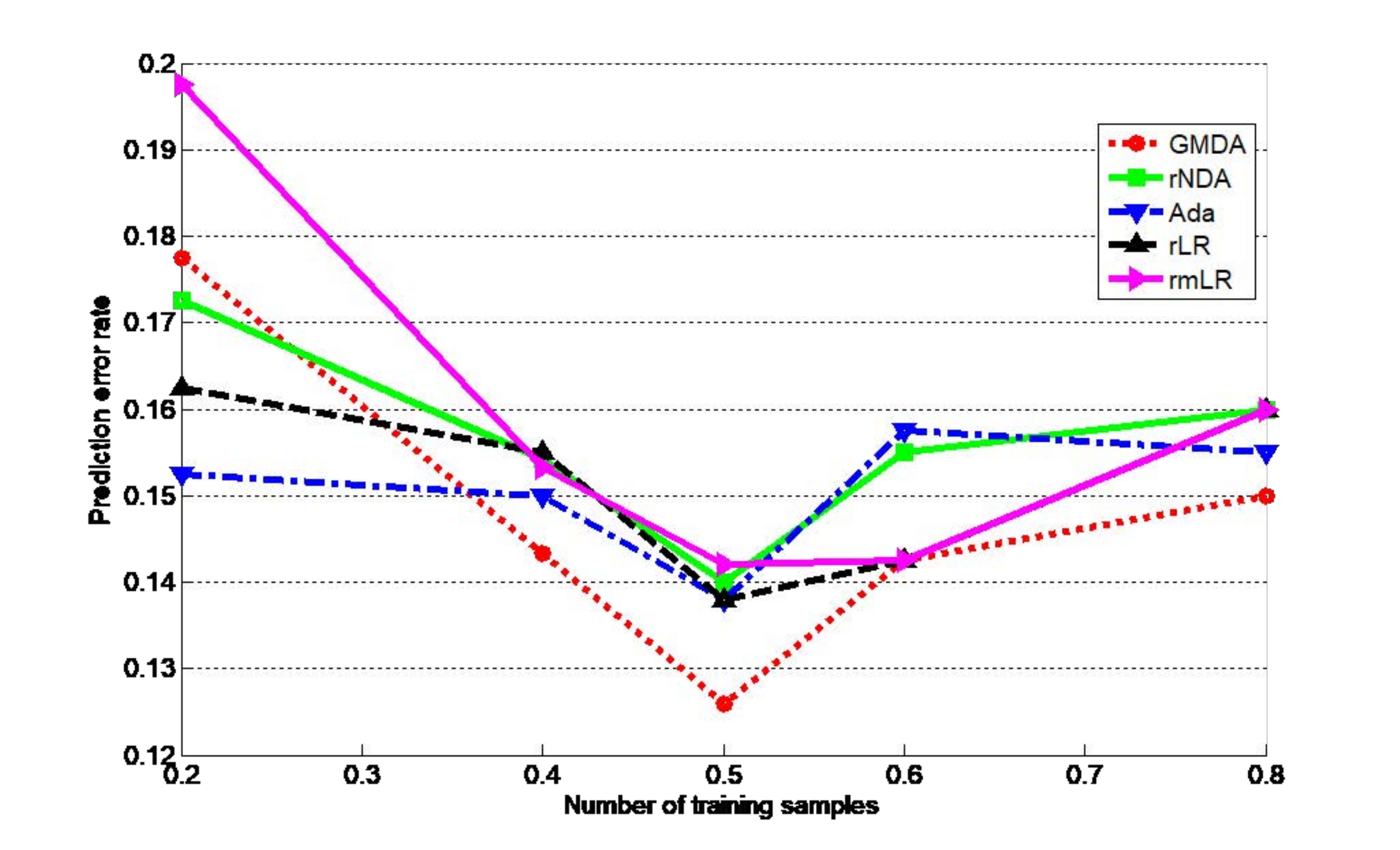}
	\caption{Effect of the number of samples}
\end{figure}
Fig. 6 shows the results of two synthetic datasets and six real-world datasets using the proposed method and the other four methods mentioned. All results are presented in Tables III–V. The bold values of error in \% shown in Table III and IV indicate the winners.The trend is that at a higher label noise level, a higher prediction error rate is obtained. The AdaBoost method, which is a label noise-robust method, is affected most by the label errors and has mostly the largest error rate. The rLR and the rmLR methods have similar performance. The proposed method has a much better performance and the lowest error rate in most cases, and outperforms others mostly both in the symmetric and asymmetric noise cases. The AdaBoost, rLR, and rmLR methods cannot predict when samples of one class label are all flipped in binary cases.
\begin{figure}[htbp]%%%图5
	\centering
	\subfigure[Iteration 1]{\includegraphics[width=5.5cm]{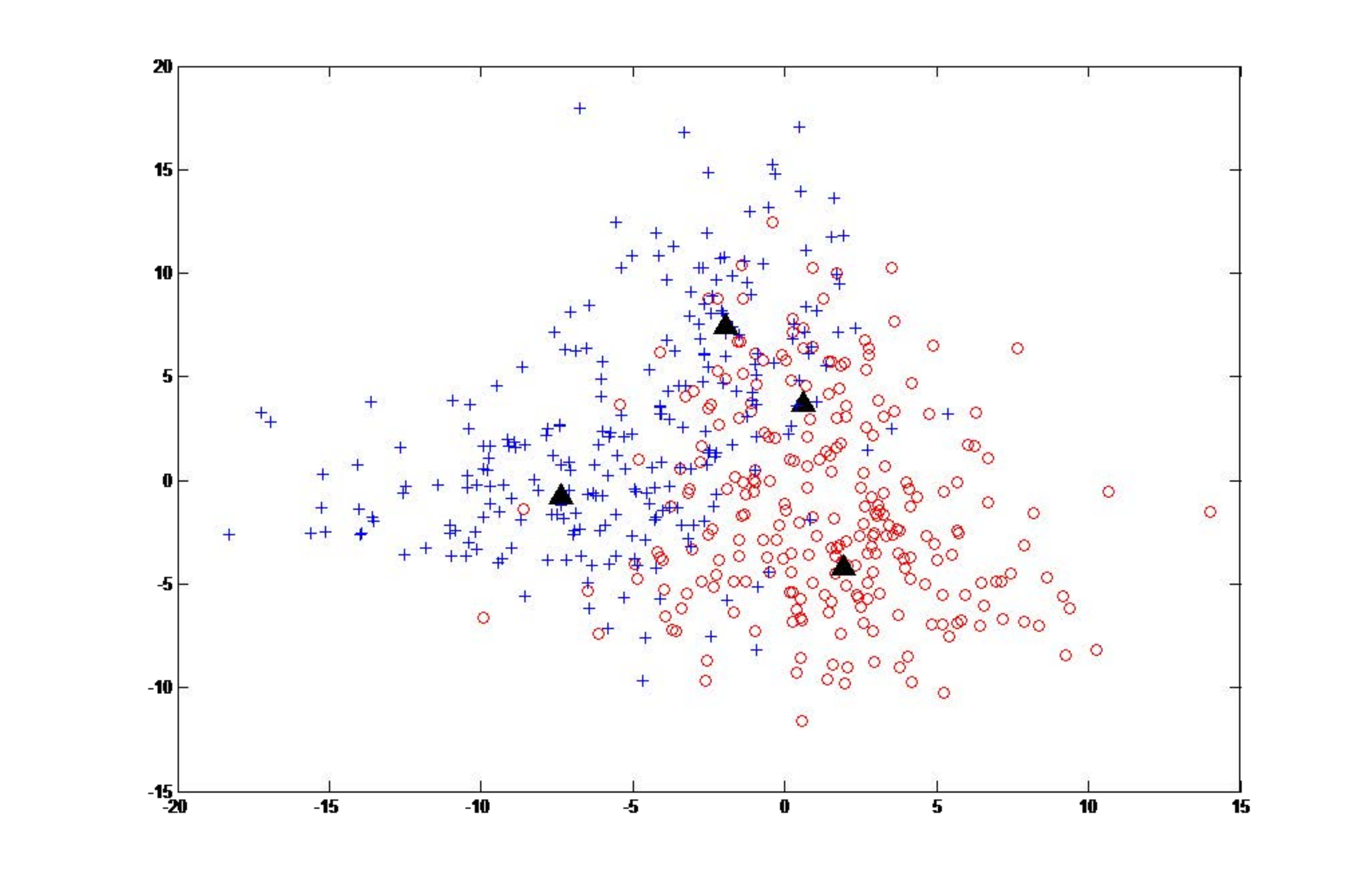}}
	\quad
	\subfigure[Iteration 5]{\includegraphics[width=5.5cm]{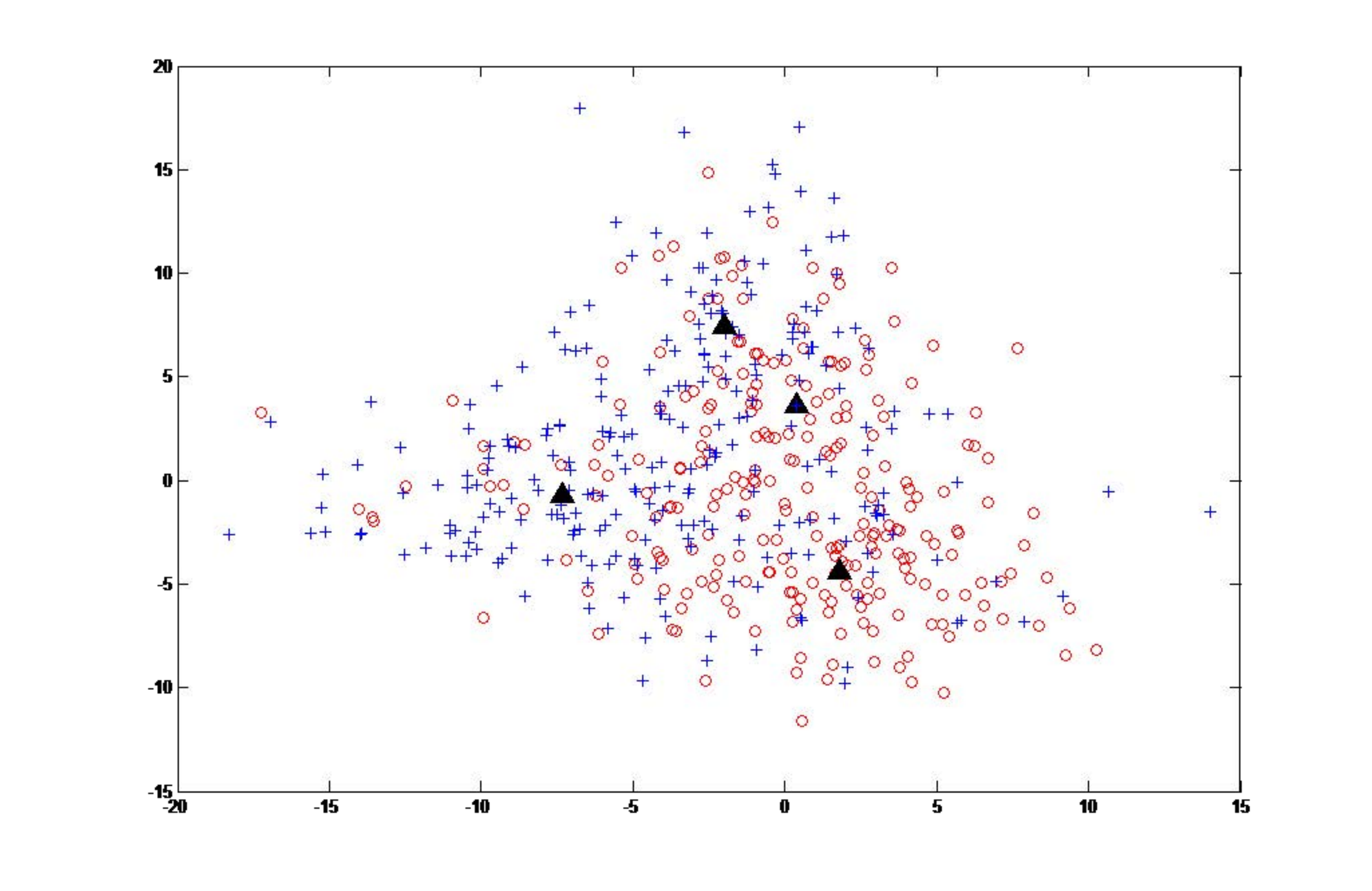}}
	\quad
	\caption{Dataset synth2 and estimated mixture centers obtained by using the proposed method}
\end{figure}
On the other hand, Fig. 6 also shows that the proposed method significantly outperforms other methods on synthetic datasets synth1 with larger dimension. We run another experiment to determine the effect of the dimension of the datasets. We create a set of datasets consisting of 1000 samples, three classes, and dimension from 5 to 30. Then, 20\% label errors are artificially injected into these datasets.
\begin{figure}[htbp]%%%图6
	\centering
	\subfigure[Dataset synth1 with symmetric label errors]{\includegraphics[width=5.5cm]{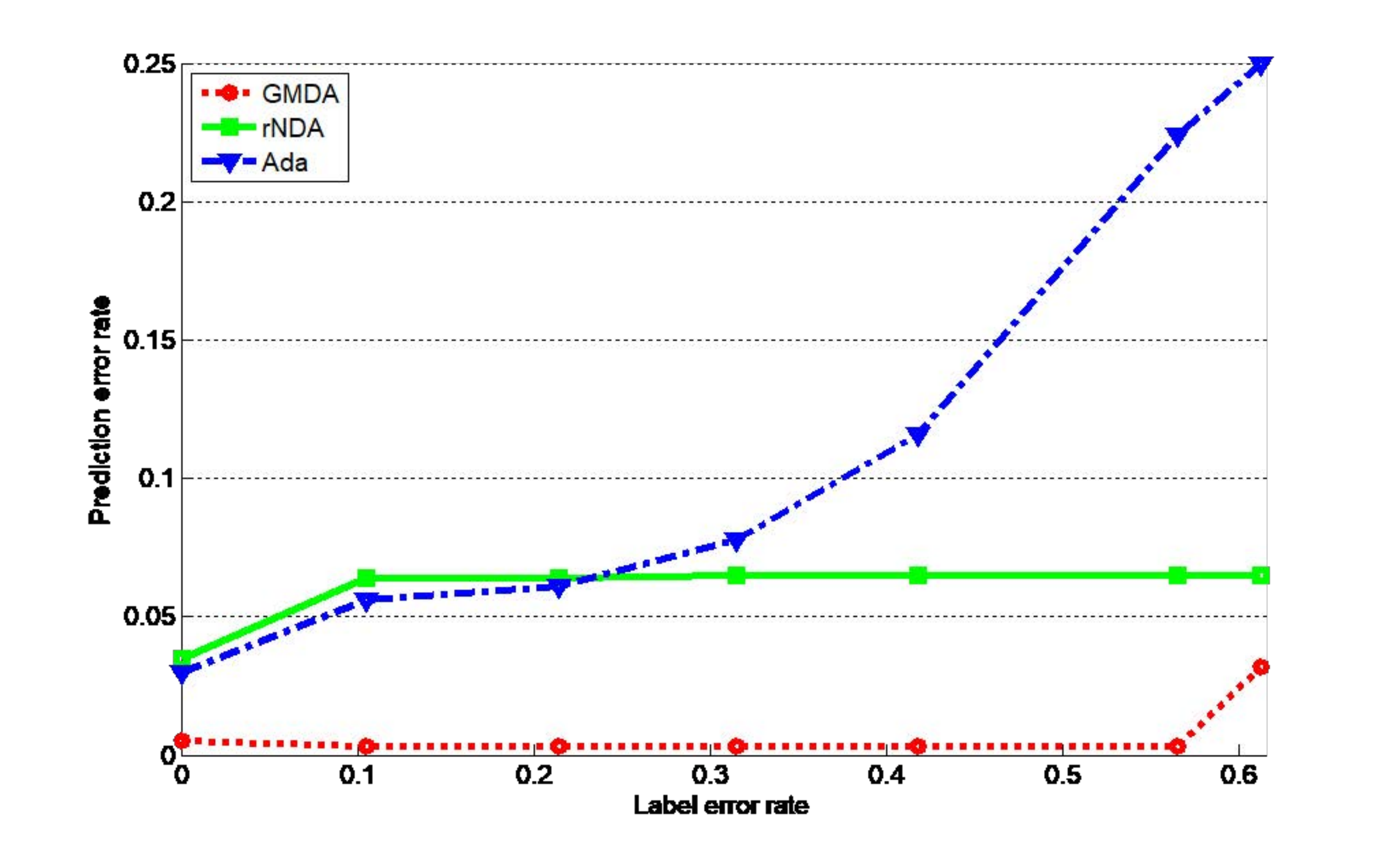}}
	\quad
	\subfigure[Dataset synth1 with asymmetric label errors]{\includegraphics[width=5.5cm]{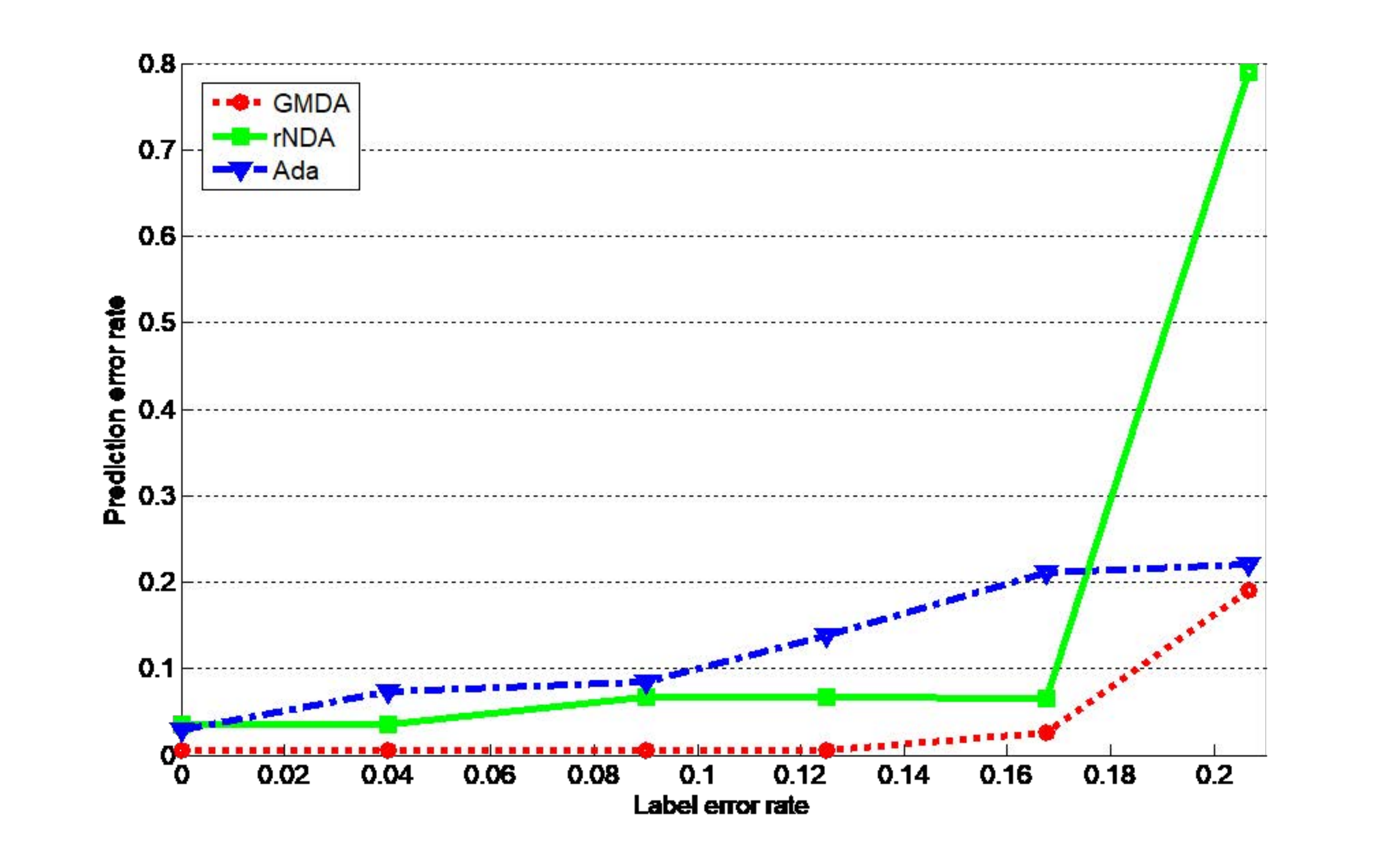}}
	\quad
	\subfigure[Dataset synth2 with symmetric label errors]{\includegraphics[width=5.5cm]{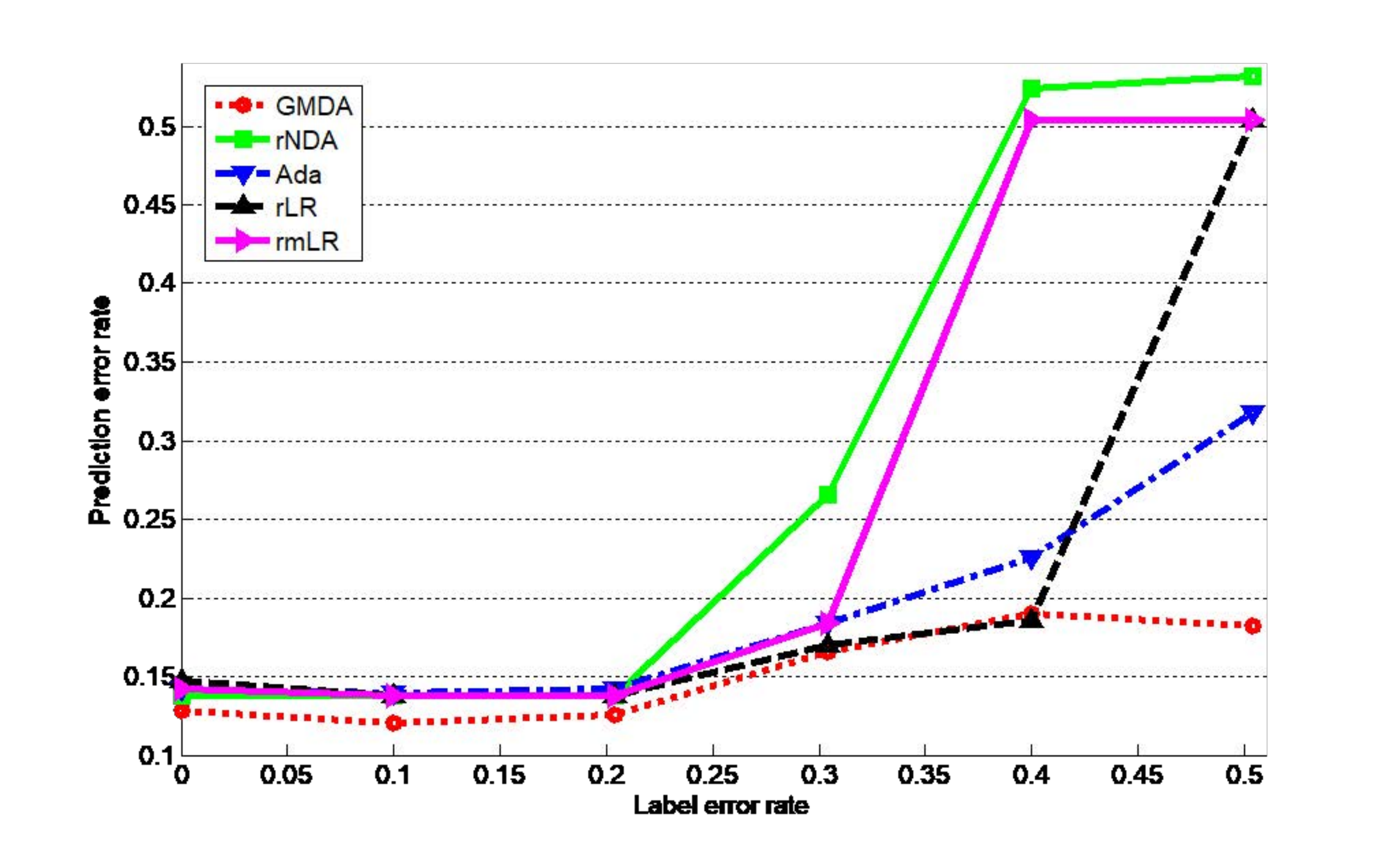}}
	\quad
	\subfigure[Dataset synth2 with asymmetric label errors]{\includegraphics[width=5.5cm]{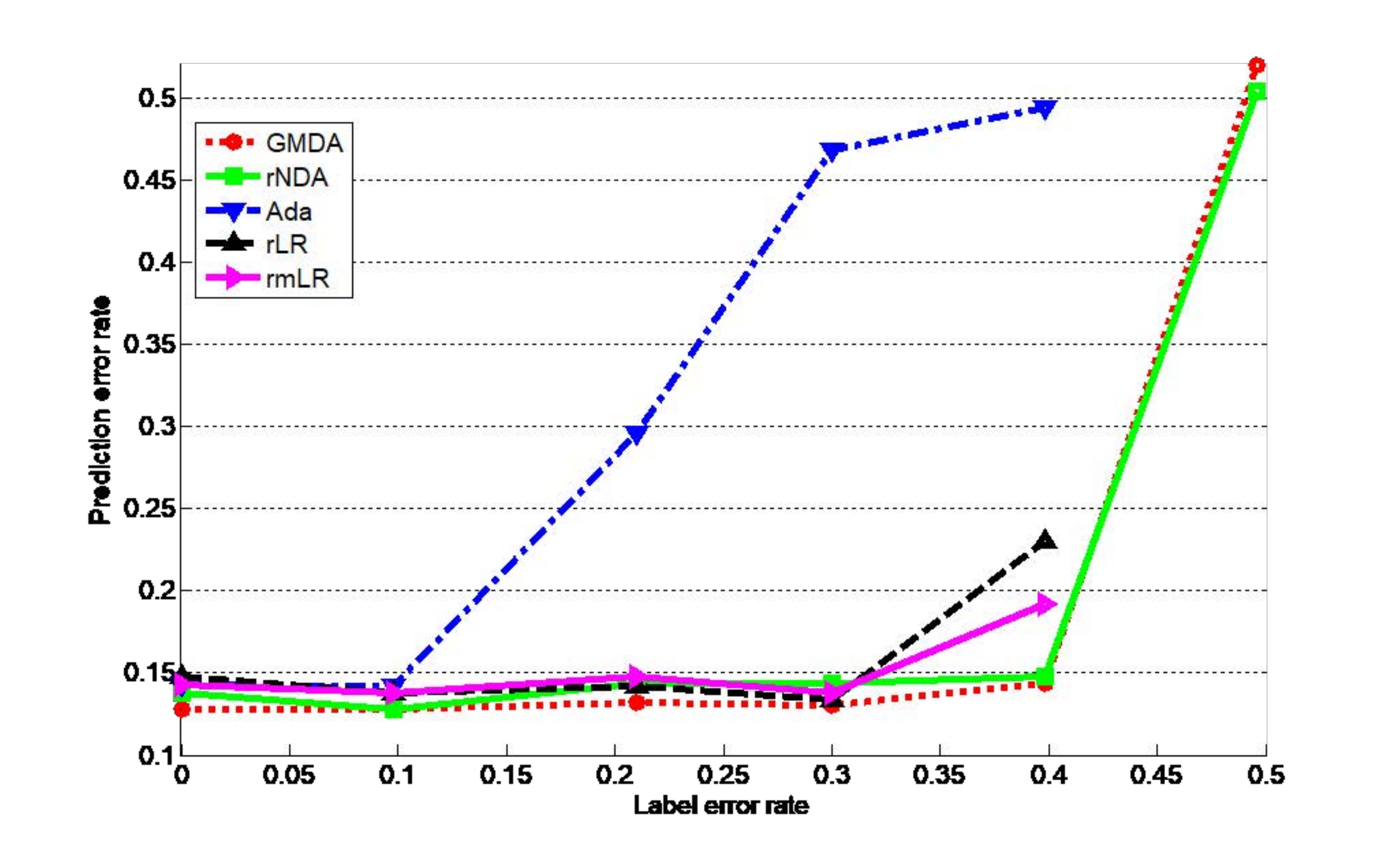}}
	\quad
	\subfigure[Boston dataset with symmetric label errors]{\includegraphics[width=5.5cm]{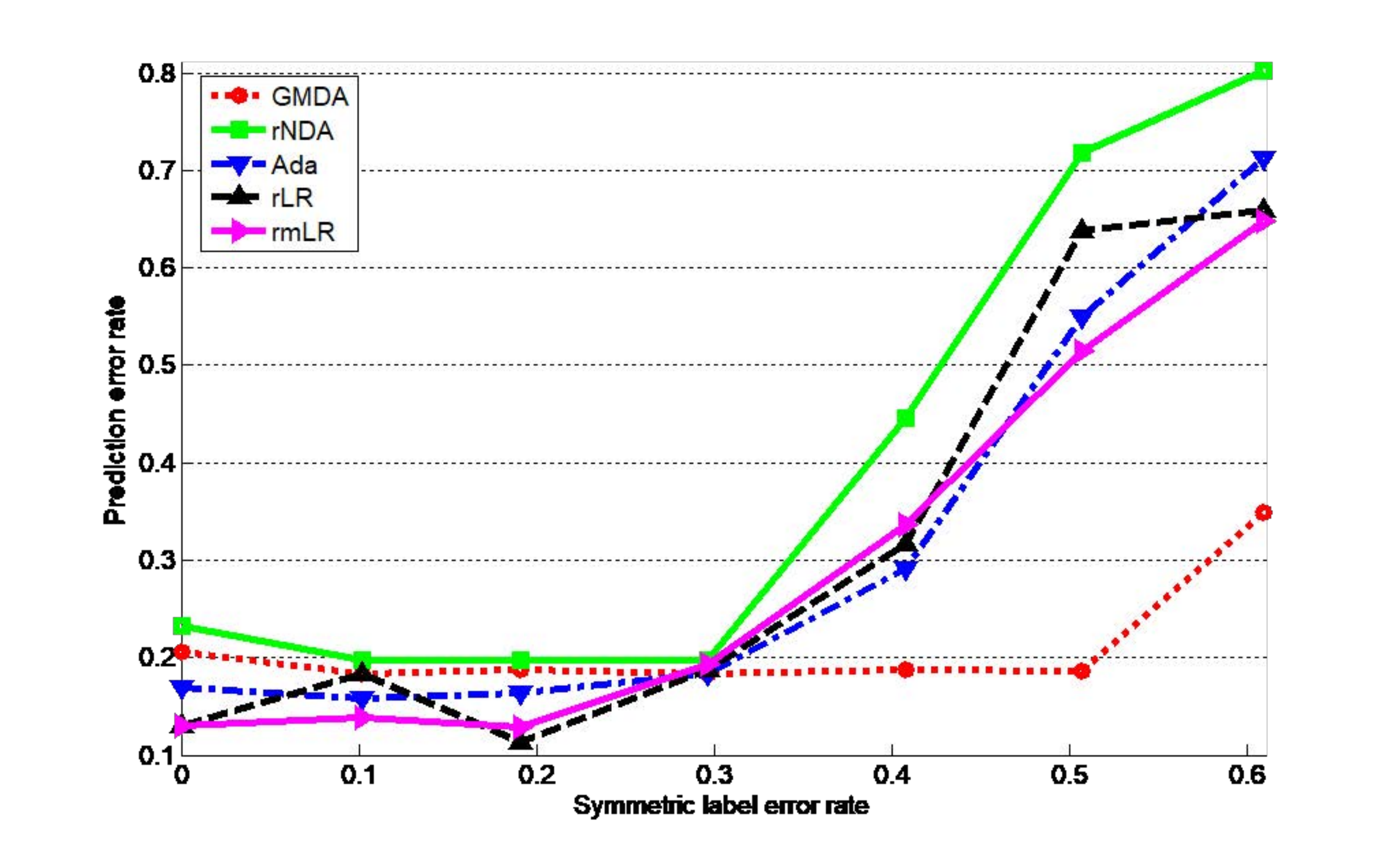}}
	\quad
	\subfigure[Boston dataset with asymmetric label errors]{\includegraphics[width=5.5cm]{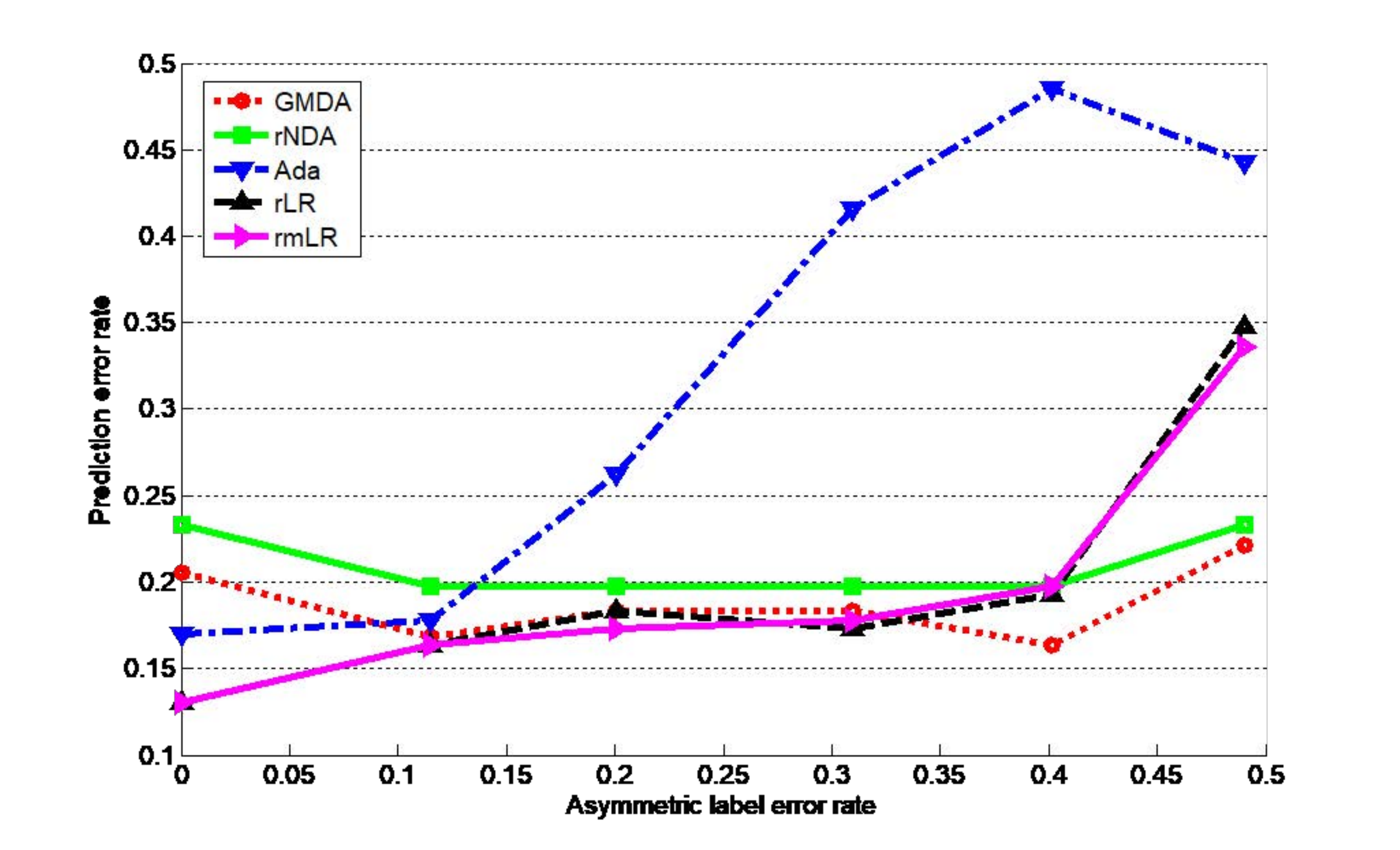}}
	\quad
	\subfigure[Breast Issue dataset with symmetric label errors]{\includegraphics[width=5.5cm]{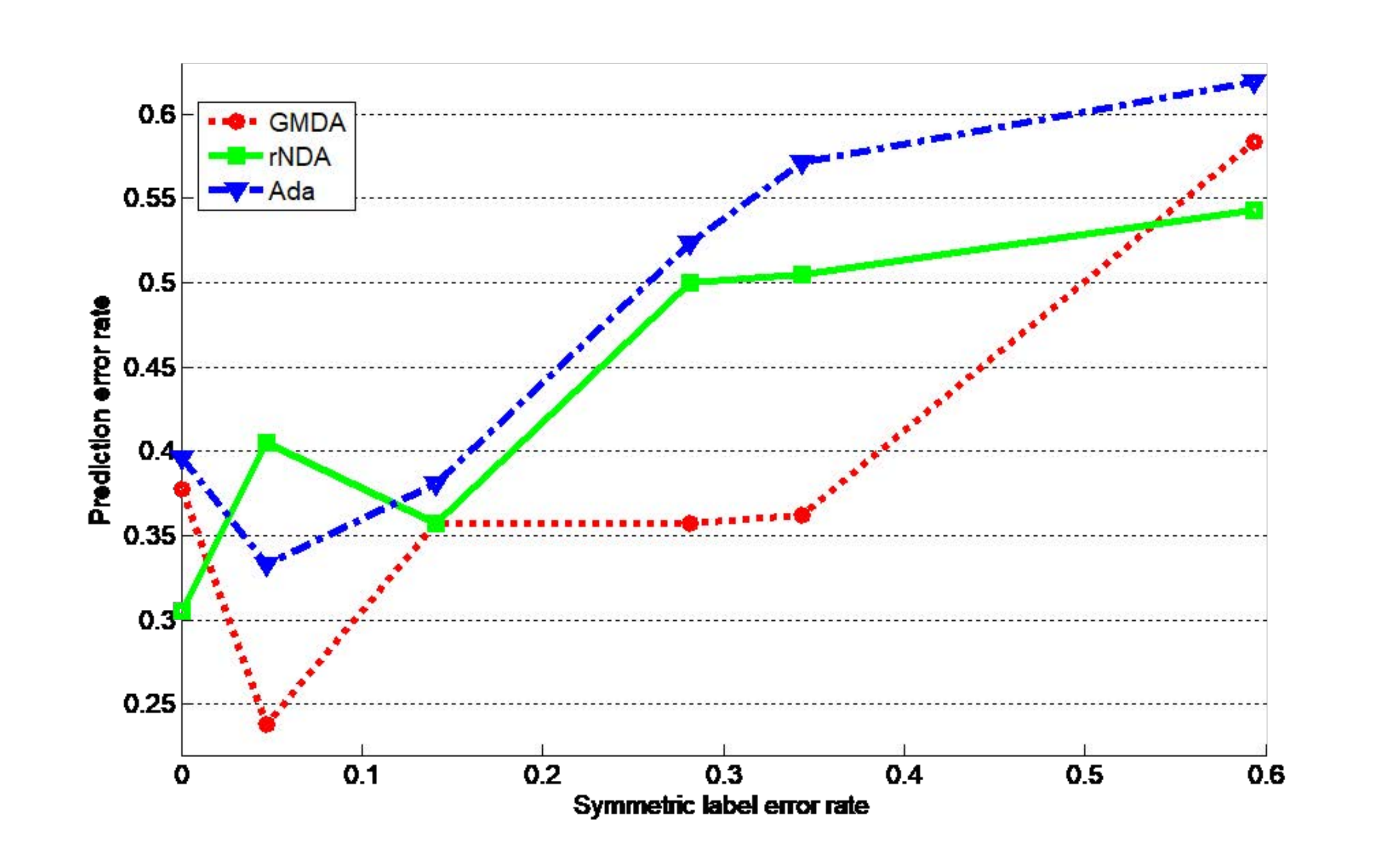}}
	\quad
	\subfigure[Breast Issue dataset with asymmetric label errors]{\includegraphics[width=5.5cm]{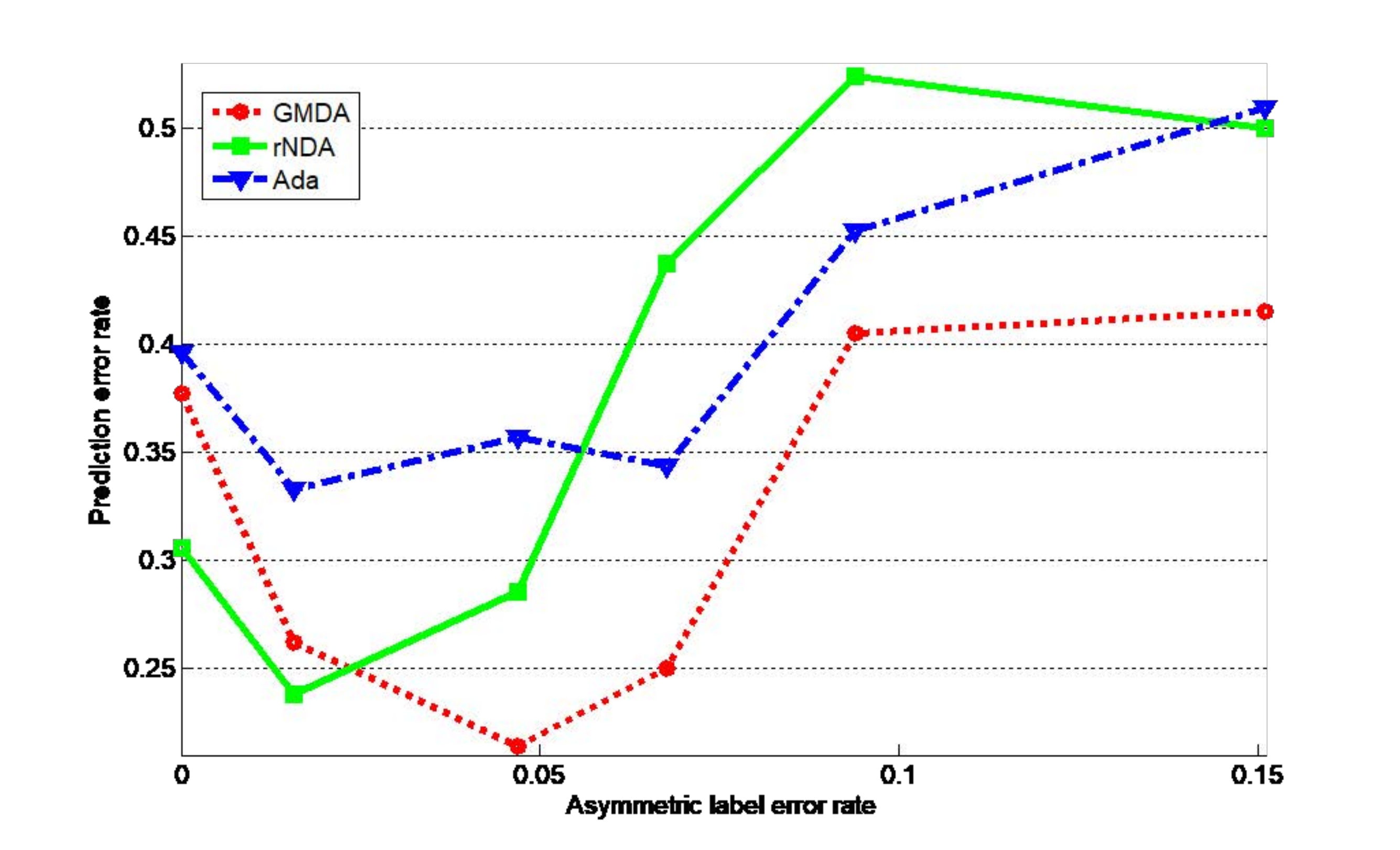}}
	\quad
	\caption{Results on different datasets with different noise levels using different methods}
\end{figure}

\begin{table}%%%表3
	\centering
	\caption{PREDICTION ERROR RATE ON TWO SYNTHETIC DATASETS AND SIX REAL-WORLD DATASETS WITH DIFFERENT SYMMETRIC LABEL NOISE RATES}
	\begin{tabular}{cccccccc}
		\hline
		\multirow{2}{*}{Dataset} & \multirow{2}{*}{Method} & \multicolumn{6}{c}{\textbf{Symmetric label noise rate}}\\ 
		 \cline{3-8}& ~ & 0 & 	0.1 &	 0.2 & 	0.3 & 	0.4 & 	0.5\\
		\hline
		\multirow{5}{*}{Synth1} & Ada & 	0.03 &	0.056 & 	0.061 & 	0.078 & 	0.116 &	0.224\\
		~ & rNDA & 	0.035 & 	0.064 & 	0.064 & 	0.065 & 	0.065 & 	0.065\\
		~ & GMDA & 	\textbf{0.005} & 	\textbf{0.003} &	\textbf{0.003} &	\textbf{0.003} & 	\textbf{0.030} &	\textbf{0.003}\\
		~ & rmLR & 	0.142 & 	0.138 &  	0.138 & 	0.184 & 	0.504 & 	0.504\\
		~ & rLR & 	0.147 & 	0.138 & 	0.138 &	0.17 & 	0.186 & 	0.504\\
		\hline
		\multirow{3}{*}{Synth2} & Ada & 	0.142 & 	0.14 & 	0.142 & 	0.184 &	0.226 &	0.318\\
		~ & rNDA &	0.138 & 	0.138 & 	0.138 & 	0.266 &	0.524 & 	0.532\\
		~ & GMDA &	\textbf{0.128} & 	\textbf{0.120} &	\textbf{0.126} &	\textbf{0.166} &	0.19 &	\textbf{0.182}\\
		\hline
		\multirow{3}{*}{Breast Issue} & Ada & 	0.396 & 	0.333 & 	0.381 &	0.523 &	0.571 & 	0.619\\
		~ & rNDA &	\textbf{0.305} &	0.404 &	\textbf{0.357} & 	0.5 & 	0.504 & 	0.542\\
		~ & GMDA &	0.377 & 	\textbf{0.238} & 	\textbf{0.357} &	\textbf{0.357} & 	\textbf{0.500} &	\textbf{0.504}\\
		\hline
		\multirow{3}{*}{Iris} & Ada & 	0.026 & 	0.05 & 	0.183 & 	0.200 &	0.333 &	0.233\\
		~ & rNDA & 	0.026 &	0.033 & 	\textbf{0.033} &	0.05 &	0.100 &	\textbf{0.08}\\
		~ & GMDA & 	\textbf{0.013} &	\textbf{0.016} &	\textbf{0.033} &	\textbf{0.05} &	\textbf{0.083} &	\textbf{0.08}\\
		\hline
		\multirow{5}{*}{Wine} & Ada & 	0.078 &	0.123 & 	0.112 &	0.157 &	0.236 & 	0.606\\
		~ & rNDA &	0.044 &	\textbf{0.011} &	\textbf{0.044} &	0.044 &	\textbf{0.044} &	0.078\\
		~ & GMDA &	\textbf{0.033} &	0.022 &	\textbf{0.044} &	\textbf{0.033} &	0.045 &	\textbf{0.076}\\
		~ & rmLR & 0.548 &	0.654 & 	0.691 &	0.737 &	0.6 & 	0.5\\
		~ & rLR &	0.533 &	0.635 &	0.672 &	0.7 & 	0.65 &	0.6\\
		\hline
		\multirow{5}{*}{Heart} & Ada & 	0.270 &	\textbf{0.261} &	0.299 &	0.537 &	0.32 & 	0.537\\
		~ & rNDA & 	0.609 &	0.626 &	0.607 &	0.637 &	0.39 &	0.7\\
		~ & GMDA & 	\textbf{0.248} &	\textbf{0.261} &	\textbf{0.261} &	\textbf{0.289} &	\textbf{0.287} &	\textbf{0.271}\\
		~ & rmLR & 	\textbf{0.130} &	0.138 &	0.128 &	0.193 & 	0.336 & 	0.514\\
		~ & rLR & 	\textbf{0.130} &	\textbf{0.183} & 	\textbf{0.113} &	0.188 & 	0.316 &	0.638\\
		\hline
		\multirow{3}{*}{Boston} & Ada & 	0.17 &	0.158 &	0.163 &	\textbf{0.183} &	0.292 &	0.549\\
		~ & rNDA & 	0.233 &	0.198 &	0.198 &	0.198 &	0.445 & 	0.717\\
		~ & GMDA &	0.205 &	\textbf{0.183} & 	0.188 &	\textbf{0.183} &	\textbf{0.188} &	\textbf{0.185}\\
		\hline
		\multirow{3}{*}{Wave form} & Ada & 	\textbf{0.188} &	\textbf{0.201} &	\textbf{0.223} &	\textbf{0.223} & 0.290 &	0.312\\
		~ & rNDA &	0.293 &	0.248 &	0.249 &	0.251 &	0.252 &	0.253\\
		~ & GMDA &	0.222 &	0.228 & 	0.231 &	0.232 &	\textbf{0.244} & 	\textbf{0.244}\\
		\hline

	\end{tabular}
\end{table}
\begin{table}%%%表4
	\centering
	\caption{PREDICTION ERROR RATE ON TWO SYNTHETIC DATASETS AND SIX REAL-WORLD DATASETS WITH DIFFERENT ASYMMETRIC LABEL NOISE RATES}
	\begin{tabular}{cccccccc}
		\hline
		\multirow{2}{*}{Dataset} & \multirow{2}{*}{Method} & \multicolumn{6}{c}{\textbf{Asymmetric label noise rate}}\\ 
		\cline{3-8}& ~ & 0 & 	0.1 &	 0.2 & 	0.3 & 	0.4 & 	0.5\\
		\hline
		\multirow{5}{*}{Synth1} & Ada & 	0.03 & 	0.073 & 	0.085 &	0.138 & 	0.211 &	0.221\\
		~ & rNDA & 	0.035 &	0.035 & 	0.067 &	0.067 & 	0.065 &	0.79\\
		~ & GMDA & 	\textbf{0.005} & 	\textbf{0.005} &	\textbf{0.005} &	\textbf{0.005} & 	\textbf{0.026} & 	\textbf{0.191}\\
		~ & rmLR &		0.142 &	0.138 &	0.148 & 	0.138 &	0.192 & 	/\\
		~ & rLR &	0.147 &	0.138 &	0.142 & 	0.134 &	0.23 & 	/\\
		\hline
		\multirow{3}{*}{Synth2} & Ada  & 	0.142 &	0.142 & 	0.296 &	0.468 & 	0.494 & 	/\\
		~ & rNDA & 	0.138 &	\textbf{0.128} &	0.144 & 	0.144 &	0.148 &	\textbf{0.504}\\
		~ & GMDA & 	\textbf{0.128} &	\textbf{0.128} &	\textbf{0.132} & 	\textbf{0.13} & 	\textbf{0.144} & 	0.52\\
		\hline
		\multirow{3}{*}{Breast Issue} & Ada & 	0.396 &	0.333 &	0.357 & 	0.343 &	0.452 & 	0.509\\
		~ & rNDA & 	\textbf{0.305} &	\textbf{0.238} &	0.285 &	0.437 &	0.523 & 	0.5\\ 
		~ & GMDA & 	0.377 &	0.261 &	\textbf{0.214} &	\textbf{0.25} & 	\textbf{0.404} &	\textbf{0.415}\\
		\hline
		\multirow{3}{*}{Iris} & Ada & 	0.026 &	0.016 &	0.183 &	0.177 &	0.433 &	/\\
		~ & rNDA &	0.026 &	\textbf{0.016} &	\textbf{0.016} & 	\textbf{0.022} &	\textbf{0.033} &	/ \\
		~ & GMDA & 	\textbf{0.013} & 	\textbf{0.016} & 	\textbf{0.016} &	\textbf{0.022} & 	\textbf{0.033} &	/\\
		\hline
		\multirow{5}{*}{Wine} & Ada & 	0.078 &	0.140 & 	0.112 & 	0.197 &	0.408 &	/\\
		~ & rNDA & 	0.044 &	0.056 & 	0.056 &	\textbf{0.042} &	\textbf{0.056} &	/\\
		~ & GMDA &	\textbf{0.033} &	\textbf{0.042} &	\textbf{0.042} &	\textbf{0.042} &	\textbf{0.056} & 	/\\
		~ & rmLR &	0.548 &	0.448 &	0.336 & 	0.355 & 	0.700 &	/\\
		~ & rLR &	0.533 &	0.420 &  	0.411 & 	0.336 & 	0.672 &	/\\
		\hline
		\multirow{5}{*}{Heart} & Ada & 	0.270 & 	0.299 &	0.307 &	0.392 &	0.719 &	/\\
		~ & rNDA &	0.609 &	0.364 &	0.271 &	0.364 &	0.729 &	/\\
		~ & GMDA & 	\textbf{0.248} &	\textbf{0.261} &	\textbf{0.261} &	\textbf{0.261} &	\textbf{0.327} &	/\\
		~ & rmLR & 	\textbf{0.130} & 	\textbf{0.163} &	\textbf{0.173} & 	\textbf{0.178} &	0.198 &	0.336\\
		~ & rLR & 	\textbf{0.130} & 	\textbf{0.163} &	0.183 &	0.173 &	0.193 &	0.347\\
		\hline
		\multirow{3}{*}{Boston} & Ada & 	0.17 &	0.178 & 	0.262 & 	0.415 & 	0.485 &	0.442\\
		~ & rNDA & 	0.233 & 	0.198 & 	0.198 & 	0.198 & 	0.445 & 	0.233\\
		~ & GMDA &	0.205 &	0.168 & 	0.183 &	0.183 & 	\textbf{0.163} &	\textbf{0.221}\\
		\hline
		\multirow{3}{*}{Wave form} & Ada & 	\textbf{0.188} &	\textbf{0.196} &	0.286 &	0.339 &	0.390 & 	/\\
		~ & rNDA & 	0.293 & 	0.23 &	0.231 & 	0.493 & 	0.500 &	/\\
		~ & GMDA & 	0.222 &	0.227 & 	\textbf{0.226} & 	\textbf{0.280} &	\textbf{0.296} &	/\\
		\hline

	\end{tabular}
\end{table}
\begin{table}%%%表5
	\centering
	\caption{WIN / DRAW / LOSE}
	\begin{tabular}{cccc}
		\hline
		\multirow{2}{*}{Dataset} & \multirow{2}{*}{Method} & \multicolumn{2}{c}{Win/Draw/Lose}\\
		\cline{3-4} &  ~ & Symmetric & 	Asymmetric\\
		\hline
		\multirow{5}{*}{Synth1} & Ada & 	0/0/6 & 	0/0/6\\
		~ & rNDA & 	0/0/6 & 	0/0/6\\
		~ & GMDA & 	6/0/0 & 	6/0/0\\
		~ & rmLR & 	0/0/6 & 	0/0/6\\
		~ & rLR & 	1/0/5 & 	0/0/6\\
		\hline
		\multirow{3}{*}{Synth2} & Ada & 	0/0/6 & 	0/0/6\\
		~ & rNDA & 	0/0/6 &	1/1/4\\
		~ & GMDA & 	5/0/1 &	4/1/1\\
		\hline
		\multirow{3}{*}{Breast Issue} & Ada & 	0/0/6 &	0/0/6\\
		~ & rNDA & 	1/1/4 & 	2/0/4\\
		~ & GMDA & 	4/1/1 &	4/0/2\\
		\hline
		\multirow{3}{*}{Iris} & Ada & 	0/0/6 & 	0/0/5\\
		~ & rNDA & 	0/3/3 &	0/4/1\\
		~ & GMDA & 	3/3/0 & 	1/4/0\\
		\hline
		\multirow{5}{*}{Wine} & Ada & 	0/0/6 &	0/0/5\\
		~ & rNDA & 	1/2/3 &	0/2/3\\
		~ & GMDA &		3/2/1 & 	3/2/0\\
		~ & rmLR & 	0/0/6 & 	0/0/5\\
		~ & rLR &  	0/0/6 &	0/0/5\\
		\hline
		\multirow{5}{*}{Heart} & Ada & 	0/1/5 & 	0/0/5\\
		~ & rNDA & 	0/0/6 & 	0/0/5\\
		~ & GMDA & 	5/1/0 & 	5/0/0\\
		~ & rmLR & 	1/1/4 &	1/2/3\\
		~ & rLR &  	1/1/4 &	1/2/3\\
		\hline
		\multirow{3}{*}{Boston} & Ada & 	0/1/5 &	0/0/6\\
		~ & rNDA & 	0/0/6 &	0/0/6\\
		~ & GMDA & 	2/1/3 & 	2/0/4\\
		\hline
		\multirow{3}{*}{Wave form} & Ada  & 	4/0/2 & 	2/0/3\\
		~ & rNDA &	   0/0/6 & 	0/0/5\\
		~ & GMDA & 	2/0/4 & 	3/0/2\\
		\hline

	\end{tabular}
\end{table}

Fig. 7 shows the results on different datasets with different dimensions using different methods. The trend is that with a higher dimension, a lower error rate is obtained. The proposed method performs much better than the others in all dimensionality cases. \\

We employ a set of synthetic datasets to study the effect of the mixture number of the datasets and that of the component number of the model. Each dataset contains 1000 samples with 10 dimensions and comprises three classes, component numbers from 1 to 5 for each class. Fig. 7 shows that the more components in each class the dataset has, the larger the error rate is. In almost all cases, the proposed method notably outperforms the other methods.
\begin{figure}[htbp]%%%图7
	\centering
	\includegraphics[scale=0.4]{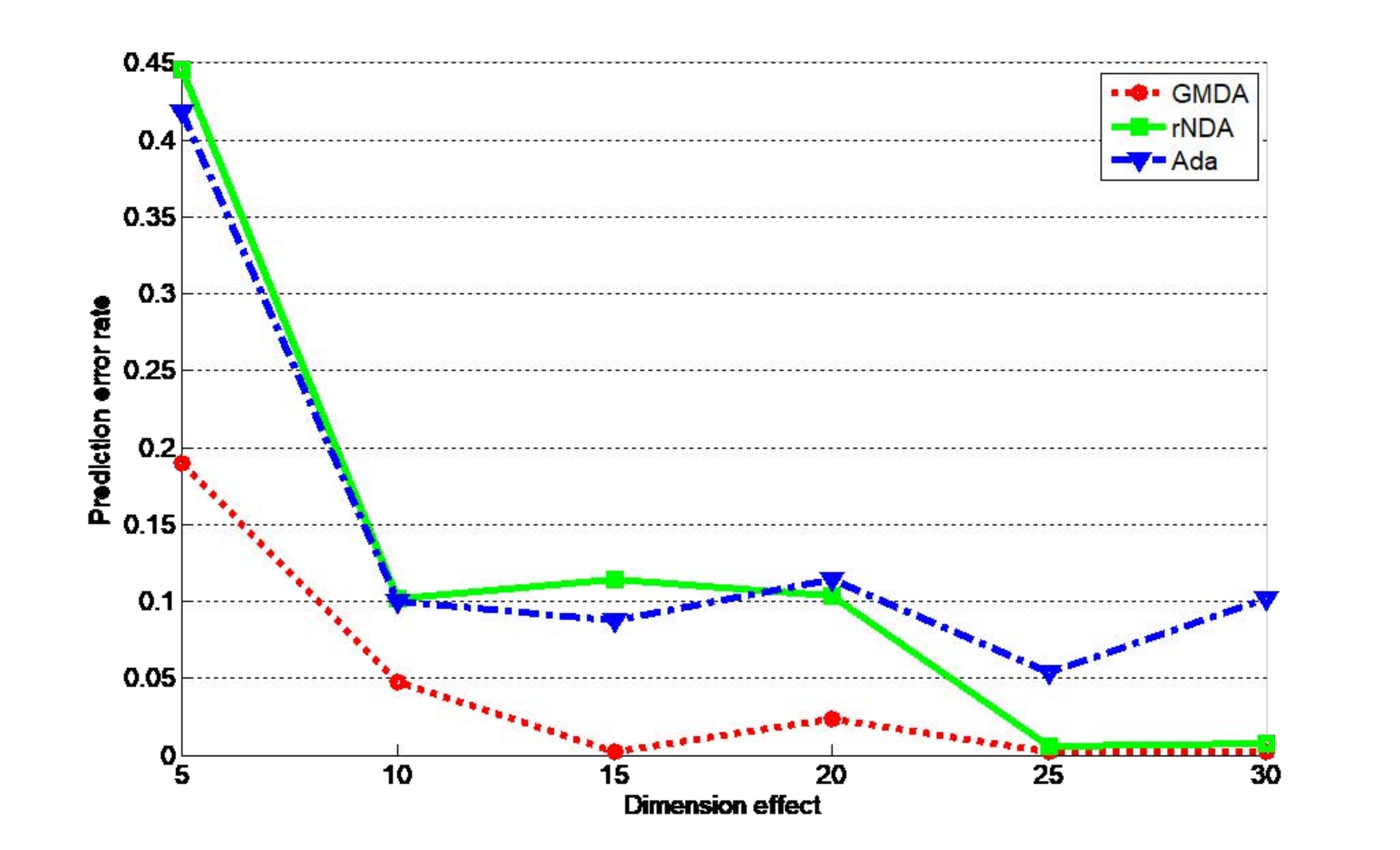}
	\caption{Effect of the dimension of datasets}
\end{figure}
According to Fig. 6 and Tables III, IV, V, our method achieves better performance in a multi-class case than in a two-class case. We run one more experiment to investigate the effect of the number of classes. Datasets used are created artificially as well; they all contain 1000 samples with 10 dimensions. Fig. 8 shows that our proposed method is affected much less than the other methods.
\begin{figure}[htbp]%%%图8
	\centering
	\includegraphics[scale=0.4]{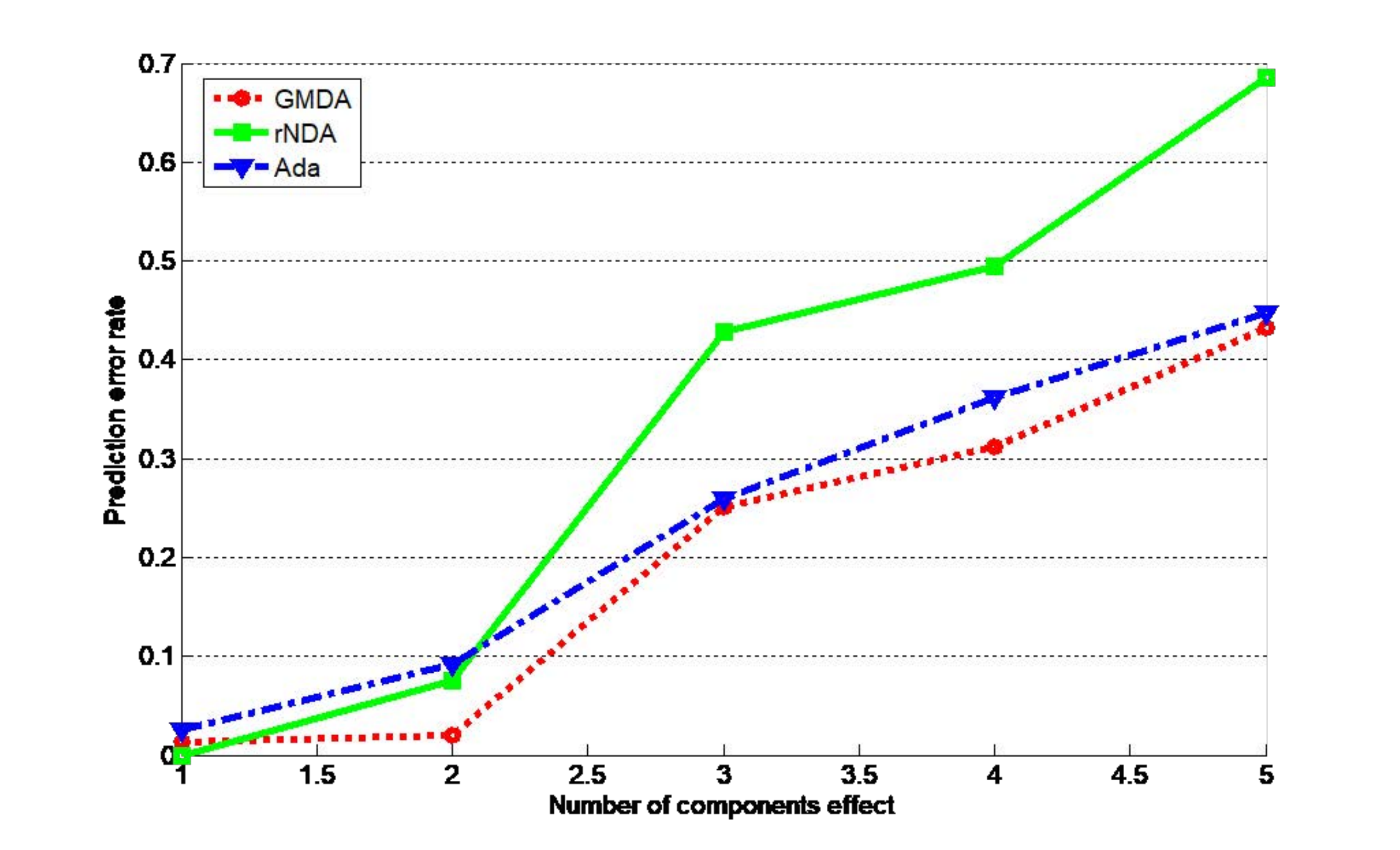}
	\caption{Effect of the number of components of the model (number of dataset mixtures is 3)}
\end{figure}
Our previous experiments on synthetic datasets have been conducted on the premise of taking the correct mixture number. In this paper, we tested the datasets using an invalid mixture number. We take a three-component dataset as a representative and set the component number of the model from 1 to 6. From Fig. 9, we can discern that an invalid mixture number makes an increment on error rate, whereas rNDA and AdaBoost have error rates of 0.5040 and 0.4660, respectively. Our method still performs remarkably better even when an invalid mixture number is used. Poor performance is expected when the number of the component of the number is 1.\\

The EM method we employed finds the local optimum each time; the initial cluster center is significant. The initial values used in this paper are obtained by k-means method, which is very sensitive to initial cluster centers. A bad initial cluster center leads to poor cluster performance and affects the error rate of the proposed method. Moreover, the difference between samples in the same class is much smaller than that between classes; thus, obtaining proper initial values for our method is difficult. The mixture model proposed in this paper is finite; the number of mixture components is provided in advance and cannot be changed to adapt to a simpler or more complicated situation. Thus, the estimation of the number of components and adapted mixture number remain to be studied.
\begin{figure}[htbp]%%%图9
	\centering
	\includegraphics[scale=0.4]{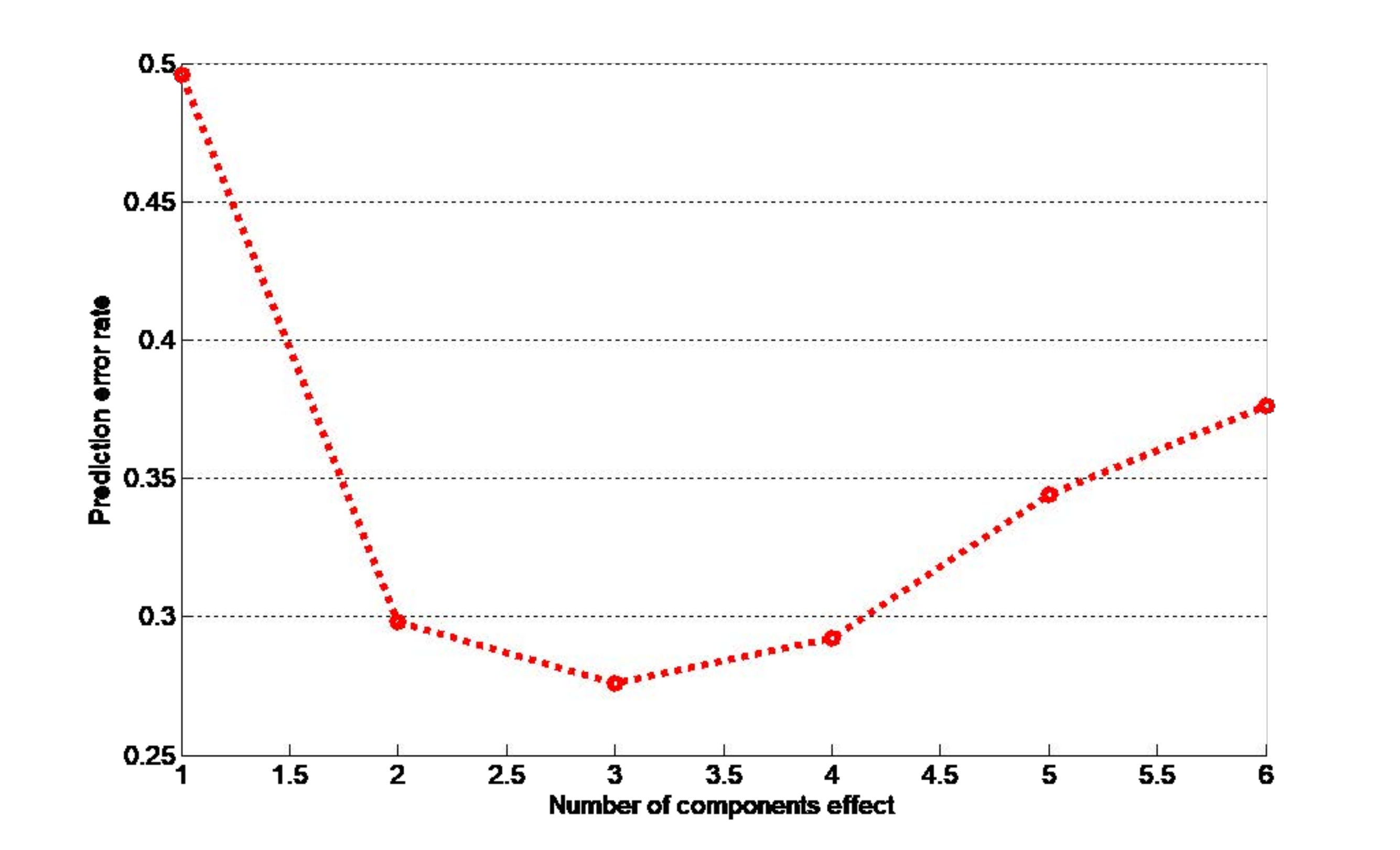}
	\caption{Effect of the number of classes}
\end{figure}
\begin{figure}[htbp]%%%图10
	\centering
	\includegraphics[scale=0.4]{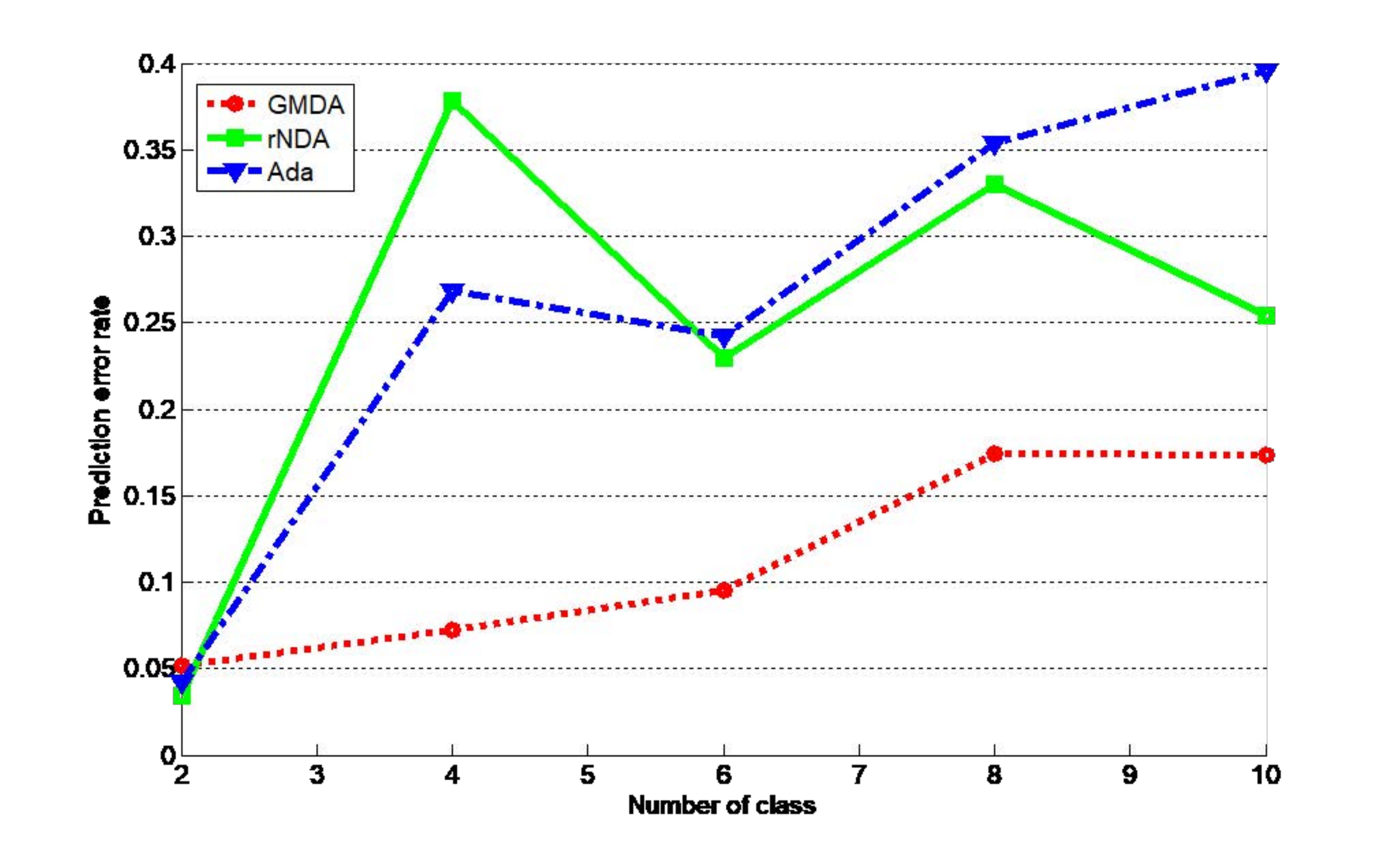}
	\caption{Effect of the number of classes}
\end{figure}
\section{EXPERIMENTAL RESULTS ON LARGE SCALE DATA SETS}%6
To verify the performance of our proposed approach on large scale data sets, we employ six synthetic datasets to study the effect of the mixture number of the datasets and that of the component number of the model. Each dataset contains 15000 samples with 200 dimensions and comprises two classes, component numbers from 1 to 5 for each class. The 6 datasets together with their sizes N and number of features D are listed in TABLE VI. More specifically, first, we randomly generate six synthetic datasets for verification goal. The data set is a mixture of two types of labels, with the covariance range from 0 to 250, which means that the correlation between these two types of labels is from uncorrelation to the maximum correlation. Similarly, for the division of data sets according to the TABLE VI, we have carried out 10 fold, 5 fold and 3 fold cross-validation respectively. In addition, we add noise labels based on 0\%, 10\%, 20\%, 30\%, 40\% and 50\% of the total number of labels. TABLE VII shows the error rate with 5, 10, and 3-cross-validation for all comparison methods. In the TABLE VII, (a1), (b1), and (c1) summarize the error rates on synthetic datasets with label correlation group, (a2), (b2), and (c2) summarize the error rates on synthetic datasets with label uncorrelation group. Fig. 11 shows the learning performances for all comparison methods, similarly, in the Fig. 11, (a1), (b1), and (c1) are the learning performances on synthetic datasets with label correlation group, (a2), (b2), and (c2) are the learning performances on synthetic datasets with label uncorrelation group.
\begin{table}%%%表6
	\centering
	\caption{CHARACTERISTICS OF THE LARGE SCALE DATASETS}
	\begin{tabular}{ccccccc}
		\hline
		\multirow{2}{*}{Dataset} &  \multicolumn{6}{c}{Characteristics}\\
		\cline{2-7} & N & 	D & 	covariance & 	ross-validation & 	Number of samples in class 1 & 	Number of samples in lass 2\\
		\hline
		Synth1 & 	15000 &	200 &	 250 & 	5 & 	7592 & 	7408\\
		Synth2 &	15000 &	200 & 	0 &  	5 & 	6619 & 	8381\\
		Synth3 & 	15000 & 	200 & 	250 & 	10 & 	8434 & 	6566\\
		Synth4 & 	15000 & 	200 & 	0 &  	10 & 	6619 & 	8381\\
		Synth5 & 	15000 & 	200 & 	250 & 	3 &  	7707 & 	7293\\
		Synth6 & 	15000 & 	200 & 	0 & 	3 & 	7796 & 	7204\\
		\hline
	\end{tabular}
\end{table}
\begin{table}%%%表7
	\centering
	\caption{EXPERIMENTAL RESULTS IN ERROR RATE ON SIX DIFFERENT CORRELATED SYNTHETIC DATASETS WITH DIFFERENT CROSS -VALIDATION PROCESSES}
	\resizebox{.5\textwidth}{!}{
	\subtable[error rate with 5-cross-validation on correlated Synth1 dataset]{%%%子表1
	\begin{tabular}{ccccccc}
		\hline
		\multirow{2}{*}{\textbf{Method}} &  \multicolumn{6}{c}{label noise rate}\\
		\cline{2-7} & 0.0 &	0.1 & 	0.2 &	0.3 & 	0.4 & 	0.5\\
		\hline
GMDA & 	0.5023 & 	0.5023 &	0.5023 &	\textbf{0.5023} & 	\textbf{0.5023} & 	0.4977\\
rNDA &	0.5063 &	0.5063 &	0.5063 & 	0.5063 & 	0.5063 & 	\textbf{0.4937}\\
Ada & 	0.4953 &	\textbf{0.4853} & 	0.4997 & 	0.5007 &	0.5090 &	0.5047\\
rLR & 	0.4927 &	0.4907 & 	0.5057 & 	0.5017 & 	0.5110 & 	0.5087\\
rmLR & 	\textbf{0.4890} &	0.4877 &	\textbf{0.4953} & 	0.5080 & 	0.5080 & 	0.5090\\
		\hline
	\end{tabular}
	%\label{firsttable}
}}

\resizebox{.5\textwidth}{!}{
\subtable[error rate with 10-cross-validation on correlated Synth3 dataset]{%%%子表2
	\begin{tabular}{ccccccc}
		\hline
		\multirow{2}{*}{\textbf{Method}} &  \multicolumn{6}{c}{label noise rate}\\
		\cline{2-7} & 0.0 &	0.1 & 	0.2 &	0.3 & 	0.4 & 	0.5\\
		\hline
GMDA & 	0.5073 & 	0.5073 &	0.4927 & 	0.4927 & 	0.4927 & 	\textbf{0.4927}\\
rNDA & 	0.5067 &	0.5067 & 	0.4933 & 	0.4933 & 	0.4933 & 	0.4933\\
Ada & 	\textbf{0.4520} & 	\textbf{0.4520} & 	\textbf{0.4520} & 	\textbf{0.4520} & 	0.5480 & 	0.5480\\
rLR & 	0.5033 & 	0.5087 & 	0.4973 & 	0.4807 & 	0.4867 & 	0.4947\\
rmLR & 	0.5120 & 	0.4987 &  	0.4973 & 	0.4740 & 	\textbf{0.4827} & 	0.4967\\
		\hline
	\end{tabular}
	%\label{firsttable}
}}
\resizebox{.5\textwidth}{!}{
\subtable[error rate with 3-cross-validation on correlated Synth5 dataset]{%%%子表3
	\begin{tabular}{ccccccc}
		\hline
		\multirow{2}{*}{\textbf{Method}} &  \multicolumn{6}{c}{label noise rate}\\
		\cline{2-7} & 0.0 &	0.1 & 	0.2 &	0.3 & 	0.4 & 	0.5\\
		\hline
GMDA & 	0.5060 &	0.5060 & 	0.5060 &	0.5060 & 	0.5060 & 	\textbf{0.4938}\\
rNDA & 	0.4944 &	0.5056 &	0.5056 & 	0.5056 & 	0.5056 &	0.5056\\
Ada & 	\textbf{0.4812} & 	\textbf{0.4812} & 	\textbf{0.4812} & 	\textbf{0.4812} &	0.5188 & 	0.5188\\
rLR & 	0.4838 &	0.4956 & 	0.4884 & 	0.4818 & 	0.4934 & 	0.5144\\
rmLR &	0.4846 &	0.4908 & 	0.4904 & 	0.4950 & 	\textbf{0.4904} & 	0.5152\\
		\hline
	\end{tabular}
	%\label{firsttable}
}}
\resizebox{.5\textwidth}{!}{
\subtable[error rate with 5-cross-validation on uncorrelated Synth2 dataset]{%%%子表4
	\begin{tabular}{ccccccc}
		\hline
		\multirow{2}{*}{\textbf{Method}} &  \multicolumn{6}{c}{label noise rate}\\
		\cline{2-7} & 0.0 &	0.1 & 	0.2 &	0.3 & 	0.4 & 	0.5\\
		\hline
GMDA & 	0.5570 &	0.5570 &	0.5570 & 	0.5577 & 	0.9640 & 	0.5570\\
rNDA & 	0.4427 & 	0.4427 &	0.4427 & 	\textbf{0.4427} & 	\textbf{0.4427} & 	0.4427\\
Ada & 	0.4427 &	0.4427 & 	0.4427 & 	0.5573 & 	0.5573 &	0.4427\\
rLR & 	\textbf{0.1653} & 	\textbf{0.1103} & 	\textbf{0.3853} & 	0.5573 & 	0.9483 &	\textbf{0.1653}\\
rmLR &	0.2730 & 	0.2730 & 	0.4303 & 	0.5477 & 	0.7920 &	0.2730\\
		\hline
	\end{tabular}
	%\label{firsttable}
}}
\resizebox{.5\textwidth}{!}{
\subtable[error rate with 10-cross-validation on uncorrelated Synth4 dataset]{%%%子表5
	\begin{tabular}{ccccccc}
		\hline
		\multirow{2}{*}{\textbf{Method}} &  \multicolumn{6}{c}{label noise rate}\\
		\cline{2-7} & 0.0 &	0.1 & 	0.2 &	0.3 & 	0.4 & 	0.5\\
		\hline
GMDA &	\textbf{0.0420} &  	0.5440 & 	0.5453 & 	0.5440 & 	0.5473 & 	0.9613\\
rNDA & 	0.4560 &	0.4560 & 	0.4560 &	0.4560 &	0.4560 &	\textbf{0.4560}\\
Ada & 	0.4560 &	0.4560 & 	0.4560 & 	0.4560 &	0.5440 &	0.5440\\
rLR & 	0.0647 &	\textbf{0.2313} & 	\textbf{0.2320} & 	\textbf{0.2133} &	\textbf{0.4267} &	0.9353\\
rmLR &	0.0600 & 	0.2313 &	0.3453 &	0.3733 &	0.4473 &	0.9400\\
		\hline
	\end{tabular}
	%\label{firsttable}
}}
\resizebox{.5\textwidth}{!}{
\subtable[error rate with 3-cross-validation on uncorrelated Synth6 dataset]{%%%子表6
	\begin{tabular}{ccccccc}
		\hline
		\multirow{2}{*}{\textbf{Method}} &  \multicolumn{6}{c}{label noise rate}\\
		\cline{2-7} & 0.0 &	0.1 & 	0.2 &	0.3 & 	0.4 & 	0.5\\
		\hline
GMDA & 	0.0424 &	0.5572 &	0.5572 & 	0.5570 & 	0.5673 & 	0.9628\\
rNDA & 	0.4422 & 	0.4422 &	0.4422 &	0.4422 &	0.4422 & 	0.4422\\
Ada & 	0.4422 &	0.4422 & 	0.4422 & 	0.4422 &	0.5578 & 	0.5578\\
rLR & 	0.0754 & 	0.1754 & 	0.1562 &	0.1902 & 	0.4370 & 	0.9264\\
rmLR & 	0.0708 & 	0.2686 & 	0.2376 & 	0.3258 & 	0.4577 & 	0.9112\\
		\hline
	\end{tabular}
	%\label{firsttable}
}}

\end{table}

\begin{figure}[htbp]%%%图11
	\centering
	\subfigure[correlated synth1 dataset]{\includegraphics[width=5.5cm]{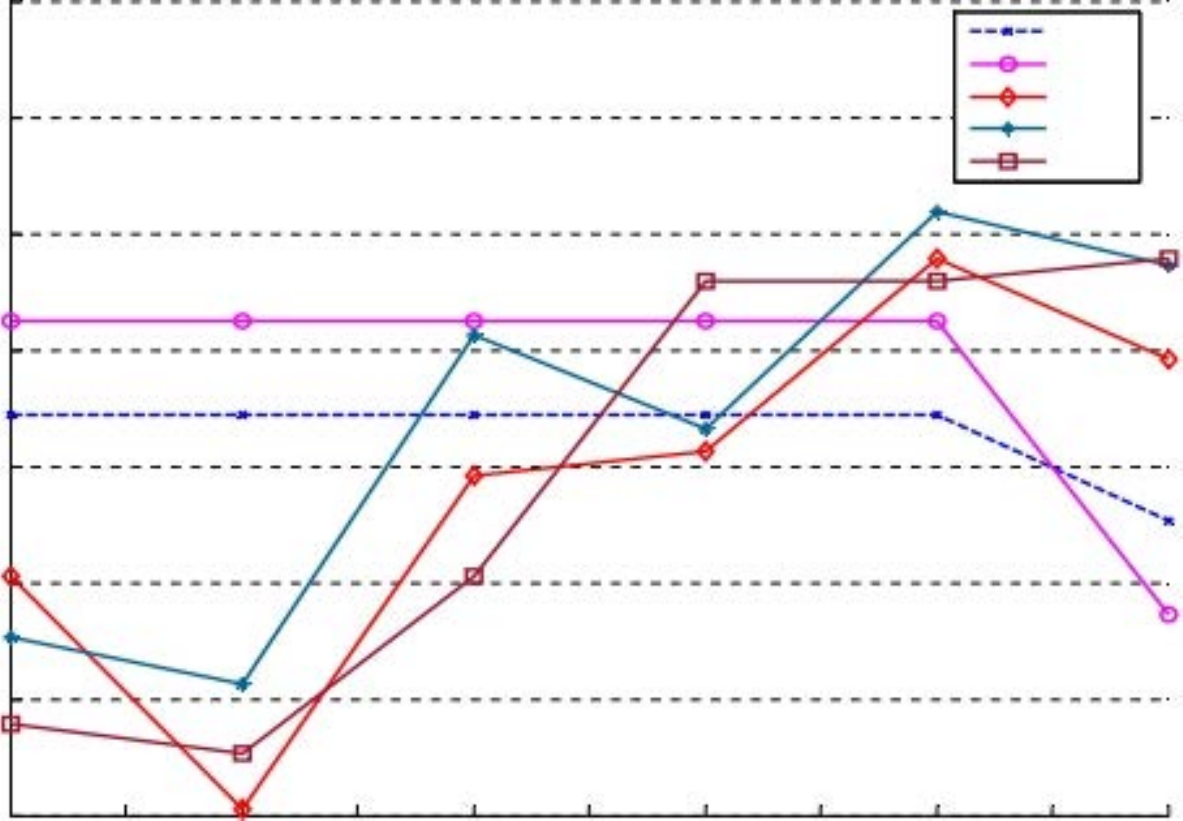}}
	\quad
	\subfigure[correlated synth3 dataset]{\includegraphics[width=5.5cm]{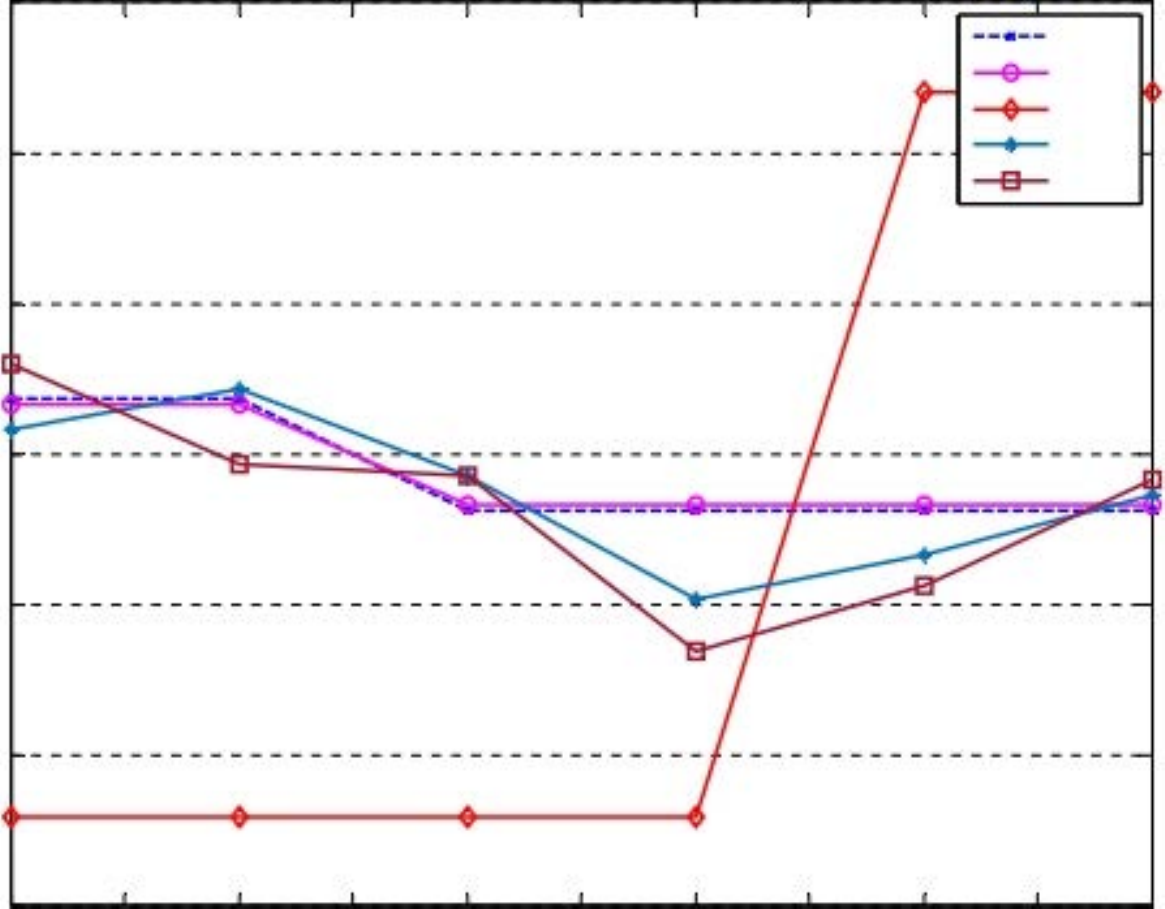}}
	\quad
	\subfigure[correlated synth5 dataset ]{\includegraphics[width=5.5cm]{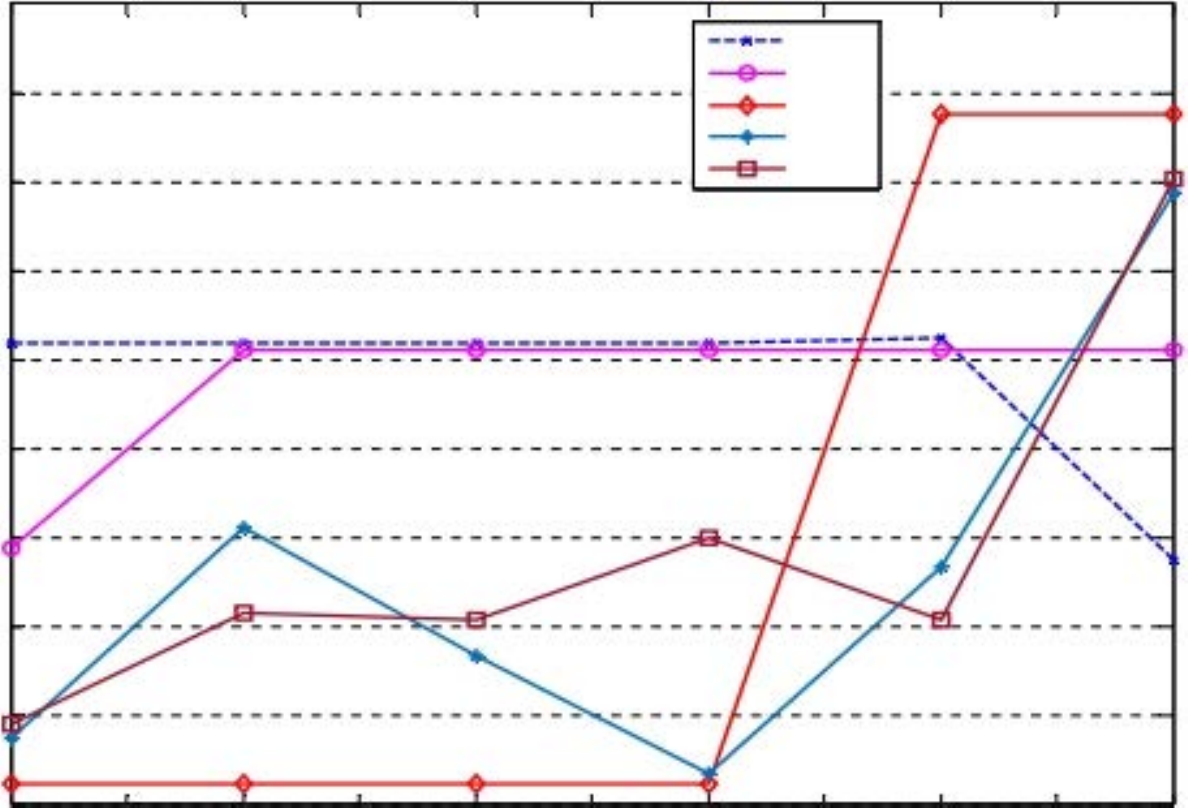}}
	\quad
	\subfigure[uncorrelated synth2 dataset ]{\includegraphics[width=5.5cm]{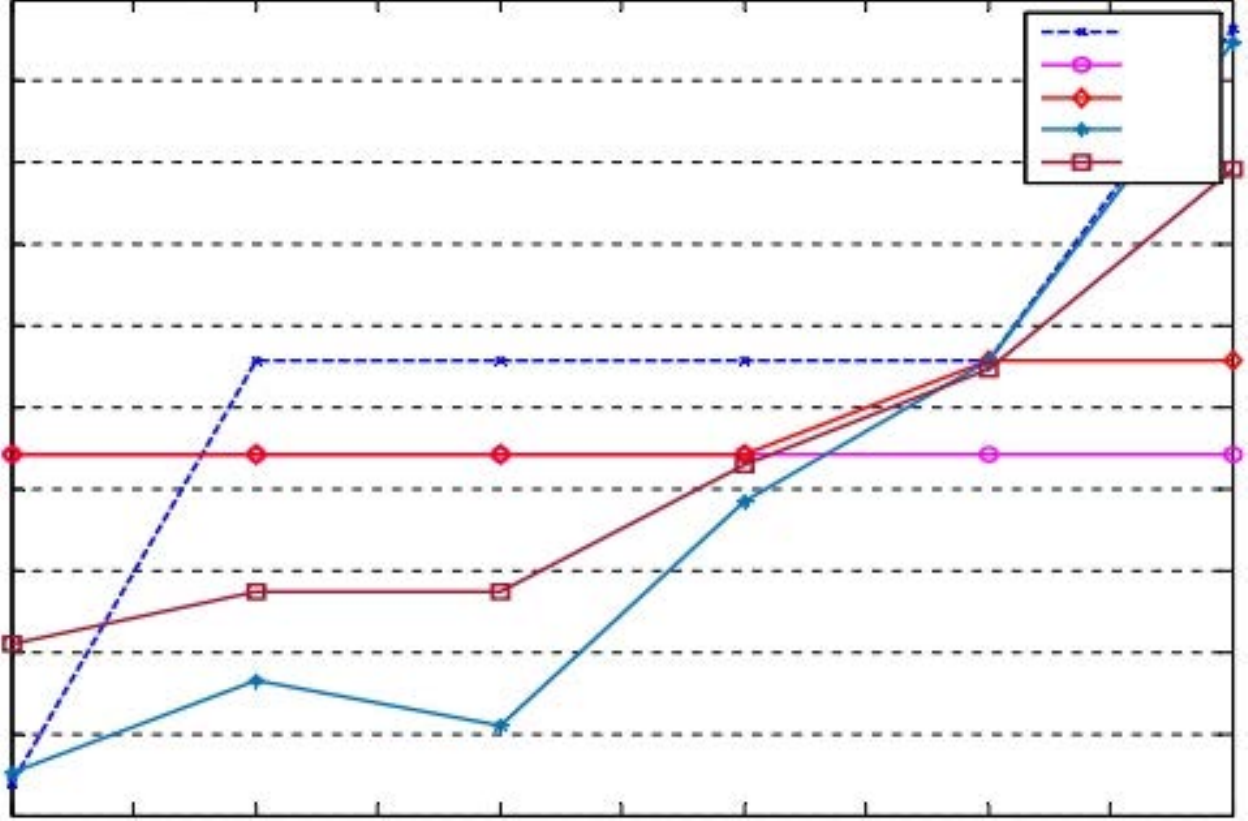}}
	\quad
	\subfigure[uncorrelated synth4 dataset]{\includegraphics[width=5.5cm]{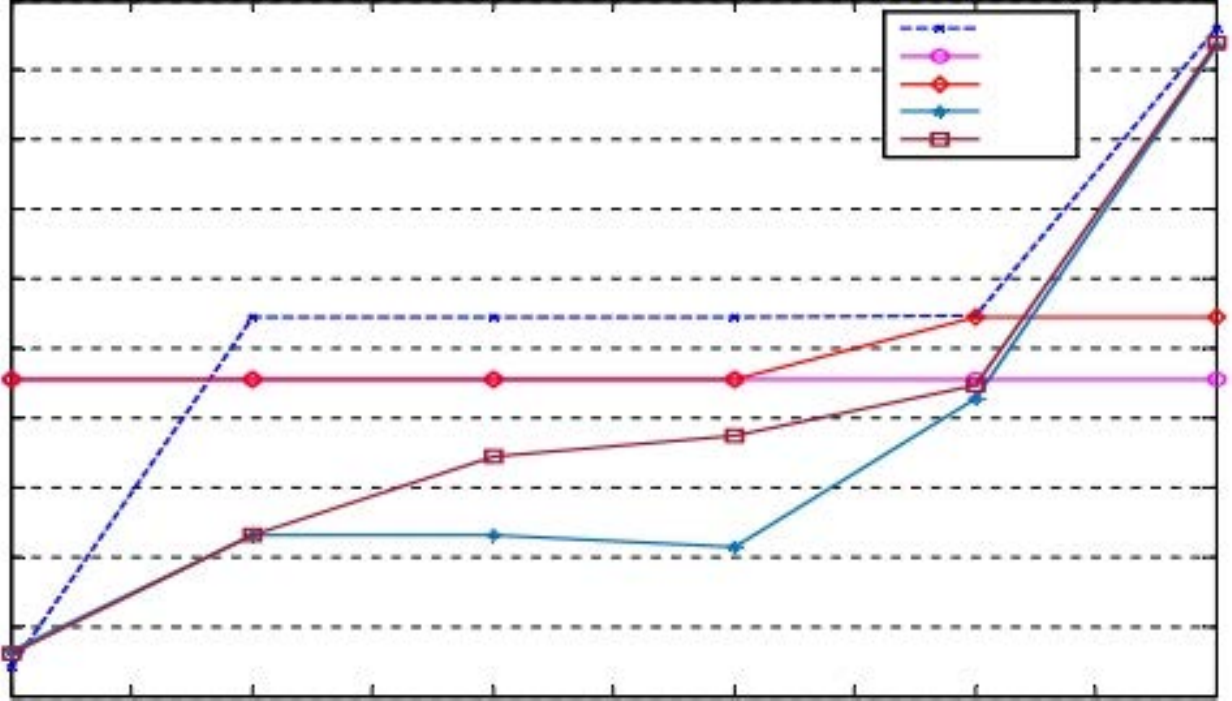}}
	\quad
	\subfigure[uncorrelated synth6 dataset ]{\includegraphics[width=5.5cm]{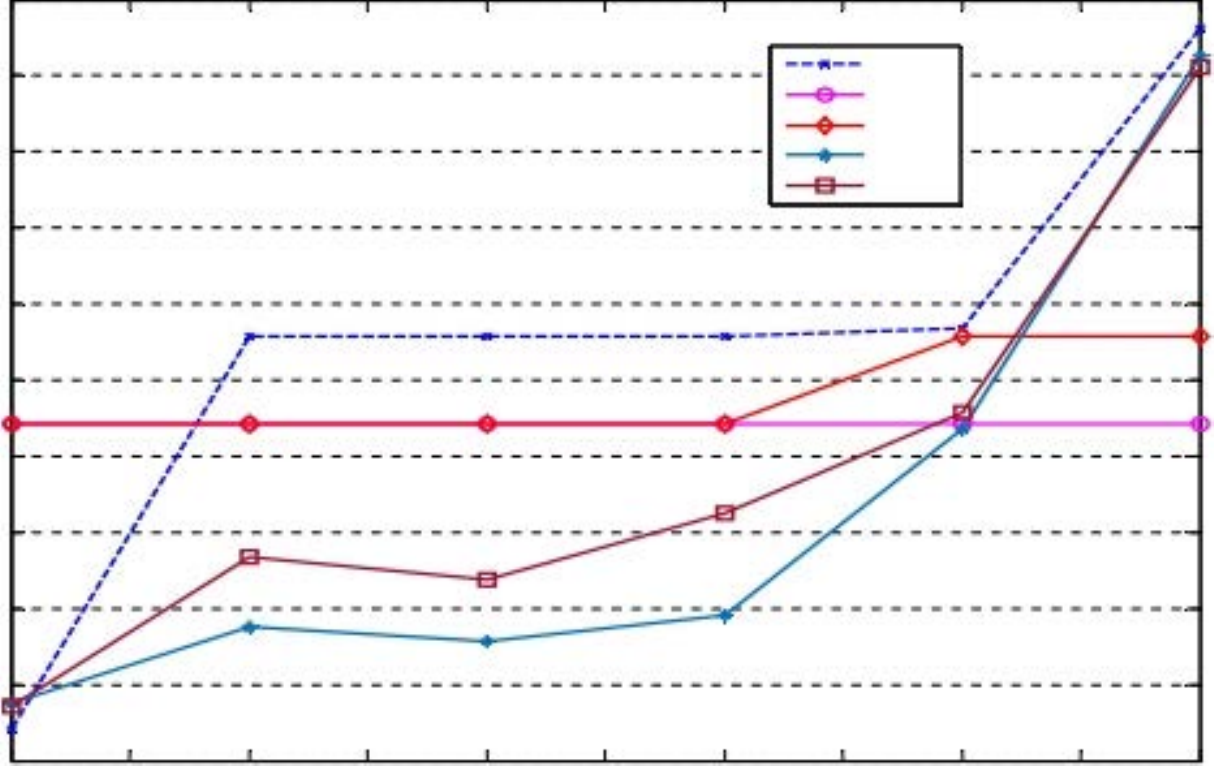}}
	\quad
\end{figure}
From TABLE VII and Fig. 11, we can see that:\\

(a) The number of these two types of tags in synthetic data sets is comparable to each other. They belong to the synthetic data sets with relatively balanced class size and large sample size, it has certain representativeness. \\

(b) When the characteristics difference between the two classes is small, that is, the covariance between two classes is equal to 250. The effect of the increase of noise labels is not very obvious. Compared with other methods, the GMDA has a certain capability of  noise resistance, and its error rate is up and down at 50\%. Secondly, with the increase of noise tags, the error rate of the GMDA decreases by about 1\%. Similar to the GMDA, under these synthetic datasets, the other comparison algorithms also fluctuate at specific values. We suspect that this is caused by the randomly generated synthetic datasets, and this small fluctuation does not affect the evaluation of the noise resistance performance of various models. \\

(c) When the characteristics difference between the two classes is large, that is, the covariance between two classes is equal to 0, the addition of noise tags have different degrees of impact on these algorithms. Generally speaking, the more noise tags are, the higher the classification error rate is. But when we look the label noise rate at the interval [0.2, 0.4], these algorithms all are with good anti-noise performance and have ability of active noise cancelation. It is worthy of note that ADA algorithm is different from the trend curve of other algorithms. When the difference between the two kinds of tags is small, the fluctuation is large. When the difference between the two kinds of tags is large, the fluctuation is small. We can choose the appropriate algorithm according to the actual data set. \\

In sum up, when the characteristics difference between the two classes in mixture model is obvious, it is meaningful to analyze and compare the experimental results. Whereas the correlation between the two classes in mixture model is large, the effect of the increase of noise labels is vague and limited. As increase of noise labels, prediction error rate does not change much.  

\section{CONCLUSION}%7
This paper presented a discriminant analysis based on Gaussian mixture models and applied to classification in the presence of label noise. We derived the updating formulas of the parameters. The experiments on two synthetic datasets and several real-world datasets showed that the proposed method was convergent and effective and mostly outperformed the other methods. Compared with the other methods, our method was less affected by the factors discussed in the preceding sections.

We found that the number of training samples affected the performance significantly, that is, the number of training samples is increased if necessary. If the samples were insufficient for maximum likelihood to estimate, Bayes estimation was used, where prior information was utilized, or domain adaptation learning was used, where a source dataset that was akin to a target dataset was used to help.

The number of components of a model given in advance may not be adapted to all the classes; it might lead to further calculation on a simpler case or less approximation on a more complicated case. Therefore, we considered a more flexible and adaptable infinite mixture model that estimates the hidden number of components from the training datasets. 

\section{ACKNOWLEDGMENTS}%8
The authors would like to thank the reviewers for their valuable comments and suggestions.
\bibliography{GMM}
\end{document}